\documentclass[acmtog]{acmart}
\acmSubmissionID{563}

\usepackage{booktabs} 

\usepackage{amsmath}
\usepackage{graphicx}
\usepackage{multirow,bigstrut}
\usepackage{soul,color} 
\usepackage{float}

\usepackage[ruled]{algorithm2e} 

\SetAlFnt{\small}
\SetAlCapFnt{\small}
\SetAlCapNameFnt{\small}
\SetAlCapHSkip{0pt}

\copyrightyear{2022} 
\acmYear{2022} 
\acmBooktitle{SIGGRAPH Asia 2022 Conference Papers (SA '22 Conference Papers), December 6--9, 2022, Daegu, Republic of Korea}
\acmPrice{15.00}
\acmDOI{10.1145/3550469.3555424}
\acmISBN{978-1-4503-9470-3/22/12}

\setcopyright{acmlicensed}\acmConference[SA '22 Conference Papers]{SIGGRAPH Asia 2022 Conference Papers}{December 6--9, 2022}{Daegu, Republic of Korea}

\acmDOI{10.1145/3550469.3555424}
\acmISBN{978-1-4503-9470-3/22/12}

\begin{document}
\title{Reconstructing editable prismatic CAD from rounded voxel models}


\newif\ifshowcomments
\showcommentsfalse
\ifshowcomments
\newcommand{\joe}[1]{{\color{red}{[Joe: #1]}}}
\newcommand{\karl}[1]{{\color{purple}{[Karl: #1]}}}
\newcommand{\pradeep}[1]{{\color{cyan}{[Pradeep: #1]}}}
\newcommand{\aditya}[1]{{\color{green}{[Aditya: #1]}}}
\newcommand{\pete}[1]{{\color{yellow}{[Pete: #1]}}}
\newcommand{\hooman}[1]{{\color{blue}{[Hooman: #1]}}}
\newcommand{\markchange}[1]{{\color{blue}{#1}}}
\else
\newcommand{\joe}[1]{}
\newcommand{\karl}[1]{}
\newcommand{\pradeep}[1]{}
\newcommand{\aditya}[1]{}
\newcommand{\pete}[1]{}
\newcommand{\hooman}[1]{}
\newcommand{\markchange}[1]{#1}
\fi

\newcommand{\resultimgwidth}{0.08}
\newcommand{\resultimgwidthvoxelours}{0.15}
\newcommand{\resultsubfigwidth}{0.06\linewidth}

\newif\ifincappendix

\incappendixtrue

\ifincappendix
\newcommand{\sectionappendixoperationdecoder}{\ref{section:appendix_operation_decoder}}   
\newcommand{\sectionappendixencodertwod}{\ref{section:appendix_encoder2d}}
\newcommand{\sectionappendixdecodertwod}{\ref{section:appendix_decoder2d}}
\newcommand{\sectionappendixenvelopedecoder}{\ref{section:appendix_envelope_decoder}}
\newcommand{\sectionappendixenvelopearrays}{\ref{section:appendix_envelope_arrays}}
\newcommand{\sectionappendixdownsampler}{\ref{section:appendix_downsampler}}
\newcommand{\sectionappendixtraining}{\ref{section:training}}
\newcommand{\sectionappendixdecoderheads}{\ref{section:decoderheads}}
\newcommand{\figuredecoderheads}{\ref{figure:decoderheads}}
\newcommand{\sectionappendixprofileanalysis}{\ref{section:appendix_profile_analysis}}
\newcommand{\figureabcexperiment}{\ref{figure:abc_experiment}}
\newcommand{\sectionsearchretrievalfit}{\ref{section:search_retrieval_fit}}
\newcommand{\sectionreconstructionthreed}{\ref{section:reconstruction3d}}
\else
\newcommand{\sectionappendixenvelopedecoder}{A.8} 
\newcommand{\sectionappendixenvelopearrays}{A.3} 
\newcommand{\sectionappendixoperationdecoder}{A.4} 
\newcommand{\sectionappendixencodertwod}{A.5}
\newcommand{\sectionappendixdecodertwod}{A.6}
\newcommand{\sectionappendixdownsampler}{A.7}
\newcommand{\sectionappendixtraining}{A.9}
\newcommand{\sectionappendixdecoderheads}{A.1}
\newcommand{\figuredecoderheads}{5}
\newcommand{\figureabcexperiment}{10}
\newcommand{\sectionappendixprofileanalysis}{A.2}
\newcommand{\sectionsearchretrievalfit}{4.4}
\newcommand{\sectionreconstructionthreed}{6.2}
\fi
\begin{abstract}
Reverse Engineering a CAD shape from other representations is an important geometric processing step for many downstream applications. In this work, we introduce a novel neural network architecture to solve this challenging task and approximate a smoothed signed distance function with an editable, constrained, prismatic CAD model. During training, our method reconstructs the input geometry in the voxel space by decomposing the shape into a series of 2D profile images and 1D envelope functions.  These can then be recombined in a differentiable way allowing a geometric loss function to be defined. During inference, we obtain the CAD data by first searching a database of 2D constrained sketches to find curves which approximate the profile images, then extrude them and use Boolean operations to build the final CAD model. Our method approximates the target shape more closely than other methods and outputs highly editable constrained parametric sketches which are compatible with existing CAD software.
\end{abstract}
\author{Joseph G. Lambourne}
\email{joseph.lambourne@autodesk.com}
\affiliation{%
  \institution{Autodesk Research}
  \country{United Kingdom}
}

\author{Karl D.D. Willis}
\email{karl.willis@autodesk.com}
\affiliation{%
  \institution{Autodesk Research}
  \country{USA}
}

\author{Pradeep Kumar Jayaraman}
\email{pradeep.kumar.jayaraman@autodesk.com}
\affiliation{%
  \institution{Autodesk Research}
  \country{Canada}
}

\author{Longfei Zhang}
\email{longfei.zhang@autodesk.com}
\affiliation{%
  \institution{Autodesk}
  \country{China}
}

\author{Aditya Sanghi}
\email{aditya.sanghi@autodesk.com}
\affiliation{%
  \institution{Autodesk Research}
  \country{Canada}
}
\author{Kamal Rahimi Malekshan}
\email{kamal.malekshan@autodesk.com}
\affiliation{%
  \institution{Autodesk Research}
  \country{Canada}
}
\renewcommand\shortauthors{Lambourne et al}

%
%


%
%

\begin{teaserfigure}
    \begin{center}
        \includegraphics[width=\columnwidth]{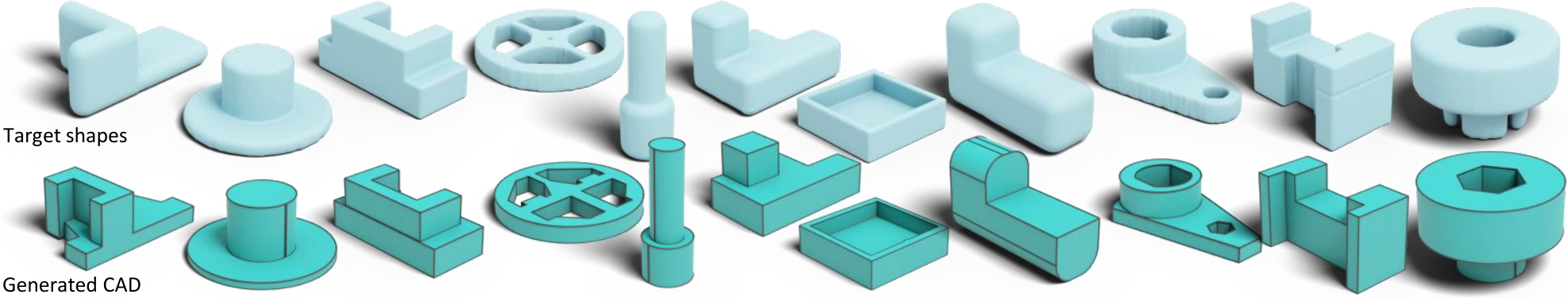}
        \caption{Conversion of rounded voxel models to prismatic CAD.}
        \label{figure:teaster}
    \end{center}
\end{teaserfigure}

\maketitle

\section{Introduction}
Reverse engineering, the creation of Computer Aided Design (CAD) models which closely match some target geometry, is one of the most sought-after geometric modeling technologies and has been extensively studied using both traditional algorithms \cite{Buonamici2018,VaradyAutomaticProcedures2008} and machine learning approaches \cite{wang2020pie,sharma2020parsenet, uy2022point2cyl}.  Traditional techniques work well in cases where the target geometry is not noisy or otherwise distorted and primitive surfaces like planes and cylinders can be correctly identified and fitted \cite{BENKO2004511,li_globFit_sigg11}.  However, many sources of geometry provide only an approximate description of the desired shape.  For example, noise in point clouds acquired by low cost scanners is often removed by smoothing the shape during surface reconstruction \cite{Calakli2011} and shape manipulation approaches working on levelsets also result in rounding of the geometry \cite{sethian1999level}.  As there are many effective methods to convert meshes and dense point clouds into signed distance functions stored in voxel grids \cite{sanchez2012efficient, baerentzen2005robust}, the ability to convert voxel models to CAD can be used as a stepping stone to reverse engineer these representations as well.

While free-form CAD data can readily be derived from rounded models by fitting splines to a quadrangulation of the surface \cite{Bommes2009}, the design of mechanical parts requires prismatic geometry composed of primitive surface types \cite{BENKO2001839}.  Planar faces are important for the interfaces where components meet and cylindrical surfaces are required for holes and as parts of rotating mechanisms.  In addition, shapes created from extrusions of connected lines and arcs are cheap to manufacture with low cost 2.5-axis Computer Numerical Control (CNC) machines.  Most modern CAD software records these curves, along with the lengths and directions of extrusions,  in a parametric recipe which designers can use to edit the model.  The curve geometry is held in 2D sketches along with constraints which maintain important aspects to design intent such as equality, symmetry, and perpendicularity.  The designer can also define parameters which control distances between curves that can be varied to modify the geometry.   By editing these parameters and replaying the recipe the shape of the final CAD model can be controlled.

Recently a number of methods have been proposed for the generation of CAD data as a sequence of geometry creation operations using transformer models \cite{willis2021engineering,Ganin2021ComputerAidedDA,para2021sketchgen,SeffVitruvion2021,wu2021deepcad,jayaraman2022solidgen}.  These are all trained utilizing supervision from ground truth sequence data.  As they do not incorporate a loss function which directly compares the generated geometry and the target shape, their ability to match target geometry is limited.   

In this work we propose a differentiable pipeline which can reconstruct a target shape in terms of voxels and extract the parametric recipe simultaneously.  The voxel model is created from a sequence of decoded profile images which are then extruded along the axes of the voxel grid and combined using Boolean operations.  A geometric loss can then be defined which brings this voxel model closer to the target data.   The recipe can then be extracted by matching the profile images with constrained parametric sketches which can be further fitted to better approximate the shape while maintaining design intent.  We show that the CAD data reconstructed from the recipe contains fewer invalid solids and provides a better approximation to the target than DeepCAD \cite{wu2021deepcad}, while having the additional benefit that the constraints and parameters in the sketches are available to designers for further editing.  

\section{Related work}
We now review work from the literature related to the generation of editable CAD models. 

\paragraph{Constructive Solid Geometry}
A common approach to producing editable CAD models is to use a Constructive Solid Geometry (CSG) representation. With CSG, 3D shapes can be expressed as a tree of parametric primitives (e.g. cuboids, spheres, and cones) that are individually positioned and combined using Boolean operations (e.g. union, subtraction, intersection). This lightweight representation enables simple edits by adjusting the 
parameters of the primitives  and their affine parameters, while also allowing  more interesting applications with program synthesis~\cite{du2018inversecsg,nandi2018functional,nandi2020synthesizing, ellis2019write, tian2018learning}. In the context of deep learning, this representation has been explored in several works such as~\cite{sharma2018csgnet, deng2020cvxnet, chen2020bsp, kania2020ucsg, yu2022capri}.   
However, the primitives used in CSG are not as flexible as the extrusions of parametric sketches used by modern mechanical design tools.  Consequently CSG shapes tend to be made from large number of primitives which are hard for users to edit to control the final geometry.

\paragraph{Engineering Sketch Generation}
With the release of a large scale engineering sketch dataset~\cite{Ari2020}, a number of learning-based approaches to engineering sketch generation have been explored~\cite{willis2021engineering,Ganin2021ComputerAidedDA,para2021sketchgen,SeffVitruvion2021}. These works successfully frame engineering sketch generation as a sequence prediction task, where geometric primitives are sequentially predicted using a Transformer~\cite{vaswani2017attention} model. 
Not addressed in these works is the generation of 3D parts compatible with parametric CAD.

\paragraph{\markchange{CAD} Generation}
Numerous approaches for 3D shape generation have been proposed~\cite{chaudhuri2020learning}. \markchange{Related to our work are approaches that predict a sequence of CAD modeling operations from which a solid CAD model can be built~\cite{willis2020fusion,Xu2021InferringCM}. In particular, DeepCAD~\cite{wu2021deepcad} is a generative model that learns from sequences of CAD modeling operations to produce editable CAD designs. In contrast to our approach, DeepCAD does not generate sketch constraints---a key factor aiding editability.
CAD generation can also be achieved by predicting geometric entities and topological connections to form 3D shapes. PolyGen~\cite{nash2020polygen} introduced the idea of separating the sequential prediction of geometric points, from the topological connections that form n-gon meshes using pointer networks~\cite{vinyals2015pointer}. SolidGen~\cite{jayaraman2022solidgen} applied a similar approach to the more difficult task of generating 3D shapes in the boundary representation (B-Rep) format commonly used in mechanical CAD. Missing from these sequence-based 3D generation approaches is a geometric loss suitable for reconstruction tasks.}


\paragraph{CAD Reconstruction}
The goal of CAD reconstruction is to \textit{reverse engineer} a 3D shape, \markchange{typically as a series of parametric primitives}, given approximate input, such as a point cloud or freehand sketch. \citet{smirnov2021patches} recover manifold 3D shapes from freehand sketch input by deforming Coons patch-based templates from set object categories.
\markchange{Given point cloud input, traditional approaches segment and then fit parametric primitives, such as planes, spheres, and cylinders, to the underlying point cloud~\cite{Schnabel2007EfficientRF}. Recent progress with learning-based approaches has addressed primitive segmentation~\cite{Yan2021HPNetDP} reconstruction of parametric curves~\cite{wang2020pie} and surfaces~\cite{sharma2020parsenet, Li2019SupervisedFO, Guo2022ComplexGen}. Missing from these works is the sequence of CAD modeling operations that greatly enhances editability.} Concurrent to our work, Point2Cyl~\cite{uy2022point2cyl} reconstructs a  collection of extrusions which can be manually ordered and combined to build CAD shapes. \markchange{An advantage of our approach is allowing for} rounded or otherwise distorted shapes to be approximated by CAD data.  As noisy or incomplete input data is often treated by smoothing, the ability to rebuild sharp prismatic shapes which approximate rounded geometry allows for a wider range of input.

\paragraph{Search and retrieval}
An alternative approach to reconstruction is to search and retrieve an appropriate shape from an existing database. Parametric CAD models are well suited to retrieval because a range of different shapes can be returned from different input parameters. \citet{Schulz2017} address retrieval for parametric shapes in a collection by approximating their manifolds in the descriptor space with a set of primitives. Recent learning-based approaches actively deform the retrieved shape to more accurately match the input query~\cite{uy2020deformation, uy2021joint}. In our work we utilize a search, retrieval and fitting procedure to identify parametric sketches with geometry close to a 2D target shape.  The parameters in the retrieved sketches can then be fitted to better approximate the shape of the target while maintaining the design intent encoded by the sketch constraints.   This allows for novel workflows like interpolation between sketches providing smooth parametric variation in some regions and topological change in others.   

\section{Data preparation}
\label{section:data_preparation}
A number of CAD related datasets have been released recently.  Notably the ABC dataset \cite{koch2019abc} which provided 1,159,257 unique 3D CAD models, the SketchGraphs dataset \cite{Ari2020} which contains 15 million constrained parametric 2D sketches, the Fusion Gallery Reconstruction dataset \cite{willis2020fusion} which provided 8,625 full sketch and extrude construction sequences along with sketch constraint information, and the DeepCAD dataset \cite{wu2021deepcad} which provided 127,267 unique sketch and extrude construction sequences without sketch constraints.   In this work we utilize 3D  data from DeepCAD and ABC, along with constrained sketches from SketchGraphs.  

\markchange{As the ABC dataset contains solids which require sophisticated free-form surface construction techniques like sweeps and lofts, we identify a subset of 52,503 models for which more than 95\% of the surface area was consistent with extrusions aligned with the x, y or z axes.  These are used as target geometry for some experiments and in the computation of the Fr\'echet inception distance (FID) metric. }

\subsection{Common extrusion combinations}
\label{section:extrusion_sequences}
Rather than attempting to produce arbitrary CAD modeling sequences, our approach focuses on the generations of B-Rep models \markchange{which are built from pre-defined extrusion recipes, containing the axes along which profiles will be extruded, the order in which the extrusions are applied and whether each extrusion should be added or subtracted from the previous geometry.} Most CAD software allows the designer to create sketches on existing planar faces of the solid being modeled, leading to some frequent patterns in the way extrusions are aligned and stacked.  In this section we analyze the combinations of extrusions which  designers employed when creating the models in the DeepCAD dataset \cite{wu2021deepcad} \markchange{and find the most frequent patterns which are used as the basis for our extrusion recipes.}
After the deduplication procedure used in \cite{jayaraman2022solidgen}, the dataset contains 127,267 files.  Of these, just 358 contain two or more extrusions which are not orthogonal to one another, accounting for just 0.28\% of the dataset.  For this reason we focus only on the generation of solids built entirely from orthogonal extrusions.  Without loss of generality, it is always possible to rotate orthogonal extrusion models such that the first extrusion will be created in the z direction.  If a second extrusion direction exists, the model can then be rotated around the z axis so the x direction is aligned with the second extrusion direction.   Transforming the models into this canonical pose allows 102,287 files (80.37\%) of the dataset to be represented by the 18 combinations of extrusions shown in \markchange{Appendix \sectionappendixdecoderheads~ Figure \figuredecoderheads.}  We see that single extrusions form the majority (56.7\%) of the dataset.  The next most frequent extrusion combination is a boss extruded from a sketch on the top face of the previous extrusion (see Figure \figuredecoderheads b), which accounts for 5.24\% of the models, followed by a cut which starts from the top face of the previous extrusion (see Figure \figuredecoderheads c) which accounts for 4.47\% of the dataset. 

\subsection{Rounded shapes}
Many processes which produce signed distance function models apply some smoothing to the shape as a kind of regularization.
To allow our method to work well with these rounded shapes, we augment the dataset as follows.  For each model we can create a signed distance voxel representation of the geometry in the canonical orientation described in Section \ref{section:extrusion_sequences} \markchange{and scaled into the unit cube.}  We then generate rounded augmentations of each model by adding a constant to the signed distance function to offset the shape inward,  re-initializing the distance function at the zero levelset \cite{sethian1999level} and reversing the process to offset outwards by the same amount.  For each model in the dataset we add rounded augmentations with radii of 2.5\%, 5.8\%, 9.1\%, 12.5\% the size of the length of the bounding cube, \markchange{corresponding to 1.6, 3.7, 5.8 and 8 voxels.}

\subsection{2D parametric sketch data}
\label{section:parametric_sketches}
Constrained parametric sketches were obtained from the SketchGraphs dataset.  As SketchGraphs contains collections of overlapping curves rather than closed profiles, the sketches were first processed to remove self-intersections.  All intersecting curves were converted to construction geometry until the remaining geometry either formed closed loops or became disjoint.  Disjoint sketches were discarded.  An auto-constraining algorithm was then used to add as many constraints and driving parameters as possible without making the sketches over-constrained. This procedure was applied to 1,081,217 sketches from SketchGraphs and yielded 177,609 sketches which formed close loops and were suitable for downstream CAD operations.  \markchange{We then conducted an analysis of the sequences of lines and arcs present in profile loops in the DeepCAD dataset.  The profile loops were clustered into groups which shared important geometric and topological properties as described in Appendix \sectionappendixprofileanalysis.  This analysis showed that 88\% of profile loops in the DeepCAD dataset fall within the 50 largest groups. We then selected a small set of 49 constrained sketches which preserved the most common profile loop shapes over a wide range of parametric variations.  Experiments showed that this small set of profile loops is able to approximate DeepCAD profiles surprisingly well, with an IoU of 93$\pm$11\% as described in Section \ref{section:retrieval2d}.  Experiments with more complex sets of profiles showed these sometimes resulted in extraneous details in the generated solids.  The parameters of the selected sketches were then varied to give rise to 1690 shapes (more detail in Appendix \sectionappendixprofileanalysis).} The closed loops of 2D curves in the sketches were converted to SVG and rendered as binary images with resolution of 128x128.   The fast marching method was used to convert the data to signed distance functions \markchange{with negative values inside the object.}  

\section{Method}
\begin{figure}
    \begin{center}
        \includegraphics[width=\columnwidth]{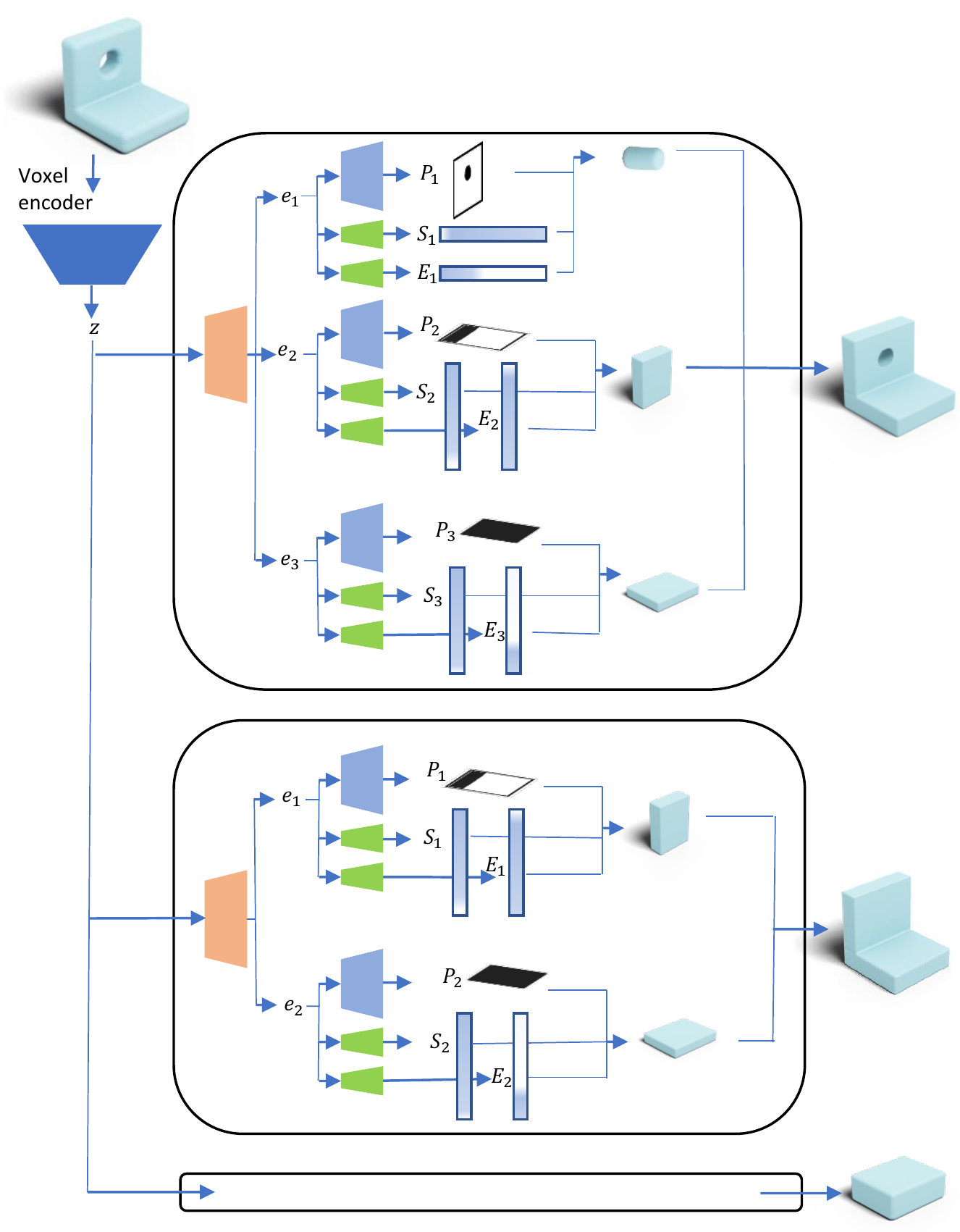}
        \caption{The differentiable extrusion decoder architecture. The input voxel model is encoded to form an embedding z.  A series of decoder modules then rebuild the shape using  hard coded construction sequences.  During training, only the ground truth decoder module is used, while at interference time the output of all decoder modules are evaluated and the one which best approximates the target shape is selected.}
        \label{figure:architecture_3d}
    \end{center}
\end{figure}
Our method for reconstructing CAD shapes utilizes an autoencoder which encodes voxel models and decodes a prismatic shape as a collection of differentiable extrusions.  The differentiable extrusions are created from 2D profile images representing the shapes to be extruded, along with 1D envelope arrays which define the start and end positions of each extrusion.  At inference time, the 2D images can be processed by a search, retrieval and fitting procedure, which yields parametric sketches suitable for extrusion by a CAD kernel, creating the final CAD model.
\subsection{Differentiable extrusions}
An overview of the architecture for decoding the differentiable extrusion voxel models is shown in Figure \ref{figure:architecture_3d}.  The input is a signed distance function, \markchange{represented as a voxel grid with resolution 64x64x64}.  A standard voxel encoder as described in \cite{OccupancyNetworks2019} can be used to convert this into a 128 length embedding vector $z$.  This is passed to a series of decoder modules, each of which will attempt to interpret the shape in terms a predefined extrusion sequence recipe as described in Section \ref{section:extrusion_sequences}.

For a decoder module with a sequence of $n$ extrusions, the voxel embedding, $z$, is split into extrusion embeddings $e_0, e_1, ..., e_{n-1}$.   This is done by a series of linear layers separated by ReLU non-linearities (see Appendix \sectionappendixoperationdecoder).   The extrusion embeddings can then be passed to a shared 2D deconvolution image decoder \markchange{(see Appendix \sectionappendixdecodertwod) which will generate images with resolution 128x128, representing each profile, $P_0, P_1, ..., P_{n-1}$.  The decoded pixel values will be negative for pixels inside the profile and positive outside it.} 

In addition the locations of the start and end of the extrusions needs to be found.  To achieve this we predict 1D arrays of envelope values which can be used to limit the extrusions in a differentiable way.  If the $i$th extrusion does not share a common start or end plane with others in the model, we decode two \markchange{arrays of length 64} to represent its limits.  The start array, $S_i$ will contain positive values below the extrusions start plane and negative values above it, while the end array, $E_i$ will be negative below the end plane and positive above it.  \markchange{These arrays are decoded from the extrusion embeddings using a 1D deconvolution network as described in Appendix \sectionappendixenvelopedecoder.   For some frequent cases where two extrusions share a common plane, a single start/end array can decoded and shared by both of them.  This is explained in more detail in Appendix \sectionappendixenvelopearrays.      Taking the max of $S_i$ and $E_i$ defines an envelope which is negative only for the parts of the $i$th extrusion which we want to appear in the model.}   

With this information we can construct a 64x64x64 voxel grid representation of the model in a differentiable way.  For each of the $n$ extrusions, the profile $P_i$ \markchange{is first downsampled as described in Appendix \sectionappendixdownsampler,} and then duplicated along the appropriate dimension of a 3D tensor as defined by the hard coded recipe in the decoder module.  The envelope is then applied by taking the max of each slice of the tensor with its corresponding value in the envelope array.  Finally the extrusions are combined with the Boolean operations in the recipe, using the min function for union, max for intersection and multiplying by -1 to form the complement.  \markchange{This gives rise to a signed function, $\phi$ of logits which, if passed through a sigmoid, would be the probability of a voxel being \emph{outside} the model.  Adopting this convention makes $\phi$ negative inside the object, allowing the shape to be visualized directly from the logits using marching cubes.  Note that $\phi$ is not a distance function as the Eikonal equation is not satisfied as we move away from the zero level set. }

\subsection{Loss functions}
The model is trained using two loss functions, a supervised loss which requires knowledge of the CAD construction sequence, and an unsupervised loss which does not require any construction sequence information.  \markchange{Both the supervised and unsupervised contributions to the loss are computed using a binary cross entropy "with logits" loss which includes the sigmoid function allowing it to operate directly on the predicted logits
\begin{equation}
    BCE(x, y) = -\text{mean} \left[ y\ln(\sigma(x)) + (1-y)\ln(1-\sigma(x))  \right]
\end{equation}
where $x$ is the tensor of predicted logits, $y$ is the binary target value, $\sigma()$ is a sigmoid function and the mean is over all the elements of the tensor. For the unsupervised loss function, the target voxel values $\hat{T}$ are derived from the CAD data before any rounded augmentation is applied.  $\hat{T}$ is 1 for voxels outside the object and 0 inside the object.   The unsupervised loss for the differentiable voxel model is then 
\begin{equation}
L_{vox} = BCE(\phi, \hat{T})
\end{equation}
The supervised loss is computed using the ground truth 2D binary profile images, $\hat{P_i}$ and 1D binary arrays, $\hat{S_i}$ and $\hat{E_i}$, corresponding to the start and end envelope arrays.   Following the convention above, these tensors have values of 1 outside the object and 0 inside.  We compute a binary cross entropy loss, averaged over the $n$ extrusions utilized by the decoder module as 
\begin{align}
    L_{profile} = \frac{1}{n}\sum_i BCE(P_i, \hat{P_i}) \label{equation:profile_loss}\\
    L_{start} = \frac{1}{n}\sum_i BCE(S_i, \hat{S_i}) \\
    L_{end} = \frac{1}{n}\sum_i BCE(E_i, \hat{E_i})
\end{align}
The final loss is then computed as
\begin{equation}
    L = L_{vox} + L_{profile} + L_{start} + L_{end} 
    \label{equation:total_loss}
\end{equation}
}

\subsection{Decoder module selection}
During supervised training, the ground truth decoder module for each training example is known.  At inference time this information is not available, so in order to make this selection we compare the input voxel model with the decoded differentiable extrusion model from each decoder module.   The unsupervised binary cross entropy loss function, $L_{vox}$, can be evaluated for each of the available decoder modules and the module which results in the smallest loss is utilized.
This procedure is also extended to allow regeneration of CAD models where the first extrusion is not in the z direction.   The input voxel grid can be rotated 90 degrees around the x and y axes before the data is passed to the network.  The differentiable extrusion model is then created for each decoder module with the target shape encoded in all three orientations.  Then the best orientation and decoder module combination can be selected.  Finally the regenerated CAD model is transformed back into its original pose.

\begingroup
\renewcommand{\arraystretch}{0.0}
\newcommand{\resultimgtwodwidth}{0.29}
\begin{figure}
  \begin{center}
    \begin{tabular}{ c c c }
    \includegraphics[width=\resultimgtwodwidth\linewidth]{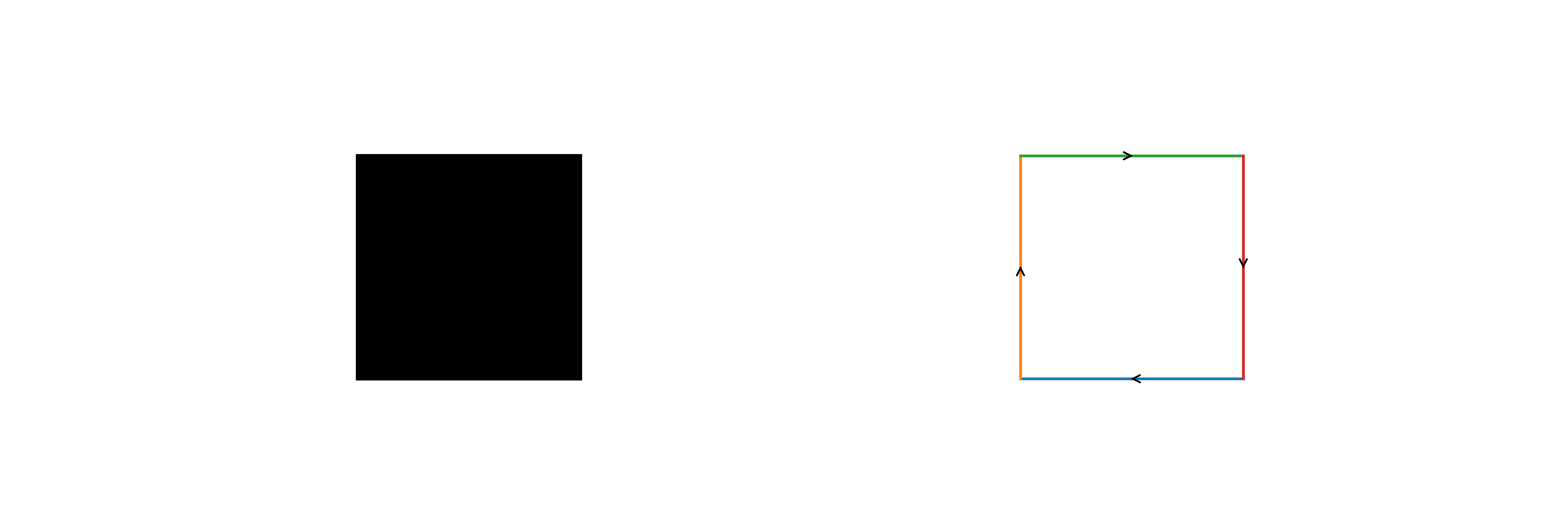} & 
    \includegraphics[width=\resultimgtwodwidth\linewidth]{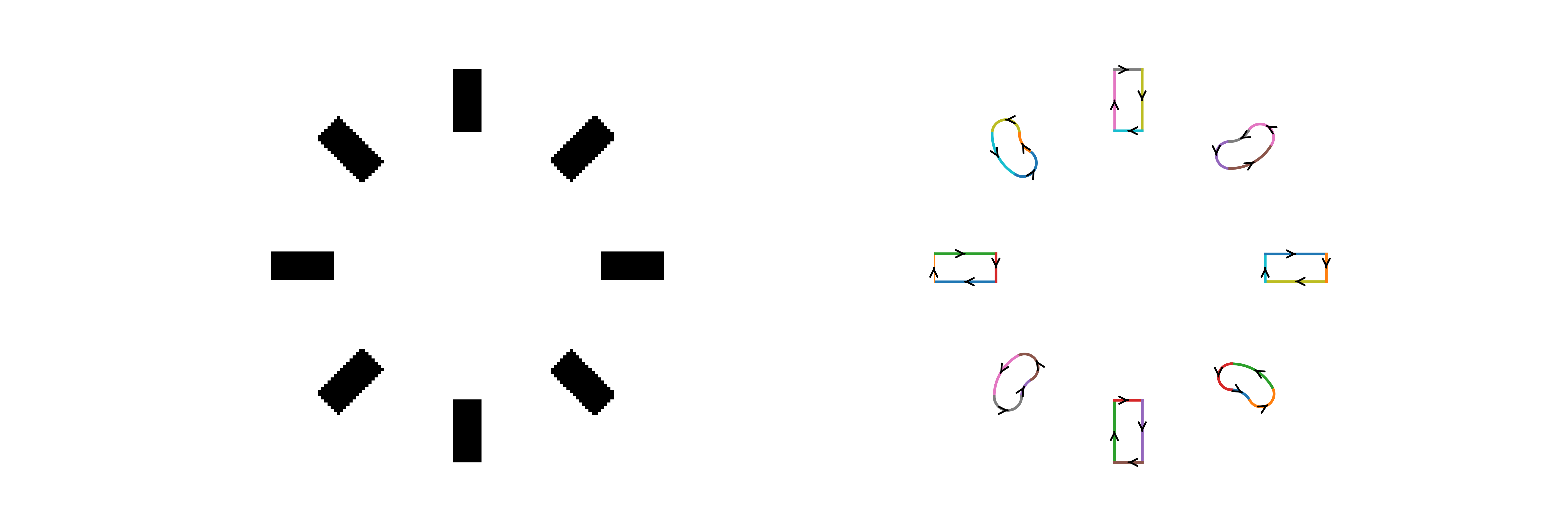} & 
    \includegraphics[width=\resultimgtwodwidth\linewidth]{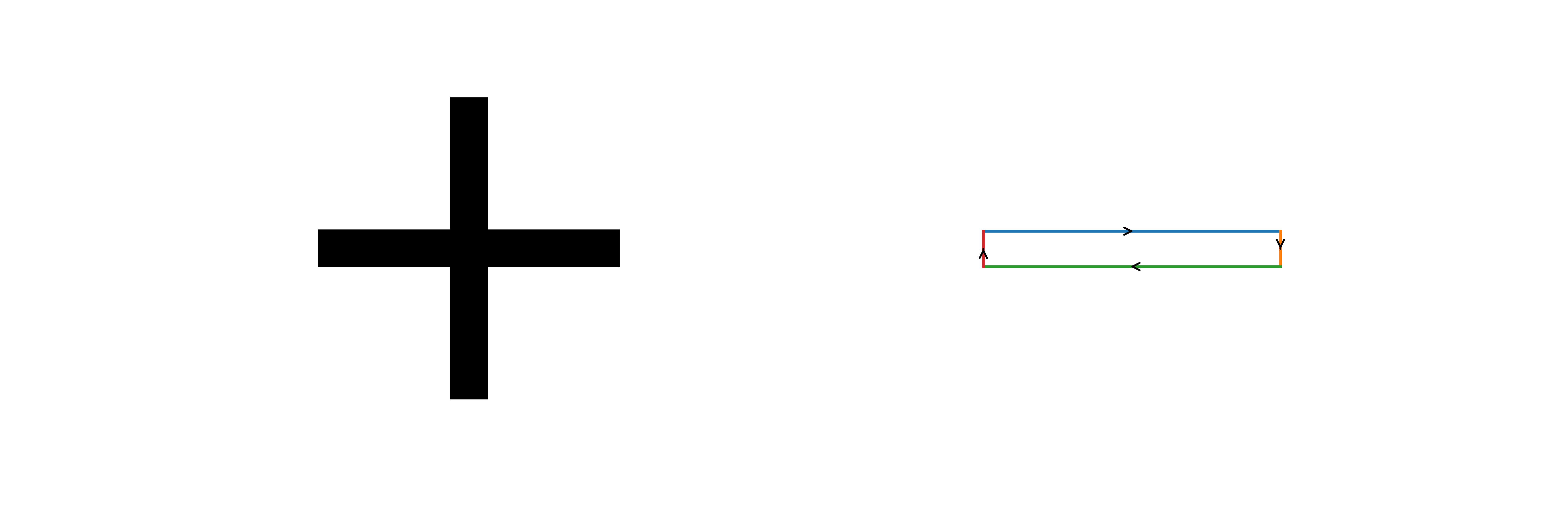} \\ 
    \includegraphics[width=\resultimgtwodwidth\linewidth]{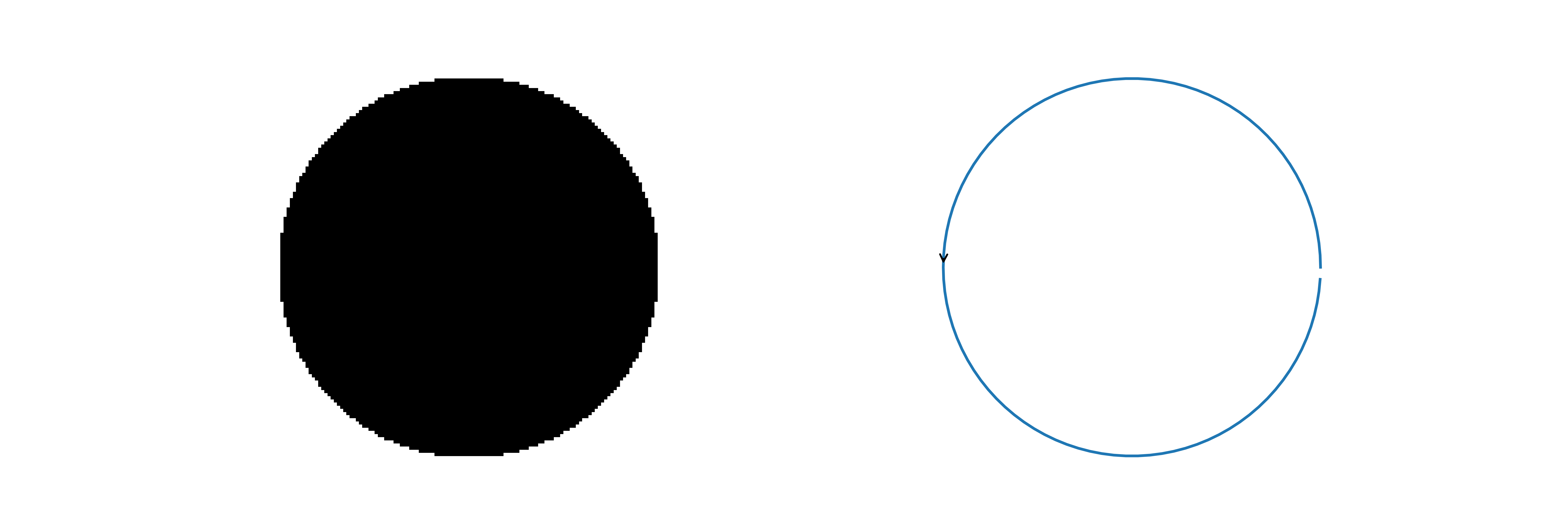} & 
    \includegraphics[width=\resultimgtwodwidth\linewidth]{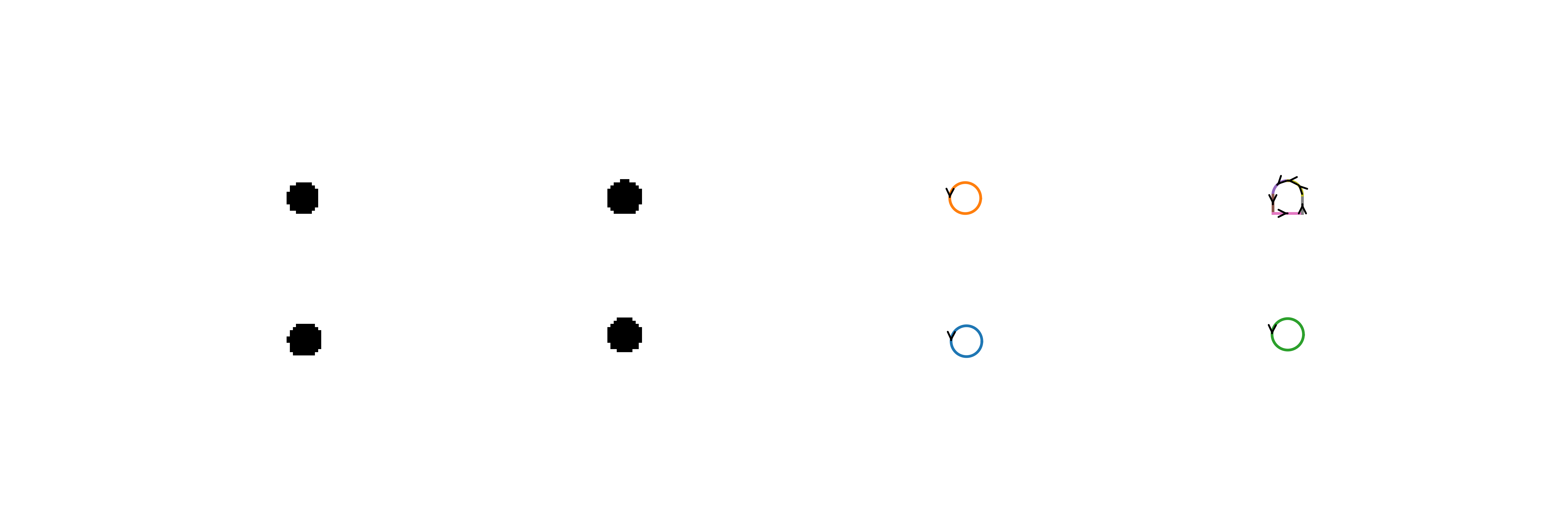} & 
    \includegraphics[width=\resultimgtwodwidth\linewidth]{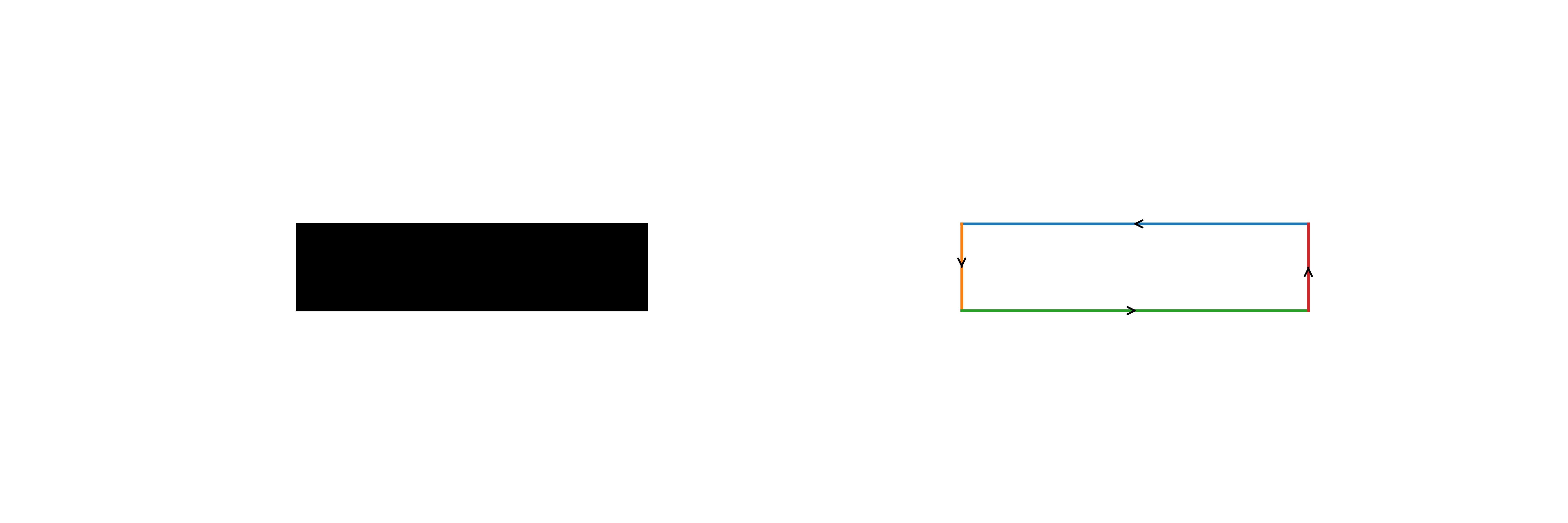} \\
    \includegraphics[width=\resultimgtwodwidth\linewidth]{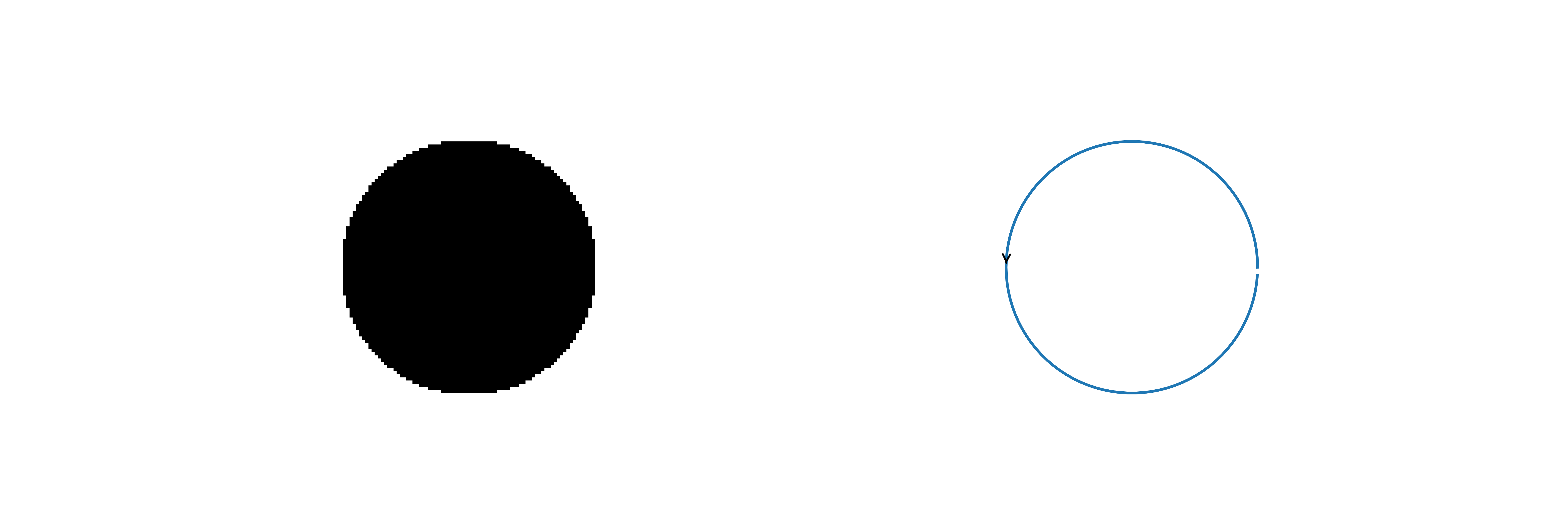} & 
    \includegraphics[width=\resultimgtwodwidth\linewidth]{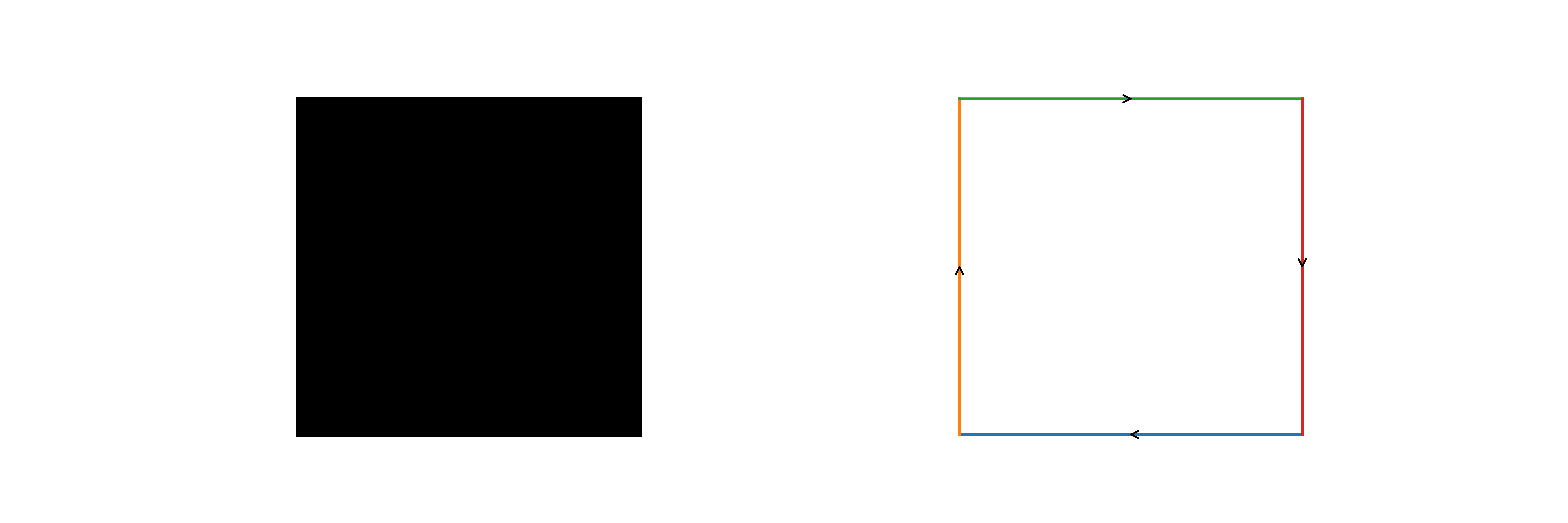} & 
    \includegraphics[width=\resultimgtwodwidth\linewidth]{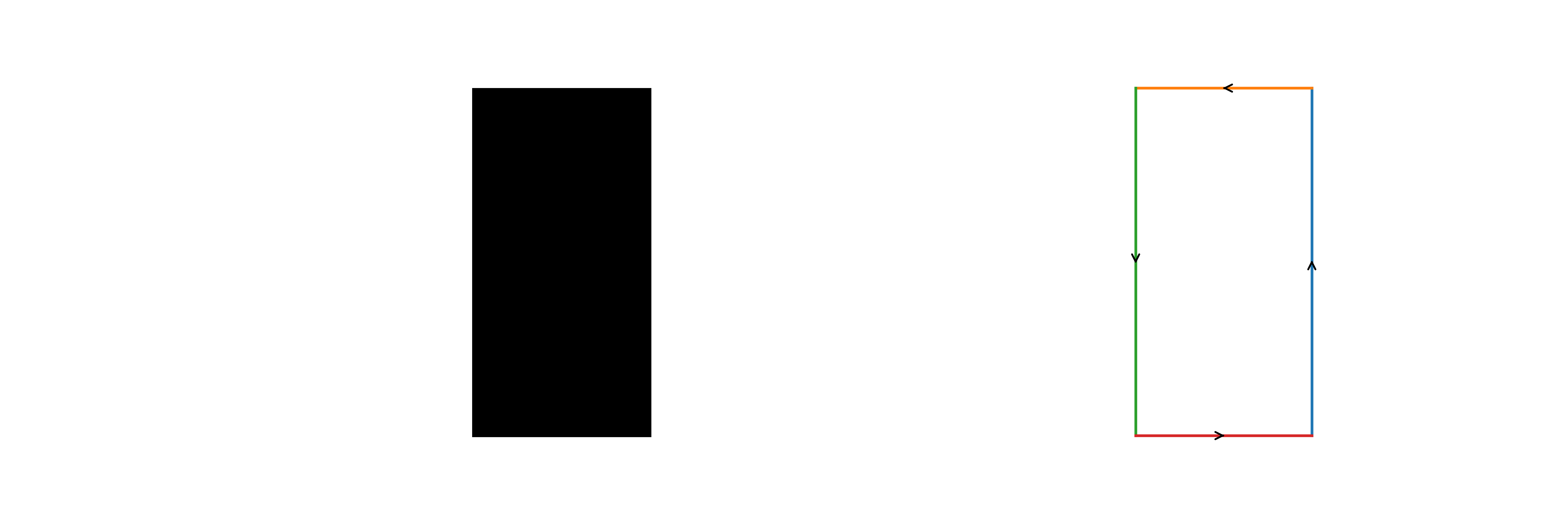} \\
    \includegraphics[width=\resultimgtwodwidth\linewidth]{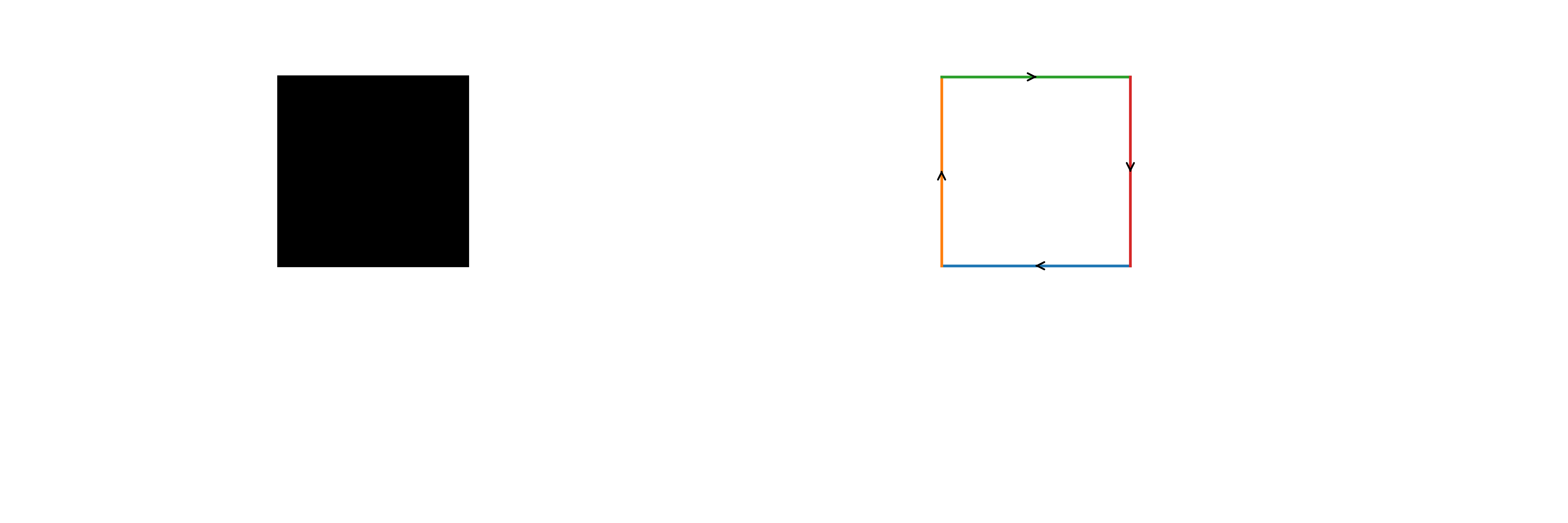} & 
    \includegraphics[width=\resultimgtwodwidth\linewidth]{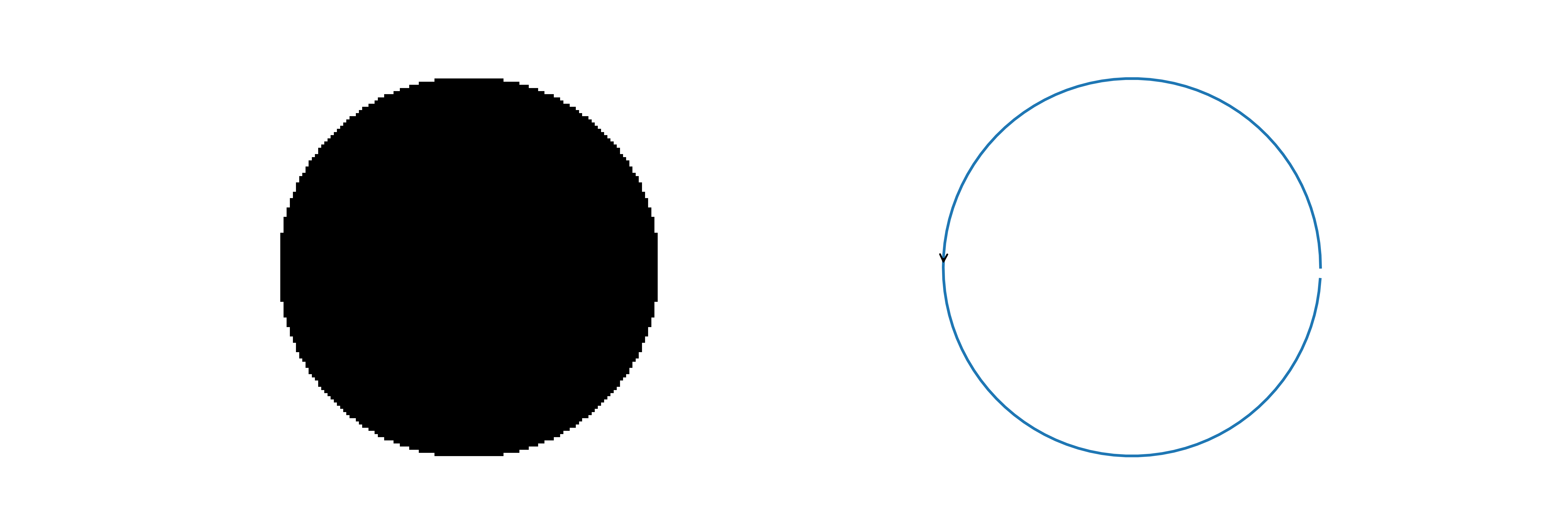} & 
    \includegraphics[width=\resultimgtwodwidth\linewidth]{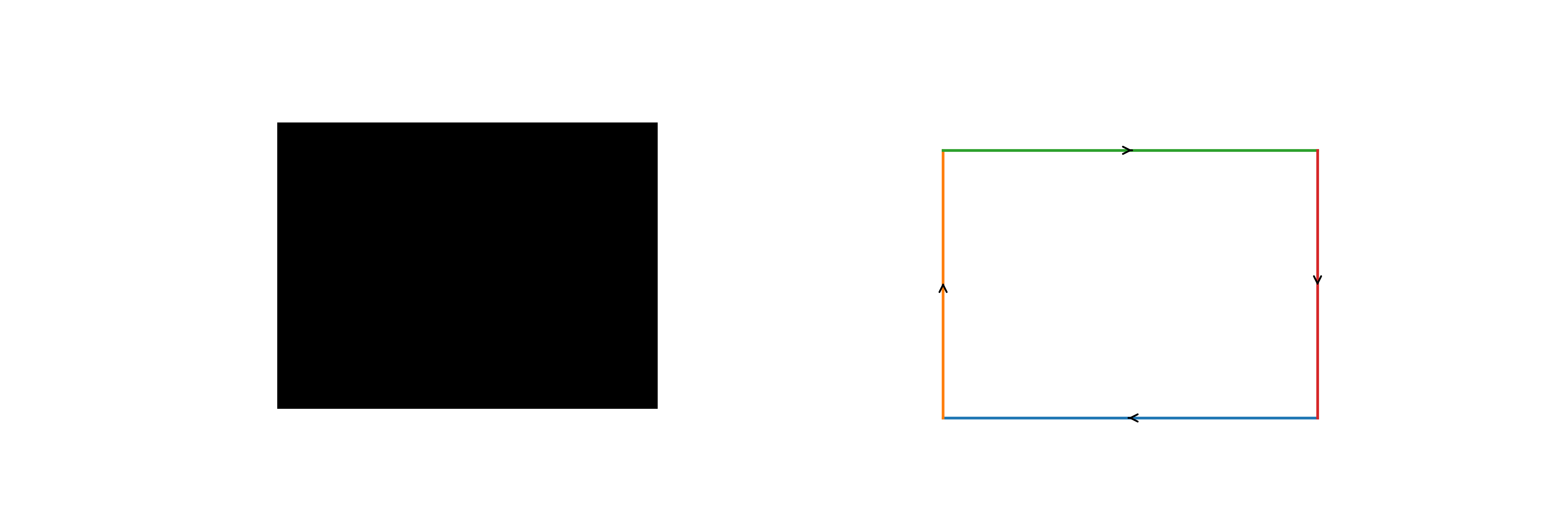} \\
    \includegraphics[width=\resultimgtwodwidth\linewidth]{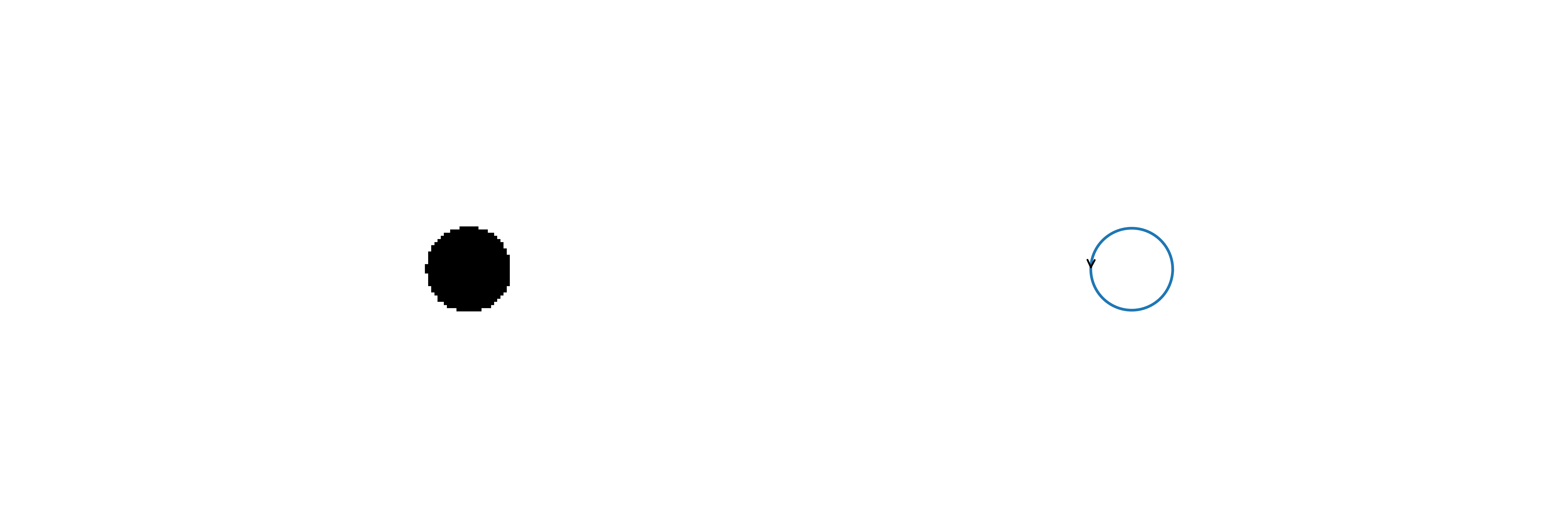} & 
    \includegraphics[width=\resultimgtwodwidth\linewidth]{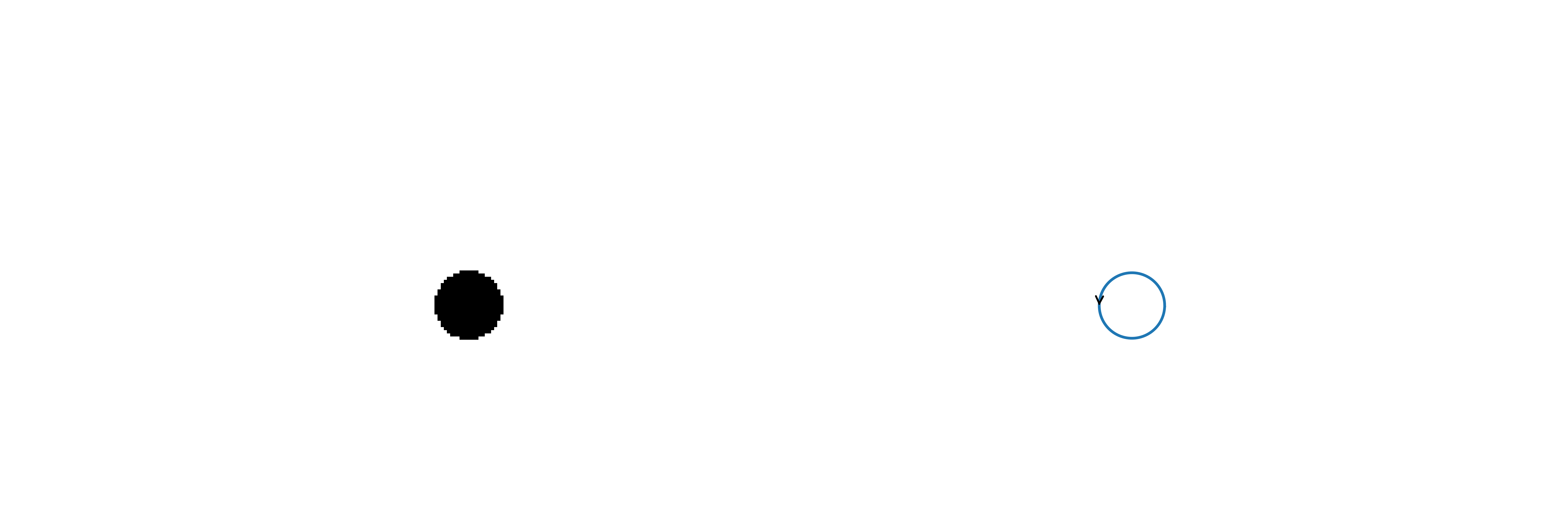} & 
    \includegraphics[width=\resultimgtwodwidth\linewidth]{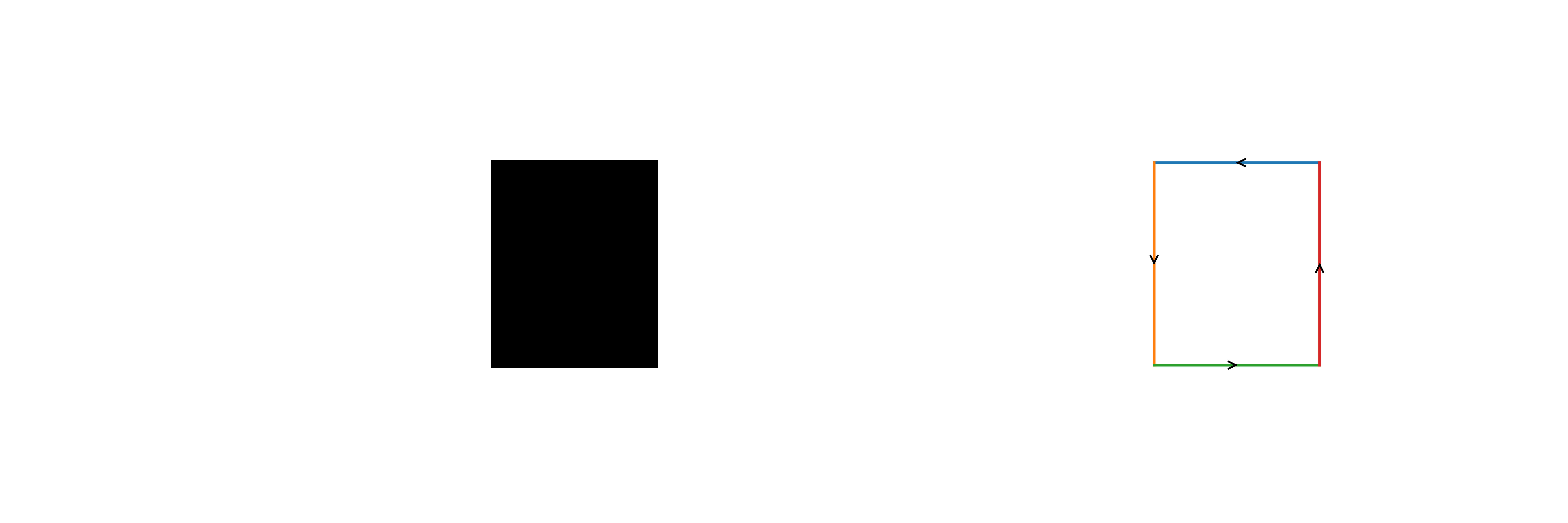} \\
    \includegraphics[width=\resultimgtwodwidth\linewidth]{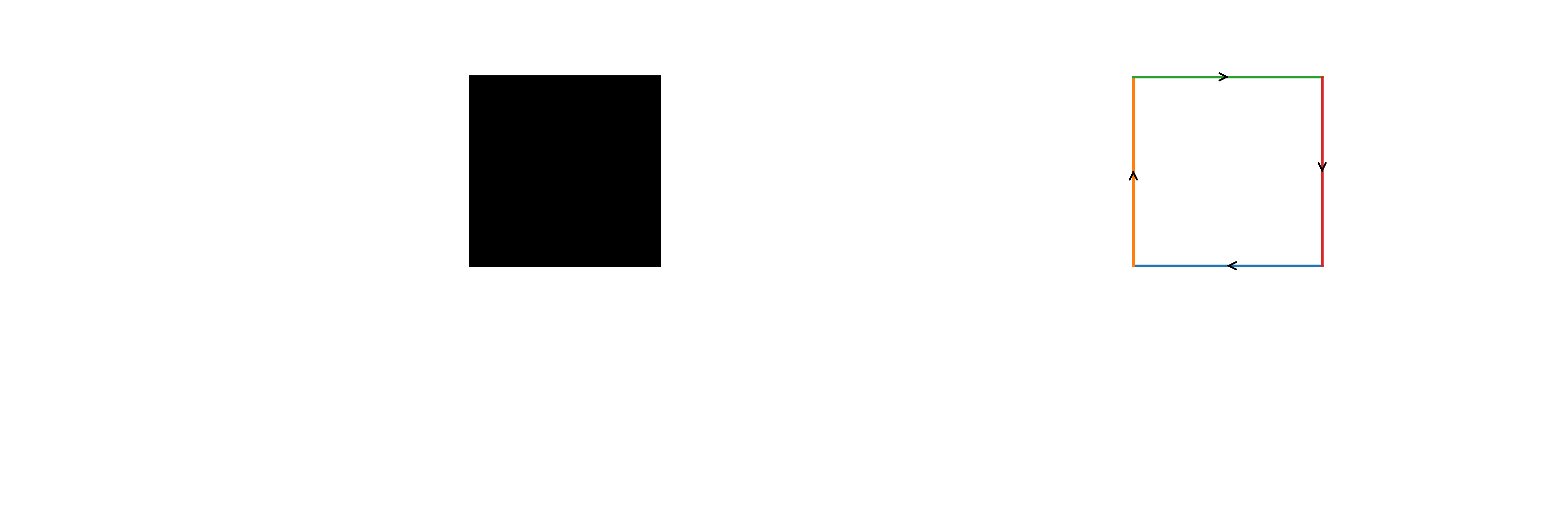} & 
    \includegraphics[width=\resultimgtwodwidth\linewidth]{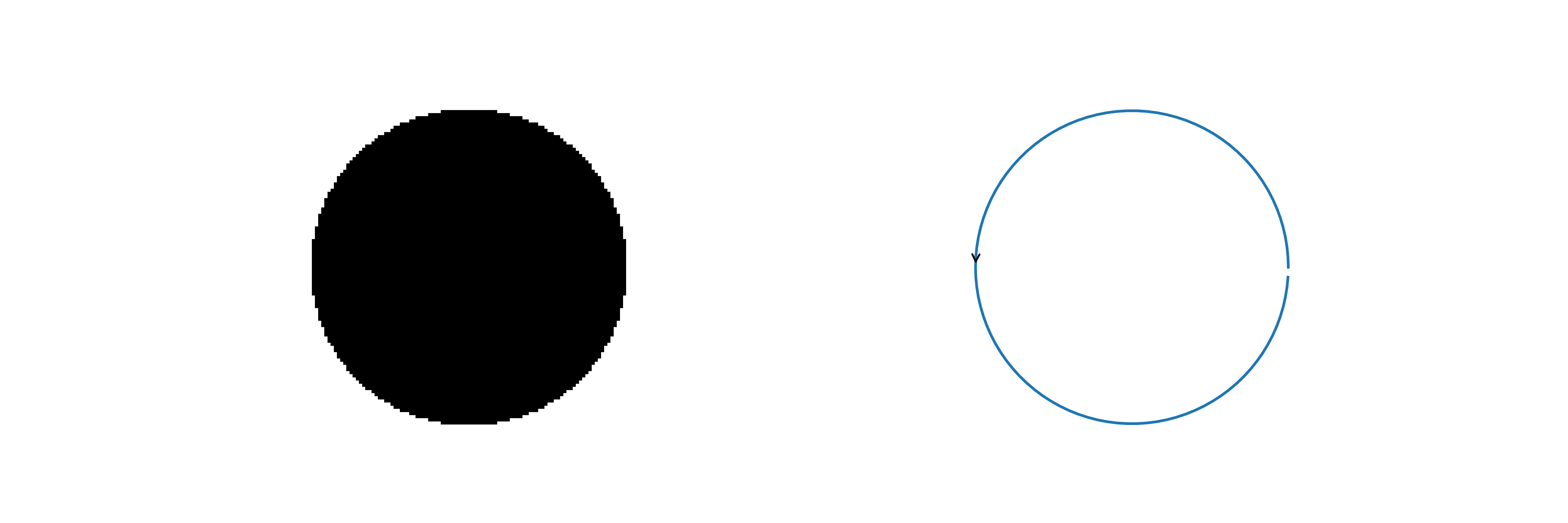} & 
    \includegraphics[width=\resultimgtwodwidth\linewidth]{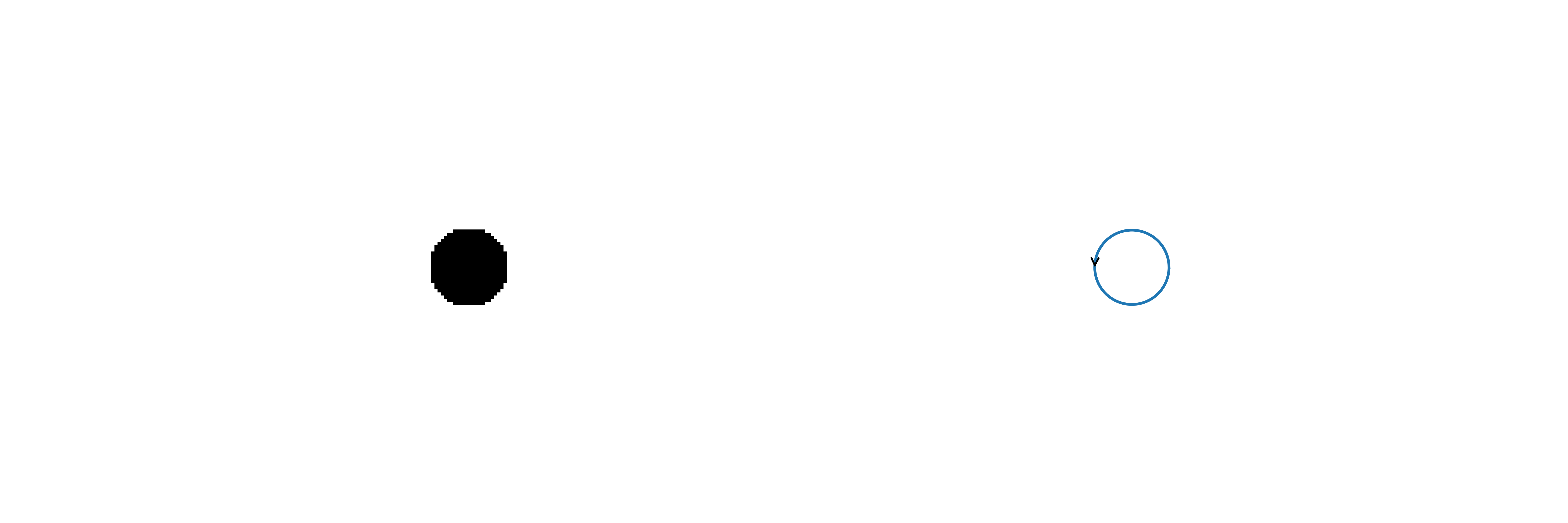} \\
    \includegraphics[width=\resultimgtwodwidth\linewidth]{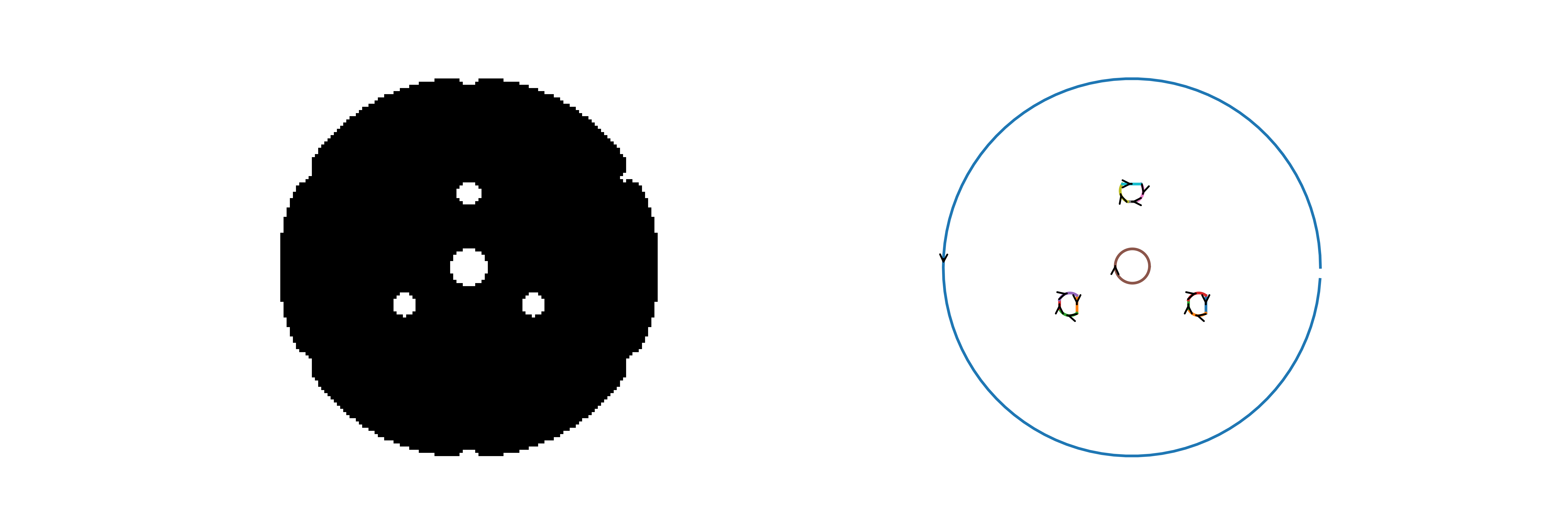} & 
    \includegraphics[width=\resultimgtwodwidth\linewidth]{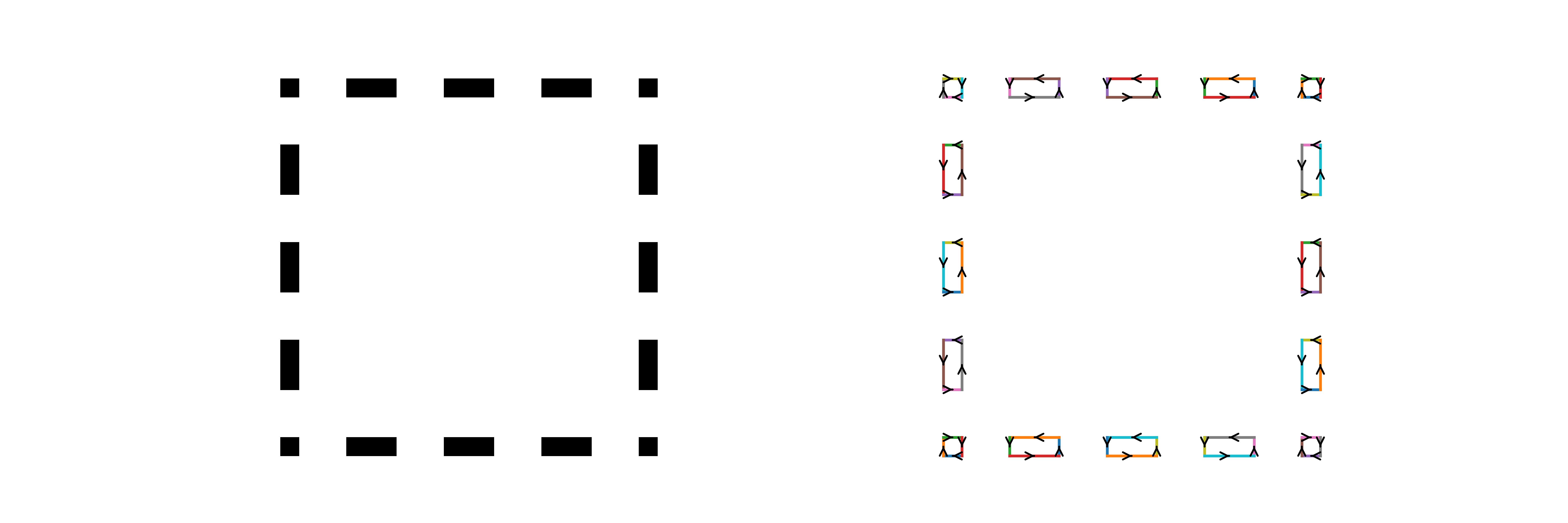} & 
    \includegraphics[width=\resultimgtwodwidth\linewidth]{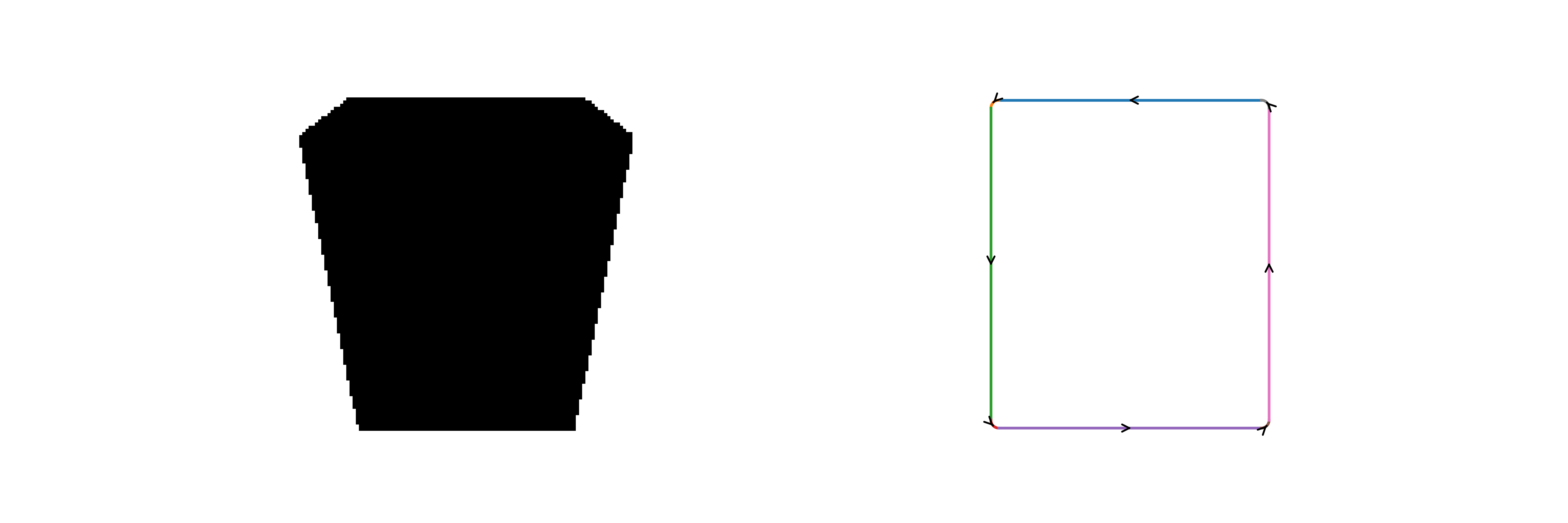} \\ 
    \includegraphics[width=\resultimgtwodwidth\linewidth]{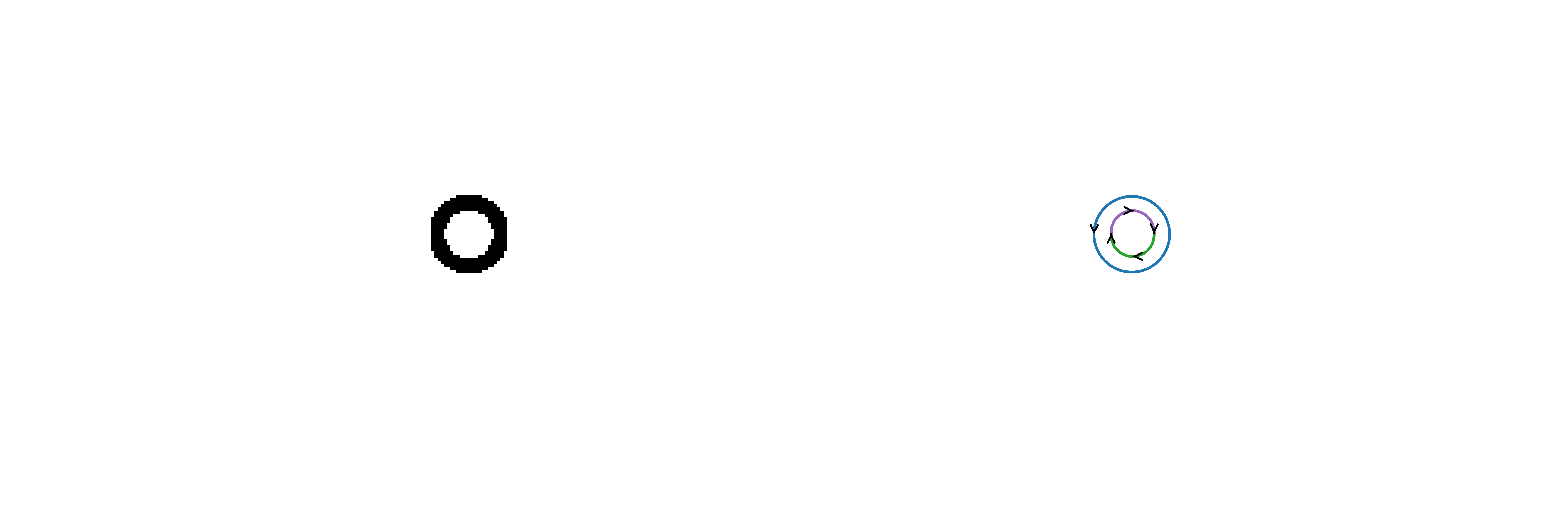} & 
    \includegraphics[width=\resultimgtwodwidth\linewidth]{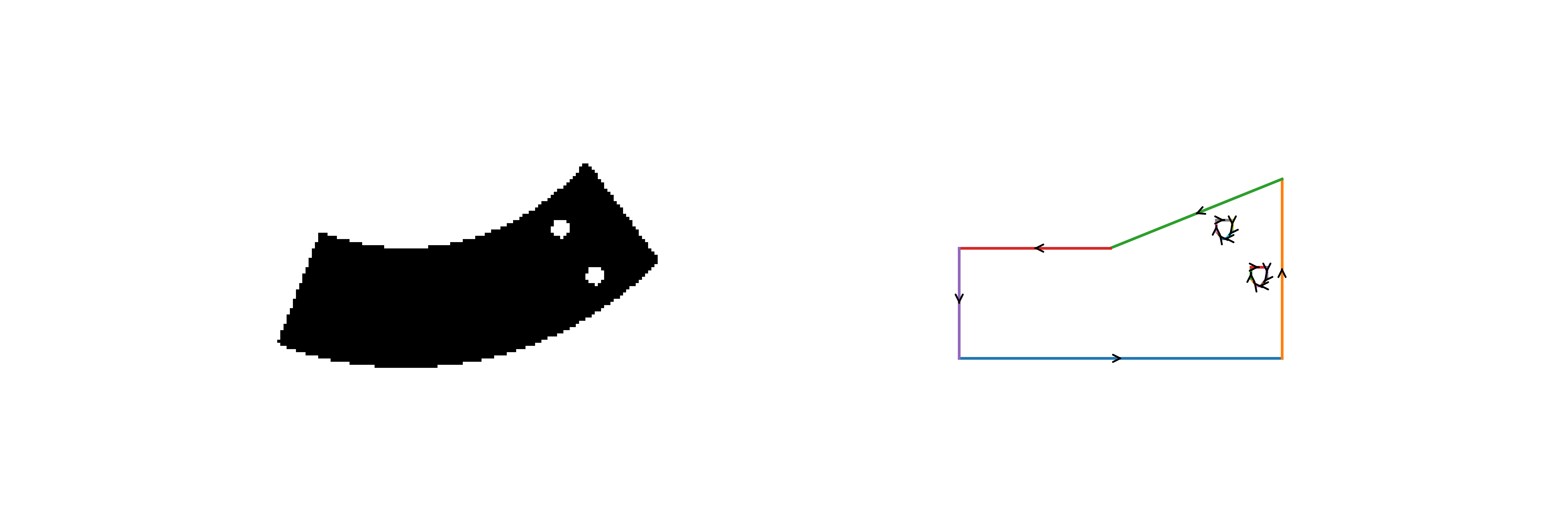} & 
    \includegraphics[width=\resultimgtwodwidth\linewidth]{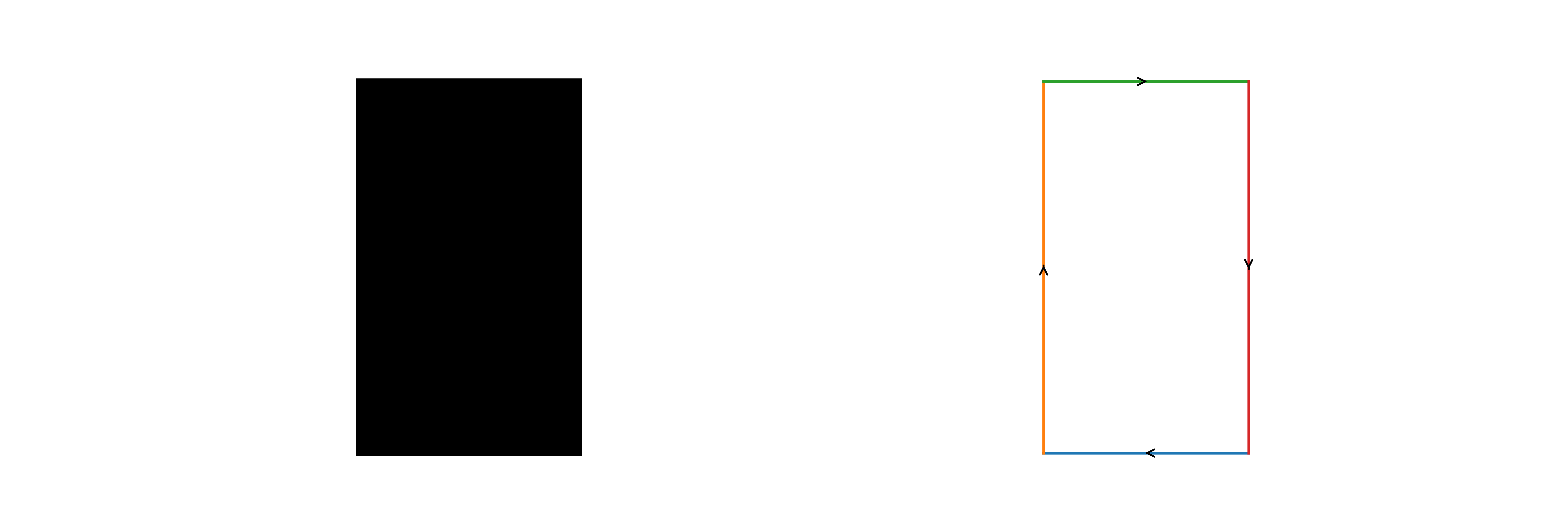} \\
    \includegraphics[width=\resultimgtwodwidth\linewidth]{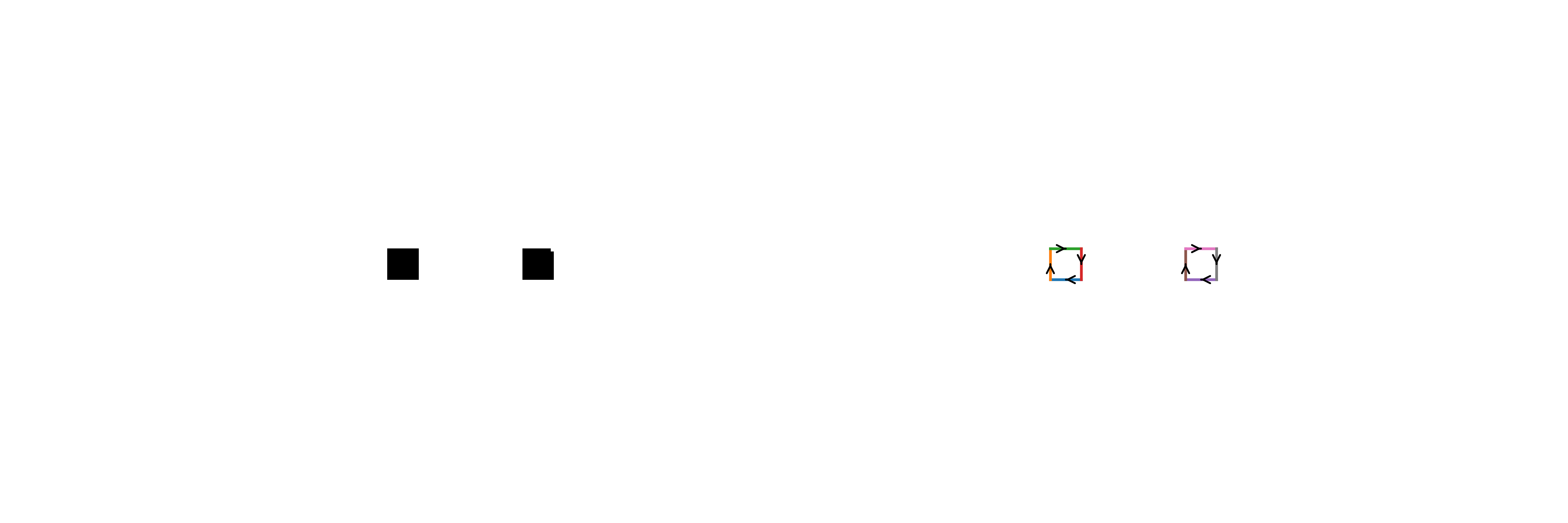} & 
    \includegraphics[width=\resultimgtwodwidth\linewidth]{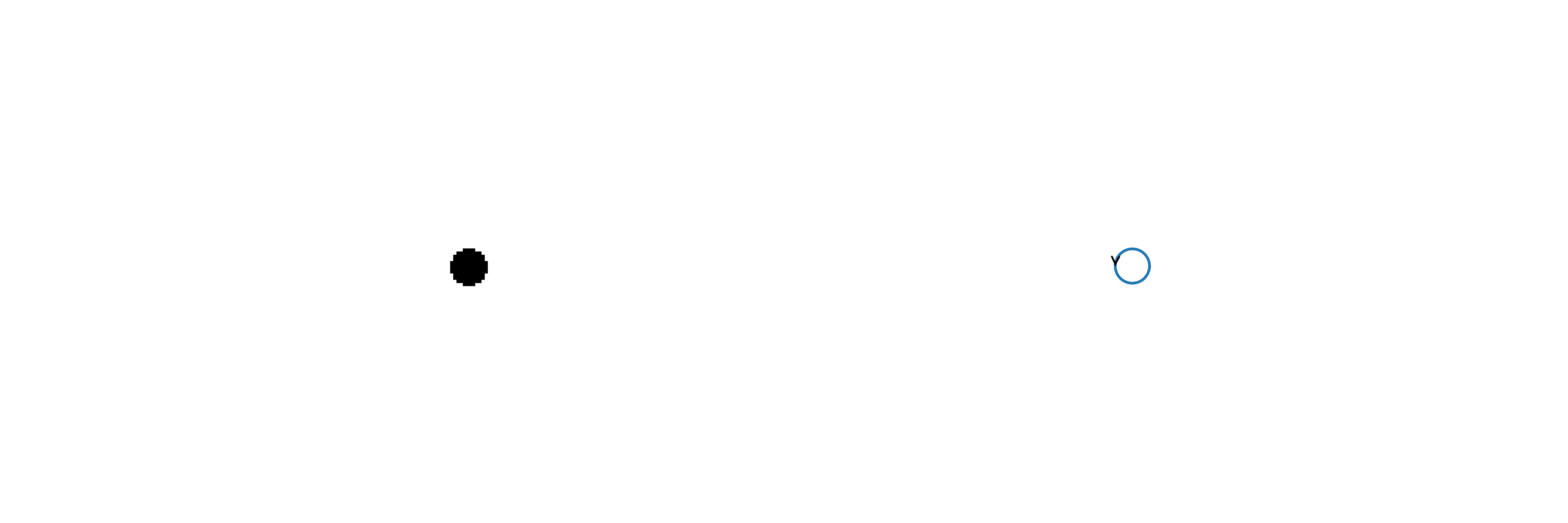} & 
    \includegraphics[width=\resultimgtwodwidth\linewidth]{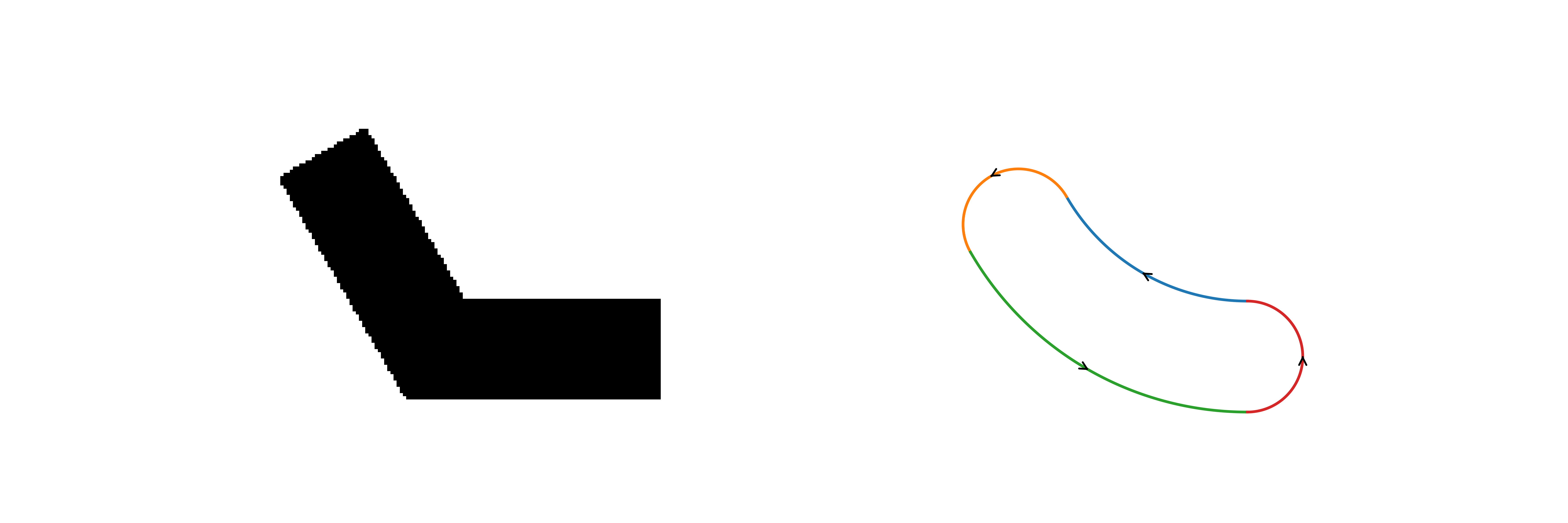} \\
    \includegraphics[width=\resultimgtwodwidth\linewidth]{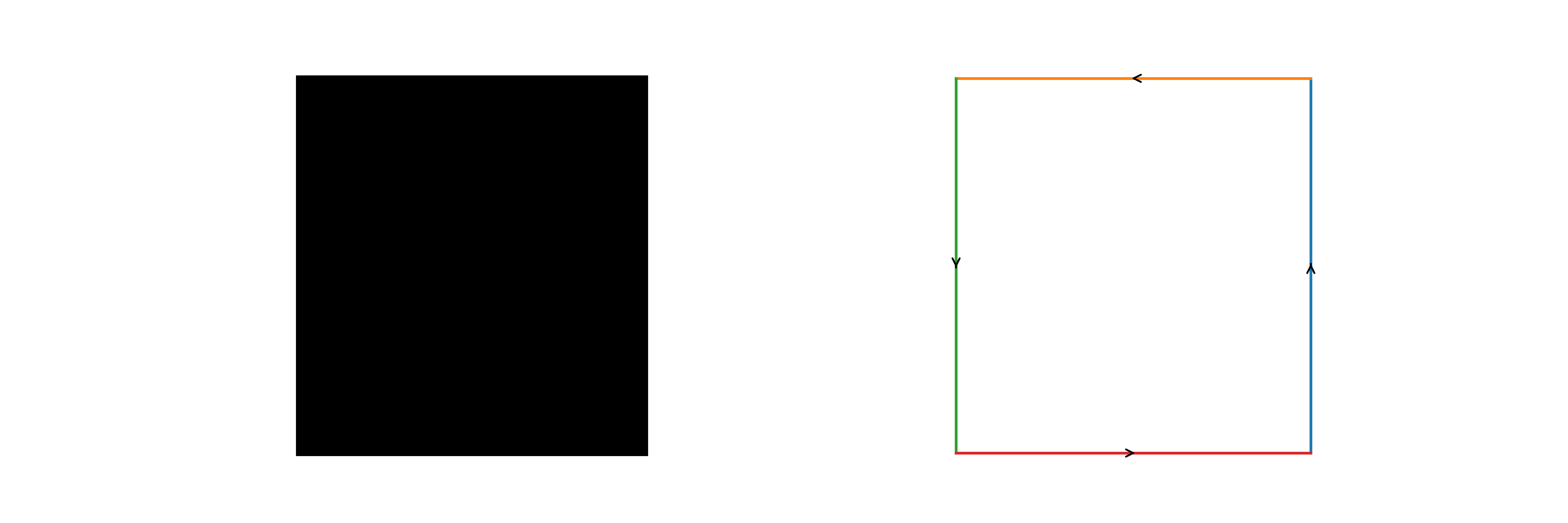} & 
    \includegraphics[width=\resultimgtwodwidth\linewidth]{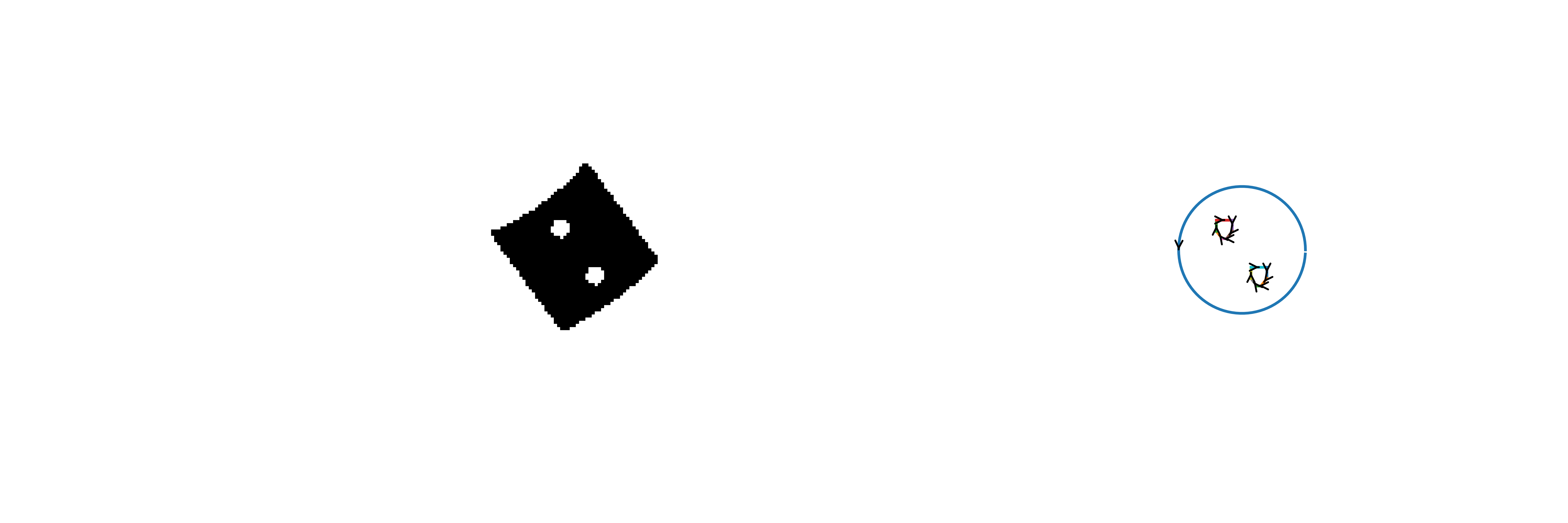} & 
    \includegraphics[width=\resultimgtwodwidth\linewidth]{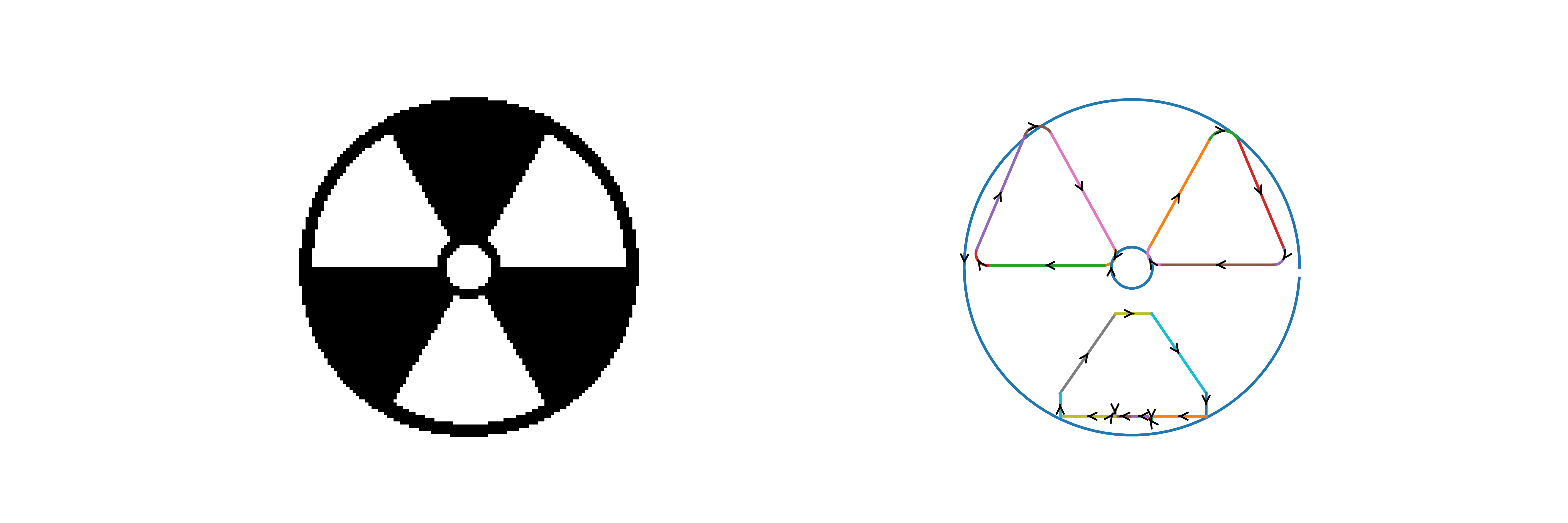} 
    \end{tabular}        
    \caption{Randomly selected retrieval results for profiles from the DeepCAD dataset.  The target shapes are shown in black and the retrieved profile loops  shown as colored curves.  As a large fraction of the dataset is created from rectangular and circular loops, many of the shapes are very well approximated even with a small search set.}
    \label{figure:2d_retrieval}
  \end{center}
\end{figure}
\endgroup
\subsection{Retrieving sketches and rebuilding CAD}
\label{section:search_retrieval_fit}
At inference time the 2D profiles images and 1D envelope arrays can be used to create a parametric recipe suitable for use in a CAD system.   The profile images are converted to CAD profiles using a search procedure in an embedding space defined by a 2D autoencoder.  This is built from the CNN encoder and deconvolution decoder as described in Appendix    \sectionappendixencodertwod ~ and  \sectionappendixdecodertwod.  As in the 3D case, the input images are signed distance functions and the model is trained using the loss from Equation \ref{equation:profile_loss}.  Once trained, the 2D autoencoder can be used to generate embeddings for each variation of a collection of constrained parametric sketches.  The 2D profile images created by the 3D decoder network can then be converted to binary images by selecting the pixels with values lower than zero.  The resulting shapes can be separated into connected components, each of which will be converted into a single CAD profile.  The shape used to search for the outer loop of each component can be found by filling any internal holes and the inner loops extracted as the difference between the original and filled shape.  For each loop extracted in this way, we can zoom in on a square around its binary mask.  As the 2D autoencoder is trained using signed distance functions as input, the zoomed images representing single loops need to be converted to signed distance functions using the fast marching method \cite{sethian1999level}.  The search then finds the parametric sketch variation with the embedding closest, in terms of euclidean distance, to the embedding of the signed distance query image.   Once the sketch has been retrieved, its parameters can be fitted to better approximate the query shape.  In addition to the internal parameters of the parametric sketch, the translation in x and y and sketch scale factor can also be optimized. To rebuild the CAD model we also need the positions of the start and end planes for each extrusion.   These are determined by \markchange{finding the positions of zero crossings of the extrusion's start and end arrays.}  If multiple zero crossings exist, the planes are chosen such that the length of the extrusion is maximized.   A CAD kernel can then be used to generate a B-Rep extrusion of the retrieved profile.   The Boolean operations from the decoder module's recipe are then used to combine the extrusions to create the CAD model.

\section{Metrics}
The following metrics are used to evaluate the performance of the generation technique.

\paragraph{Solid model validity}
The validity of a solid model is critical to it's usability in downstream processes.  Solid models may contain problems which are not visible when the model is rendered, but cause the failure of downstream modeling operations.  For this reason we test validity using the Open Cascade \verb|BRepCheck_Analyzer.IsValid()| function as in code provided with \cite{wu2021deepcad}.   The valid ratio is measured as the fraction of voxel models which generate solid data passing the Open Cascade validity test.   Models which fail the validity test are not included in downstream metrics.   Both the IoU and FID metrics described below are computed using a voxelized version of the generated CAD.  \markchange{This cannot be generated reliably unless the solid is valid and watertight.}

\paragraph{Intersection over Union}
The intersection over union metric, also known as Jaccard index, provides an intuitive way to understand how well the regenerated model represents the shape of the target.  The CAD models are first triangulated and then voxelized, taking care to correctly account for scaling and translation.  The IoU can then be computed based on the resulting binary voxel grids. IoU measurements are always made with respect to the ground truth voxel models before any rounded augmentation has been applied.   Because our approach canonicalizes the orientation of the voxel models, our voxel embeddings do not contain information about which orientation the final model needs to be constructed in.  For this reason DeepCAD often creates shapes which look approximately correct, but are reconstructed in the wrong orientation.    To ensure a fair comparison with DeepCAD, we rotate the target voxel model by all 24 non-mirror symmetries of cube and pick the biggest value when computing the DeepCAD IoU.
\paragraph{Fr\'echet inception distance}
\markchange{The Fr\'echet inception distance (FID) \cite{heusel2017gans} is a common metric used to measure quality and diversity of generated shapes as compared to a set of ground truth shapes. The metric is computed by measuring the difference in the distribution of  activations obtained from a pre-trained model for generated shapes and ground truth shapes. In our case, we compute the metric using embeddings from a 3D voxel autoencoder, trained on 36,752 models from the ABC dataset which were identified as being consistent with axis aligned extrusions.}   Both the ground truth and generated data distributions were then created by voxelizing solid models and using the encoder of this autoencoder to produce the embeddings.  For experiments with the DeepCAD dataset, the ground truth solids were created by reconstructing 5,000 randomly selected examples from the DeepCAD training set.   \markchange{For experiments on ABC the 5,000 ground truth solids were randomly selected from the same distribution of axis aligned models from ABC.  These were not used for training or as target geometry in the experiments.}
\section{Experiments}
\subsection{2D profile fitting}
\label{section:retrieval2d}

In this section we examine the performance of our 2D search retrieval and fitting technique at reproducing the shapes found in the profiles from 3,379 randomly selected models from the DeepCAD dataset.  The 2D autoencoder was trained on 128,400 2D profiles generated by cutting axis aligned slices through 5,422 models in the ABC dataset.   For each extrusion in the DeepCAD models, we create binary images from the union of the extruded profiles.  The search and retrieval procedure described in Section \ref{section:search_retrieval_fit} is then used to approximate the loops of the profiles using the 1,690 constrained parametric sketch variations created by the procedure described in Section \ref{section:parametric_sketches}.  The IoU between the target profiles and the recovered sketches is 93$\pm$11\% which is surprisingly high given the very small number of sketches over which the search is conducted.  We also investigated using the binary mask as input to the search network rather than a signed distance function.  This results in a substantially lowers the IoU to 84$\pm$21\%.  We speculate that using a signed distance function breaks the translation invariance in the CNN encoder architecture in a similar way to the solution suggested by \cite{Liu2018}.  Figure \ref{figure:2d_retrieval} shows some randomly selected examples of target 2D shapes and the retrieved profiles.   As 41.53\% of DeepCAD profile loops are circles and 25.97\% are rectangles, we see that the successful recovery of these simple shapes provides an excellent foundation for the search based approach.  
\begin{center}
\begin{table}[t]
\begin{tabular}{ l l l l l  }
    \toprule
    Dataset & Method &
    Valid ratio ($\uparrow$) &
    IoU ($\uparrow$) &
    FID ($\downarrow$) \\ 
    
    \midrule
    
    \multirow{3}{*}{DeepCAD} & DC-P & 69.47\%  &
    47 $\pm$ 11\%   & 
    1.72e+6 \\
    
    & DC-V  & 
    71.46\% & 
    48 $\pm$ 12\% 
    & \textbf{1.28e+6} \\
    
    & Ours & 
    \textbf{83.84\%} &
    \textbf{79 $\pm$ 15\%} &
    1.3e+6\\
        
    \midrule
    
    \multirow{3}{*}{\markchange{ABC}}  & \markchange{DC-P} & \markchange{56.99\%}  &
    \markchange{51 $\pm$ 7}  \%   & 
    \markchange{1.73e+6} \\
    
    & \markchange{DC-V}  & 
    \markchange{51.66\%} & \markchange{50 $\pm$ 6\%} 
    & \markchange{7.23e+5} \\
    
    & \markchange{Ours} & 
    \markchange{\textbf{81.46\%}} &
    \markchange{\textbf{74$\pm$ 13\%}} &
    \markchange{\textbf{2.43e+5}}\\
    
    \bottomrule
\end{tabular}
\caption{Performance of the networks.  DC-P is DeepCAD with the official point cloud encoder generating the embeddings.   DC-V uses our voxel encoder to generate the embeddings.  The mean and standard deviation of the IoU values are shown.  The wide distribution reflects the fact that some models are better approximated than others.}
\label{table:results}
\end{table}
\end{center}
\subsection{3D reconstruction}
\begingroup
\renewcommand{\arraystretch}{0.0}
\begin{figure*}[t]
\begin{tabular}{ c@{\hskip 0in} c@{\hskip 0in} c@{\hskip 0in} c@{\hskip 0.3in}  c@{\hskip 0in} c@{\hskip 0in} c@{\hskip 0in} c@{\hskip 0.3in}  c@{\hskip 0in} c }

    \multicolumn{4}{c}{Simple examples} & \multicolumn{4}{c}{Complex examples} & \multicolumn{2}{c}{DeepCAD failures}\\

    \includegraphics[width=\resultimgwidth\linewidth]{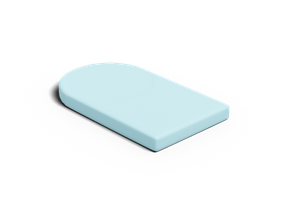} &
    \includegraphics[width=\resultimgwidth\linewidth]{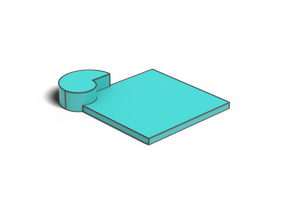} &
    \includegraphics[width=\resultimgwidth\linewidth]{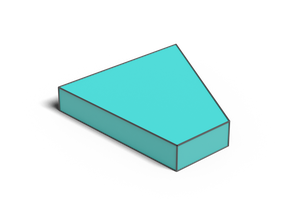} &
    \includegraphics[width=\resultimgwidth\linewidth]{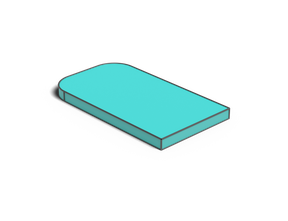} &
    
    \includegraphics[width=\resultimgwidth\linewidth]{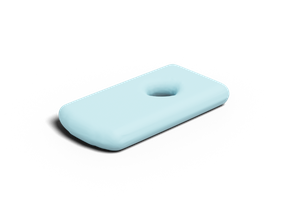} &
    \includegraphics[width=\resultimgwidth\linewidth]{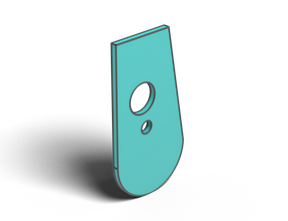} &
    \includegraphics[width=\resultimgwidth\linewidth]{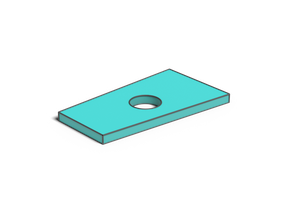} &
    \includegraphics[width=\resultimgwidth\linewidth]{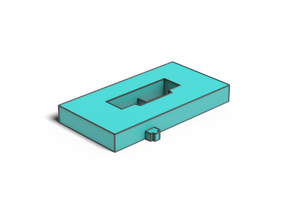} &
    
    \includegraphics[width=\resultimgwidth\linewidth]{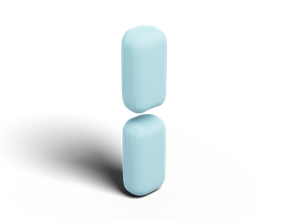} &
    \includegraphics[width=\resultimgwidth\linewidth]{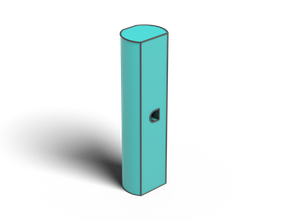} \\

    \includegraphics[width=\resultimgwidth\linewidth]{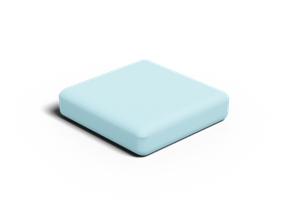} &
    \includegraphics[width=\resultimgwidth\linewidth]{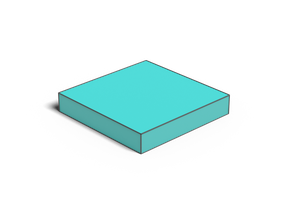} &
    \includegraphics[width=\resultimgwidth\linewidth]{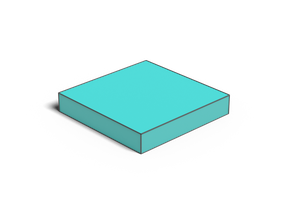} &
    \includegraphics[width=\resultimgwidth\linewidth]{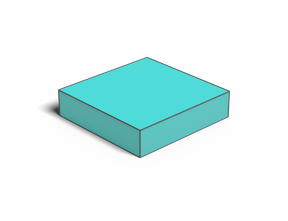} &
    
    \includegraphics[width=\resultimgwidth\linewidth]{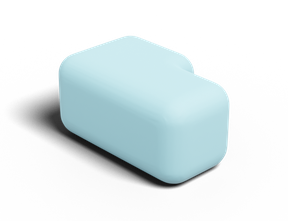} &
    \includegraphics[width=\resultimgwidth\linewidth]{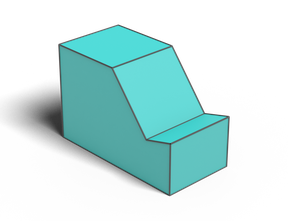} &
    \includegraphics[width=\resultimgwidth\linewidth]{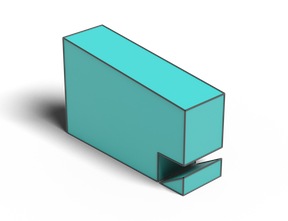} &
    \includegraphics[width=\resultimgwidth\linewidth]{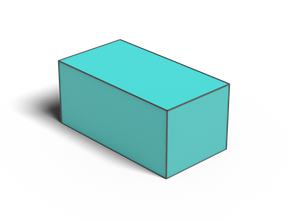} &
    
    \includegraphics[width=\resultimgwidth\linewidth]{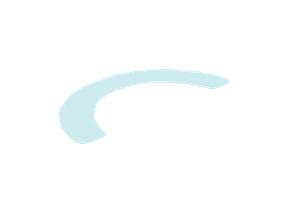} &
    \includegraphics[width=\resultimgwidth\linewidth]{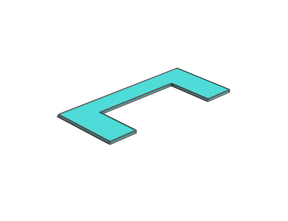} \\
    
    \includegraphics[width=\resultimgwidth\linewidth]{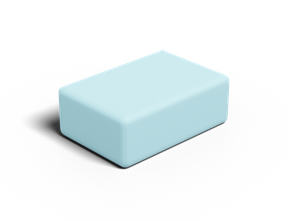} &
    \includegraphics[width=\resultimgwidth\linewidth]{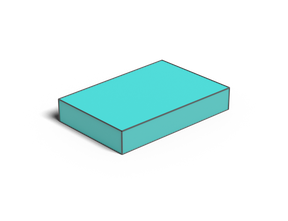} &
    \includegraphics[width=\resultimgwidth\linewidth]{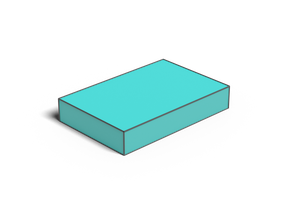} &
    \includegraphics[width=\resultimgwidth\linewidth]{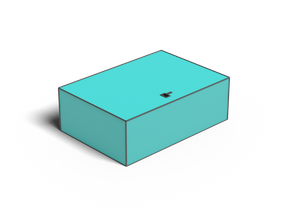} &

    \includegraphics[width=\resultimgwidth\linewidth]{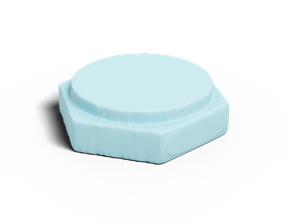} &
    \includegraphics[width=\resultimgwidth\linewidth]{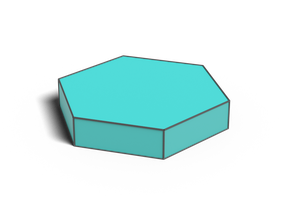} &
    \includegraphics[width=\resultimgwidth\linewidth]{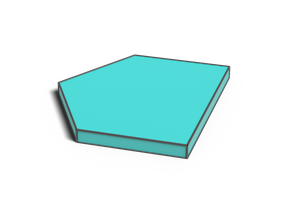} &
    \includegraphics[width=\resultimgwidth\linewidth]{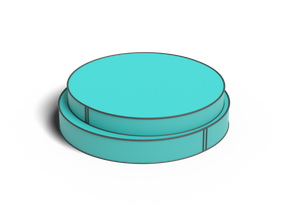} &
    
    \includegraphics[width=\resultimgwidth\linewidth]{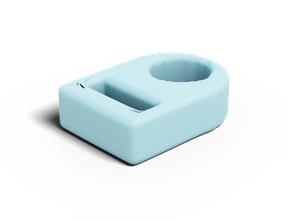} &
    \includegraphics[width=\resultimgwidth\linewidth]{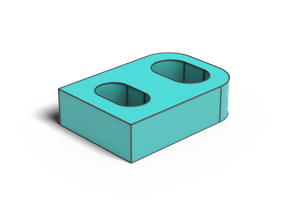} \\

    \includegraphics[width=\resultimgwidth\linewidth]{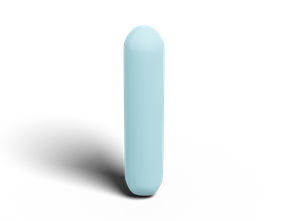} &
    \includegraphics[width=\resultimgwidth\linewidth]{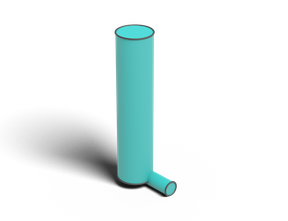} &
    \includegraphics[width=\resultimgwidth\linewidth]{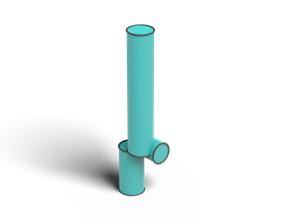} &
    \includegraphics[width=\resultimgwidth\linewidth]{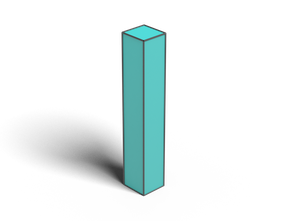} &
    
    \includegraphics[width=\resultimgwidth\linewidth]{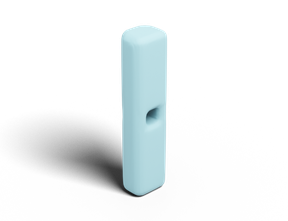} &
    \includegraphics[width=\resultimgwidth\linewidth]{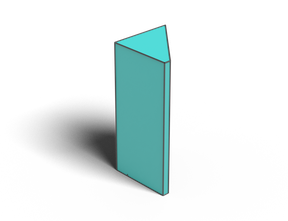} &
    \includegraphics[width=\resultimgwidth\linewidth]{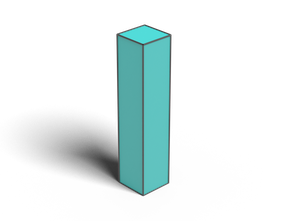} &
    \includegraphics[width=\resultimgwidth\linewidth]{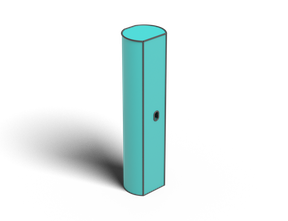} &
    
    \includegraphics[width=\resultimgwidth\linewidth]{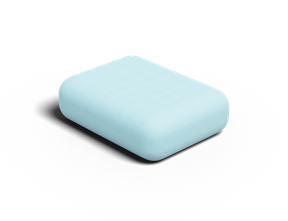} &
    \includegraphics[width=\resultimgwidth\linewidth]{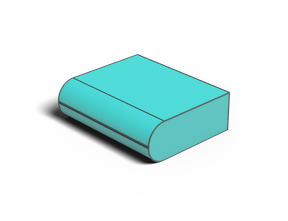} \\

    \includegraphics[width=\resultimgwidth\linewidth]{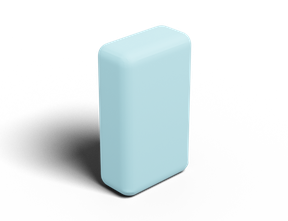} &
    \includegraphics[width=\resultimgwidth\linewidth]{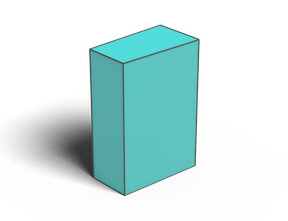} &
    \includegraphics[width=\resultimgwidth\linewidth]{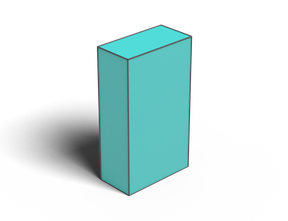} &
    \includegraphics[width=\resultimgwidth\linewidth]{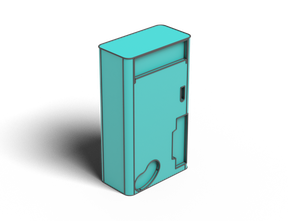} &
    
    \includegraphics[width=\resultimgwidth\linewidth]{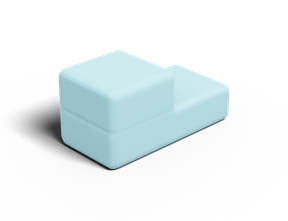} &
    \includegraphics[width=\resultimgwidth\linewidth]{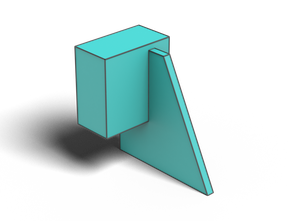} &
    \includegraphics[width=\resultimgwidth\linewidth]{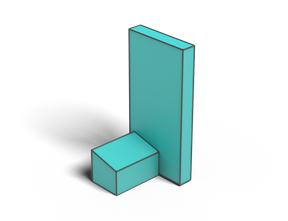} &
    \includegraphics[width=\resultimgwidth\linewidth]{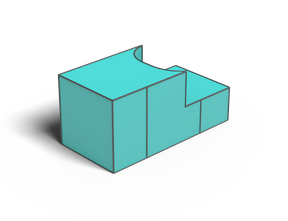} &
      
    \includegraphics[width=\resultimgwidth\linewidth]{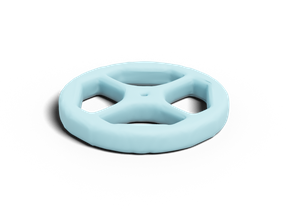} &
    \includegraphics[width=\resultimgwidth\linewidth]{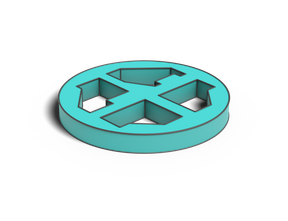} \\
    
    \includegraphics[width=\resultimgwidth\linewidth]{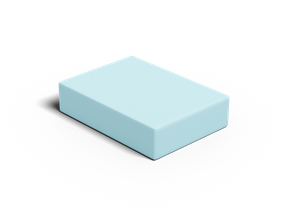} &
    \includegraphics[width=\resultimgwidth\linewidth]{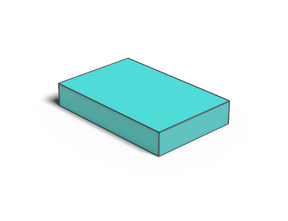} &
    \includegraphics[width=\resultimgwidth\linewidth]{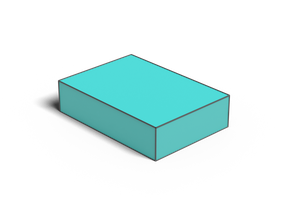} &
    \includegraphics[width=\resultimgwidth\linewidth]{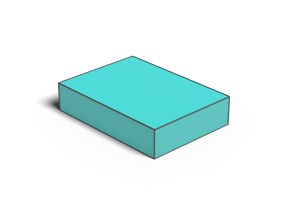} &
    
    \includegraphics[width=\resultimgwidth\linewidth]{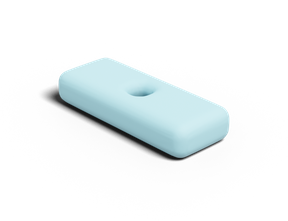} &
    \includegraphics[width=\resultimgwidth\linewidth]{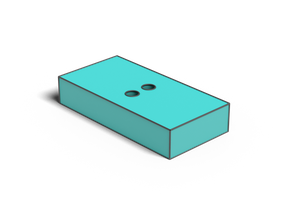} &
    \includegraphics[width=\resultimgwidth\linewidth]{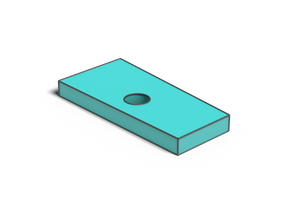} &
    \includegraphics[width=\resultimgwidth\linewidth]{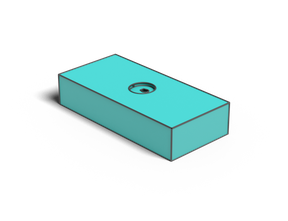} &
      
    \includegraphics[width=\resultimgwidth\linewidth]{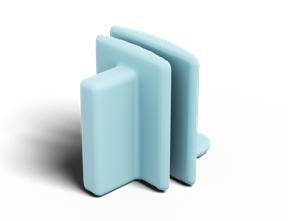} &
    \includegraphics[width=\resultimgwidth\linewidth]{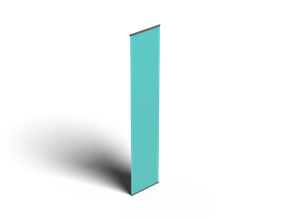} \\
    
    \includegraphics[width=\resultimgwidth\linewidth]{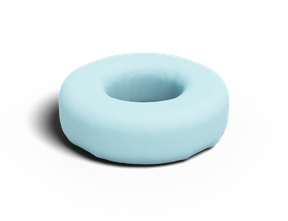} &
    \includegraphics[width=\resultimgwidth\linewidth]{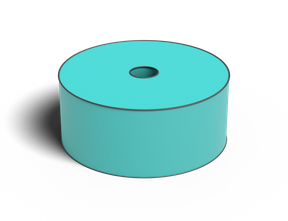} &
    \includegraphics[width=\resultimgwidth\linewidth]{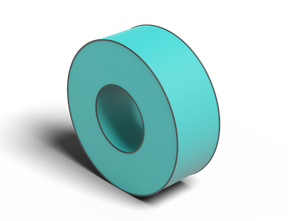} &
    \includegraphics[width=\resultimgwidth\linewidth]{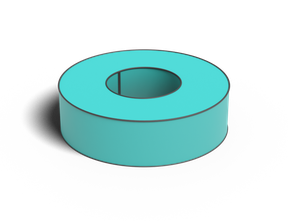} &
      
    \includegraphics[width=\resultimgwidth\linewidth]{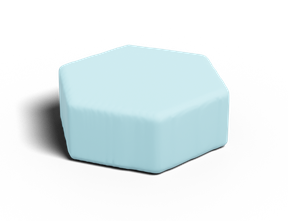} &
    \includegraphics[width=\resultimgwidth\linewidth]{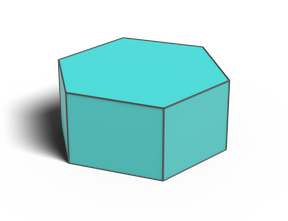} &
    \includegraphics[width=\resultimgwidth\linewidth]{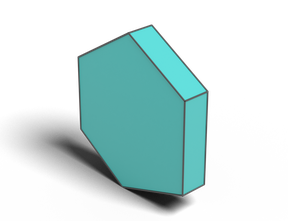} &
    \includegraphics[width=\resultimgwidth\linewidth]{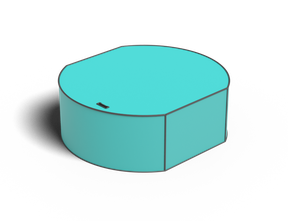} &
                
    \includegraphics[width=\resultimgwidth\linewidth]{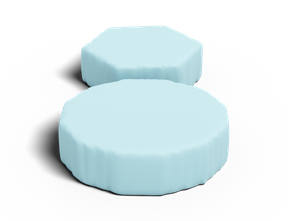} &
    \includegraphics[width=\resultimgwidth\linewidth]{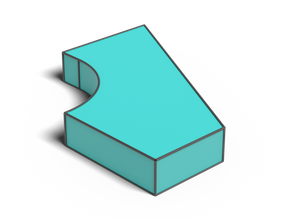} \\
    
    \includegraphics[width=\resultimgwidth\linewidth]{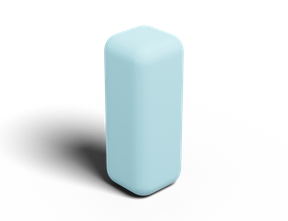} &
    \includegraphics[width=\resultimgwidth\linewidth]{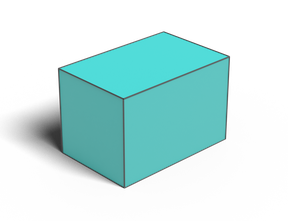} &
    \includegraphics[width=\resultimgwidth\linewidth]{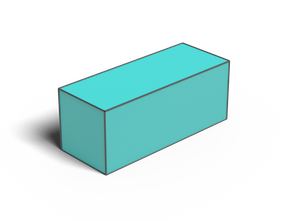} &
    \includegraphics[width=\resultimgwidth\linewidth]{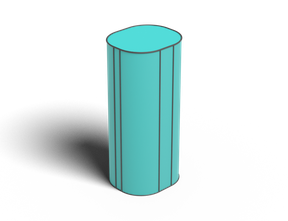} &
    
    \includegraphics[width=\resultimgwidth\linewidth]{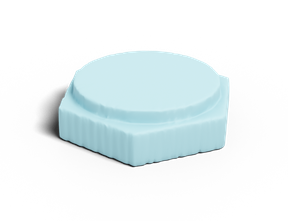} &
    \includegraphics[width=\resultimgwidth\linewidth]{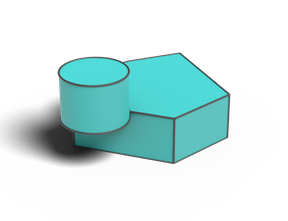} &
    \includegraphics[width=\resultimgwidth\linewidth]{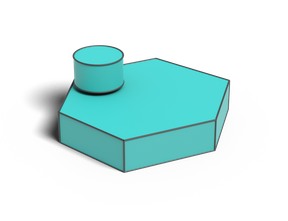} &
    \includegraphics[width=\resultimgwidth\linewidth]{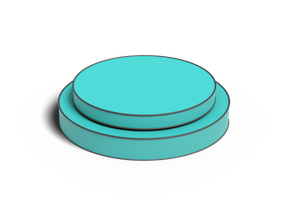} &
    
    \includegraphics[width=\resultimgwidth\linewidth]{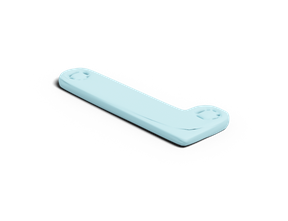} &
    \includegraphics[width=\resultimgwidth\linewidth]{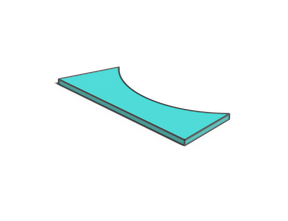} \\
    
    (a) Target & (b) DC-P & (c) DC-V & (d) Ours & (a) Target & (b) DC-P & (c) DC-V & (d) Ours & (a) Target & (d) Ours
    
    
    
    
\end{tabular}  
\caption{Randomly chosen reconstructed models.  On the left we show models which all methods could create, including blocks and cylinders.  In the middle we show more complex examples.  We exclude blocks and cylinders and then randomly sample from the remaining reconstructed results.  On the right we show examples where DeepCAD failed to generate valid solids, but our method was able to do so successfully.     (a) Target voxel model.  (b) DeepCAD using the point cloud encoder.  (c) DeepCAD using the voxel encoder.  (d) Ours}
\label{figure:random_examles_inc_blocks}
\end{figure*}
\endgroup

\label{section:reconstruction3d}
Here we compare the performance of our reconstruction technique with DeepCAD.  As our method can only generate the extrusion combinations for 80.37\% of the models in the DeepCAD dataset, we use our own train/validation/test split of 70\%/15\%/15\%. The official DeepCAD split provides only 5\% of the models in the test set and does not account for duplicates.  \markchange{As target geometry, we utilize solids from the DeepCAD test set and from ABC.  The ABC data is chosen such that can be built from extrusions, as described in Section \ref{section:data_preparation}. } 

Our network is trained on the DeepCAD dataset as described in Appendix \sectionappendixtraining. We reconstruct CAD for 6,338 voxel models from DeepCAD \markchange{ and 7,876 models from ABC.} For comparison with DeepCAD we use their pre-trained sequence encoder to generate sequence embeddings for our training set.  We then train two encoders to replicate these embeddings.   The original DeepCAD point cloud model is trained using point clouds extracted from triangulations of our voxels models, including the rounded augmentations.  We also utilize our voxel encoder to consume the same voxel models as our network and produce the  embeddings required for DeepCAD sequence generation.   Both encoders are trained with MSE loss as in the official code from \cite{wu2021deepcad}. 

The results are shown in table \ref{table:results}.   Our technique achieves the highest valid ratios of 83.84\% on the DeepCAD dataset \markchange{and 81.46\% on ABC.}  DeepCAD has a tendency to create self-intersecting profile loops which result in invalid solids.  In IoU, our method also outperforms DeepCAD.  We observe that DeepCAD often creates plausible models with approximately the correct shape, but these do not fit as closely to the target shape as ours.
We also record the IoU values for the differentiable voxel model produced by our approach.  On the DeepCAD dataset this gives a value of $88 \pm 12$\% \markchange{and for ABC $81\pm13$\%.  This shows that between 7\% and 9\% of the mismatch between the target and final model results from CAD conversion.}  This is approximately consistent with the 2D search and retrieval results where we observe the parametric sketches can approximate the ground truth profiles with a mean IoU of 93\%.  

In FID, the DeepCAD model marginally outperforms ours on the DeepCAD dataset.  We observe that DeepCAD tends to create data which is very close to the original dataset, with little adaptation to the target shape as shown quantitatively by the IoU metric. \markchange{ On the ABC dataset our method outperforms DeepCAD in FID, showing it is better able to generalize to different data distributions.  }

To show results qualitatively we randomly sample models from the DeepCAD experiments for display in Figure \ref{figure:random_examles_inc_blocks}.  Only 51\% of the voxel models were successfully regenerated by all three methods.  The DeepCAD dataset contains a large number of blocks and cylinders, so when we randomly sample examples from the successfully regenerated CAD, (see left of Figure \ref{figure:random_examles_inc_blocks}), the results are dominated by these simple shapes.   To show results of a wider variety of geometries we filter out blocks and cylinders, and then randomly sample the models in the middle of Figure  \ref{figure:random_examles_inc_blocks}.   Our technique is able to generate valid solids in cases where the DeepCAD result was invalid.  We show random samples from these on the right hand side of Figure \ref{figure:random_examles_inc_blocks}.  \markchange{We show similar results using ABC as the target models in the appendix Figure \figureabcexperiment.  We observe that as the target geometry become more complex, our technique discovers simplified shapes which approximate it, while the DeepCAD results become random collections of simple extrusions.}
\section{Conclusions and limitations}
This paper described a method for reconstructing CAD geometry from rounded voxel models.  It allows prismatic reconstruction of axis aligned extrusion models \markchange{from the ABC and }DeepCAD dataset with state of the art IoU between the reconstructed geometry and the original model before rounding is added.   The method is limited in that only a small set of pre-defined combinations of extrusions can be regenerated.  In practice we observe that a wide variety of useful shapes can be modeled using the small set of extrusion combinations supported. The search and retrieval strategy employed to recover the CAD geometry is also limited in that it can only recover sketches from a predefined search set.  While this narrows the range of topologies of profile loops which can be created, it works relatively well with the simple shapes in the available data.   The main limitation of the approach is that each profile loop is converted to CAD geometry independently.  This allows misalignment between curves and causes mismatched and incongruent profiles to be combined in a single model.   For example the right hand model in Figure \ref{figure:teaster} shows a case where a hexagon was retrieved in place of a     circular hole.  In future work we plan to refine the CAD geometry conversion technique to consider the association between loops and generate geometry consistent with the entire solid.

\bibliographystyle{ACM-Reference-Format}
\bibliography{main}


\begin{thebibliography}{51}


\ifx \showCODEN    \undefined \def \showCODEN     #1{\unskip}     \fi
\ifx \showDOI      \undefined \def \showDOI       #1{#1}\fi
\ifx \showISBNx    \undefined \def \showISBNx     #1{\unskip}     \fi
\ifx \showISBNxiii \undefined \def \showISBNxiii  #1{\unskip}     \fi
\ifx \showISSN     \undefined \def \showISSN      #1{\unskip}     \fi
\ifx \showLCCN     \undefined \def \showLCCN      #1{\unskip}     \fi
\ifx \shownote     \undefined \def \shownote      #1{#1}          \fi
\ifx \showarticletitle \undefined \def \showarticletitle #1{#1}   \fi
\ifx \showURL      \undefined \def \showURL       {\relax}        \fi
\providecommand\bibfield[2]{#2}
\providecommand\bibinfo[2]{#2}
\providecommand\natexlab[1]{#1}
\providecommand\showeprint[2][]{arXiv:#2}

\bibitem[\protect\citeauthoryear{B{\ae}rentzen}{B{\ae}rentzen}{2005}]%
        {baerentzen2005robust}
\bibfield{author}{\bibinfo{person}{J~Andreas B{\ae}rentzen}.}
  \bibinfo{year}{2005}\natexlab{}.
\newblock \showarticletitle{Robust generation of signed distance fields from
  triangle meshes}. In \bibinfo{booktitle}{\emph{Fourth International Workshop
  on Volume Graphics, 2005.}} IEEE, \bibinfo{publisher}{IEEE},
  \bibinfo{pages}{167--239}.
\newblock


\bibitem[\protect\citeauthoryear{Benkő, Martin, and Várady}{Benkő
  et~al\mbox{.}}{2001}]%
        {BENKO2001839}
\bibfield{author}{\bibinfo{person}{Pál Benkő}, \bibinfo{person}{Ralph~R.
  Martin}, {and} \bibinfo{person}{Tamás Várady}.}
  \bibinfo{year}{2001}\natexlab{}.
\newblock \showarticletitle{Algorithms for reverse engineering boundary
  representation models}.
\newblock \bibinfo{journal}{\emph{Computer-Aided Design}} \bibinfo{volume}{33},
  \bibinfo{number}{11} (\bibinfo{year}{2001}), \bibinfo{pages}{839--851}.
\newblock
\showISSN{0010-4485}
\urldef\tempurl%
\url{https://doi.org/10.1016/S0010-4485(01)00100-2}
\showDOI{\tempurl}


\bibitem[\protect\citeauthoryear{Benkő and Várady}{Benkő and
  Várady}{2004}]%
        {BENKO2004511}
\bibfield{author}{\bibinfo{person}{Pál Benkő} {and} \bibinfo{person}{Tamás
  Várady}.} \bibinfo{year}{2004}\natexlab{}.
\newblock \showarticletitle{Segmentation methods for smooth point regions of
  conventional engineering objects}.
\newblock \bibinfo{journal}{\emph{Computer-Aided Design}} \bibinfo{volume}{36},
  \bibinfo{number}{6} (\bibinfo{year}{2004}), \bibinfo{pages}{511--523}.
\newblock
\showISSN{0010-4485}
\urldef\tempurl%
\url{https://doi.org/10.1016/S0010-4485(03)00159-3}
\showDOI{\tempurl}


\bibitem[\protect\citeauthoryear{Bommes, Zimmer, and Kobbelt}{Bommes
  et~al\mbox{.}}{2009}]%
        {Bommes2009}
\bibfield{author}{\bibinfo{person}{David Bommes}, \bibinfo{person}{Henrik
  Zimmer}, {and} \bibinfo{person}{Leif Kobbelt}.}
  \bibinfo{year}{2009}\natexlab{}.
\newblock \showarticletitle{Mixed-Integer Quadrangulation}.
\newblock \bibinfo{journal}{\emph{ACM Trans. Graph.}} \bibinfo{volume}{28},
  \bibinfo{number}{3}, Article \bibinfo{articleno}{77} (\bibinfo{date}{jul}
  \bibinfo{year}{2009}), \bibinfo{numpages}{10}~pages.
\newblock
\showISSN{0730-0301}
\urldef\tempurl%
\url{https://doi.org/10.1145/1531326.1531383}
\showDOI{\tempurl}


\bibitem[\protect\citeauthoryear{Buonamici, Carfagni, Furferi, Governi, Lapini,
  and Volpe}{Buonamici et~al\mbox{.}}{2018}]%
        {Buonamici2018}
\bibfield{author}{\bibinfo{person}{Francesco Buonamici},
  \bibinfo{person}{Monica Carfagni}, \bibinfo{person}{Rocco Furferi},
  \bibinfo{person}{Lapo Governi}, \bibinfo{person}{Alessandro Lapini}, {and}
  \bibinfo{person}{Yary Volpe}.} \bibinfo{year}{2018}\natexlab{}.
\newblock \showarticletitle{Reverse engineering modeling methods and tools: a
  survey}.
\newblock \bibinfo{journal}{\emph{Computer-Aided Design and Applications}}
  \bibinfo{volume}{15}, \bibinfo{number}{3} (\bibinfo{year}{2018}),
  \bibinfo{pages}{443--464}.
\newblock
\urldef\tempurl%
\url{https://doi.org/10.1080/16864360.2017.1397894}
\showDOI{\tempurl}
\showeprint{https://doi.org/10.1080/16864360.2017.1397894}


\bibitem[\protect\citeauthoryear{Calakli and Taubin}{Calakli and
  Taubin}{2011}]%
        {Calakli2011}
\bibfield{author}{\bibinfo{person}{F. Calakli} {and} \bibinfo{person}{Gabriel
  Taubin}.} \bibinfo{year}{2011}\natexlab{}.
\newblock \showarticletitle{SSD: Smooth Signed Distance Surface
  Reconstruction}.
\newblock \bibinfo{journal}{\emph{Computer Graphics Forum}}
  \bibinfo{volume}{30} (\bibinfo{date}{11} \bibinfo{year}{2011}),
  \bibinfo{pages}{1993 -- 2002}.
\newblock
\urldef\tempurl%
\url{https://doi.org/10.1111/j.1467-8659.2011.02058.x}
\showDOI{\tempurl}


\bibitem[\protect\citeauthoryear{Chaudhuri, Ritchie, Wu, Xu, and
  Zhang}{Chaudhuri et~al\mbox{.}}{2020}]%
        {chaudhuri2020learning}
\bibfield{author}{\bibinfo{person}{Siddhartha Chaudhuri},
  \bibinfo{person}{Daniel Ritchie}, \bibinfo{person}{Jiajun Wu},
  \bibinfo{person}{Kai Xu}, {and} \bibinfo{person}{Hao Zhang}.}
  \bibinfo{year}{2020}\natexlab{}.
\newblock \showarticletitle{Learning generative models of 3D structures}. In
  \bibinfo{booktitle}{\emph{Computer Graphics Forum}},
  Vol.~\bibinfo{volume}{39}. Wiley Online Library, \bibinfo{publisher}{Wiley},
  \bibinfo{pages}{643--666}.
\newblock


\bibitem[\protect\citeauthoryear{Chen, Tagliasacchi, and Zhang}{Chen
  et~al\mbox{.}}{2020}]%
        {chen2020bsp}
\bibfield{author}{\bibinfo{person}{Zhiqin Chen}, \bibinfo{person}{Andrea
  Tagliasacchi}, {and} \bibinfo{person}{Hao Zhang}.}
  \bibinfo{year}{2020}\natexlab{}.
\newblock \showarticletitle{Bsp-net: Generating compact meshes via binary space
  partitioning}. In \bibinfo{booktitle}{\emph{Proceedings of the IEEE/CVF
  Conference on Computer Vision and Pattern Recognition}}.
  \bibinfo{pages}{45--54}.
\newblock


\bibitem[\protect\citeauthoryear{Deng, Genova, Yazdani, Bouaziz, Hinton, and
  Tagliasacchi}{Deng et~al\mbox{.}}{2020}]%
        {deng2020cvxnet}
\bibfield{author}{\bibinfo{person}{Boyang Deng}, \bibinfo{person}{Kyle Genova},
  \bibinfo{person}{Soroosh Yazdani}, \bibinfo{person}{Sofien Bouaziz},
  \bibinfo{person}{Geoffrey Hinton}, {and} \bibinfo{person}{Andrea
  Tagliasacchi}.} \bibinfo{year}{2020}\natexlab{}.
\newblock \showarticletitle{Cvxnet: Learnable convex decomposition}. In
  \bibinfo{booktitle}{\emph{Proceedings of the IEEE/CVF Conference on Computer
  Vision and Pattern Recognition}}. \bibinfo{pages}{31--44}.
\newblock


\bibitem[\protect\citeauthoryear{Dijkstra}{Dijkstra}{1959}]%
        {Dijkstra59anote}
\bibfield{author}{\bibinfo{person}{E.~W. Dijkstra}.}
  \bibinfo{year}{1959}\natexlab{}.
\newblock \showarticletitle{A Note on Two Problems in Connexion with Graphs}.
\newblock \bibinfo{journal}{\emph{NUMERISCHE MATHEMATIK}} \bibinfo{volume}{1},
  \bibinfo{number}{1} (\bibinfo{year}{1959}), \bibinfo{pages}{269--271}.
\newblock


\bibitem[\protect\citeauthoryear{Du, Inala, Pu, Spielberg, Schulz, Rus,
  Solar-Lezama, and Matusik}{Du et~al\mbox{.}}{2018}]%
        {du2018inversecsg}
\bibfield{author}{\bibinfo{person}{Tao Du}, \bibinfo{person}{Jeevana~Priya
  Inala}, \bibinfo{person}{Yewen Pu}, \bibinfo{person}{Andrew Spielberg},
  \bibinfo{person}{Adriana Schulz}, \bibinfo{person}{Daniela Rus},
  \bibinfo{person}{Armando Solar-Lezama}, {and} \bibinfo{person}{Wojciech
  Matusik}.} \bibinfo{year}{2018}\natexlab{}.
\newblock \showarticletitle{Inversecsg: Automatic conversion of 3d models to
  csg trees}.
\newblock \bibinfo{journal}{\emph{Annual Conference on Computer Graphics and
  Interactive Techniques (SIGGRAPH)}} \bibinfo{volume}{37}, \bibinfo{number}{6}
  (\bibinfo{year}{2018}), \bibinfo{pages}{1--16}.
\newblock


\bibitem[\protect\citeauthoryear{Ellis, Nye, Pu, Sosa, Tenenbaum, and
  Solar-Lezama}{Ellis et~al\mbox{.}}{2019}]%
        {ellis2019write}
\bibfield{author}{\bibinfo{person}{Kevin Ellis}, \bibinfo{person}{Maxwell Nye},
  \bibinfo{person}{Yewen Pu}, \bibinfo{person}{Felix Sosa},
  \bibinfo{person}{Josh Tenenbaum}, {and} \bibinfo{person}{Armando
  Solar-Lezama}.} \bibinfo{year}{2019}\natexlab{}.
\newblock \showarticletitle{Write, execute, assess: Program synthesis with a
  repl}. In \bibinfo{booktitle}{\emph{Proceedings of the 33rd International
  Conference on Neural Information Processing Systems}}.
  \bibinfo{publisher}{Curran Associates Inc.}, \bibinfo{address}{Red Hook, NY,
  USA}, Article \bibinfo{articleno}{822}, \bibinfo{numpages}{10}~pages.
\newblock


\bibitem[\protect\citeauthoryear{Ganin, Bartunov, Li, Keller, and
  Saliceti}{Ganin et~al\mbox{.}}{2021}]%
        {Ganin2021ComputerAidedDA}
\bibfield{author}{\bibinfo{person}{Yaroslav Ganin}, \bibinfo{person}{Sergey
  Bartunov}, \bibinfo{person}{Yujia Li}, \bibinfo{person}{Ethan Keller}, {and}
  \bibinfo{person}{Stefano Saliceti}.} \bibinfo{year}{2021}\natexlab{}.
\newblock \showarticletitle{Computer-aided design as language}. In
  \bibinfo{booktitle}{\emph{Advances in Neural Information Processing Systems
  (NeurIPS)}}. \bibinfo{publisher}{Advances in Neural Information Processing
  Systems (NeurIPS)}.
\newblock


\bibitem[\protect\citeauthoryear{Guo, Liu, Pan, Liu, Tong, and Guo}{Guo
  et~al\mbox{.}}{2022}]%
        {Guo2022ComplexGen}
\bibfield{author}{\bibinfo{person}{Haoxiang Guo}, \bibinfo{person}{Shilin Liu},
  \bibinfo{person}{Hao Pan}, \bibinfo{person}{Yang Liu}, \bibinfo{person}{Xin
  Tong}, {and} \bibinfo{person}{Baining Guo}.} \bibinfo{year}{2022}\natexlab{}.
\newblock \showarticletitle{ComplexGen: CAD Reconstruction by B-Rep Chain
  Complex Generation}.
\newblock \bibinfo{journal}{\emph{ACM Transactions on Graphics (TOG)}}
  \bibinfo{volume}{41} (\bibinfo{year}{2022}), \bibinfo{pages}{1 -- 18}.
\newblock


\bibitem[\protect\citeauthoryear{Heusel, Ramsauer, Unterthiner, Nessler, and
  Hochreiter}{Heusel et~al\mbox{.}}{2017}]%
        {heusel2017gans}
\bibfield{author}{\bibinfo{person}{Martin Heusel}, \bibinfo{person}{Hubert
  Ramsauer}, \bibinfo{person}{Thomas Unterthiner}, \bibinfo{person}{Bernhard
  Nessler}, {and} \bibinfo{person}{Sepp Hochreiter}.}
  \bibinfo{year}{2017}\natexlab{}.
\newblock \showarticletitle{Gans trained by a two time-scale update rule
  converge to a local nash equilibrium}.
\newblock \bibinfo{journal}{\emph{Advances in neural information processing
  systems}}  \bibinfo{volume}{30} (\bibinfo{year}{2017}).
\newblock


\bibitem[\protect\citeauthoryear{Jayaraman, Lambourne, Desai, Willis, Sanghi,
  and Morris}{Jayaraman et~al\mbox{.}}{2022}]%
        {jayaraman2022solidgen}
\bibfield{author}{\bibinfo{person}{Pradeep~Kumar Jayaraman},
  \bibinfo{person}{Joseph~G Lambourne}, \bibinfo{person}{Nishkrit Desai},
  \bibinfo{person}{Karl~DD Willis}, \bibinfo{person}{Aditya Sanghi}, {and}
  \bibinfo{person}{Nigel~JW Morris}.} \bibinfo{year}{2022}\natexlab{}.
\newblock \showarticletitle{SolidGen: An Autoregressive Model for Direct B-rep
  Synthesis}.
\newblock \bibinfo{journal}{\emph{arXiv preprint arXiv:2203.13944}}
  (\bibinfo{year}{2022}).
\newblock


\bibitem[\protect\citeauthoryear{Kania, Zi{\k{e}}ba, and Kajdanowicz}{Kania
  et~al\mbox{.}}{2020}]%
        {kania2020ucsg}
\bibfield{author}{\bibinfo{person}{Kacper Kania}, \bibinfo{person}{Maciej
  Zi{\k{e}}ba}, {and} \bibinfo{person}{Tomasz Kajdanowicz}.}
  \bibinfo{year}{2020}\natexlab{}.
\newblock \showarticletitle{UCSG-Net--Unsupervised Discovering of Constructive
  Solid Geometry Tree}. In \bibinfo{booktitle}{\emph{Advances in Neural
  Information Processing Systems (NeurIPS)}} (Vancouver, BC, Canada)
  \emph{(\bibinfo{series}{NIPS'20})}. \bibinfo{publisher}{Curran Associates
  Inc.}, \bibinfo{address}{Red Hook, NY, USA}, Article
  \bibinfo{articleno}{736}, \bibinfo{numpages}{11}~pages.
\newblock


\bibitem[\protect\citeauthoryear{Koch, Matveev, Jiang, Williams, Artemov,
  Burnaev, Alexa, Zorin, and Panozzo}{Koch et~al\mbox{.}}{2019}]%
        {koch2019abc}
\bibfield{author}{\bibinfo{person}{Sebastian Koch}, \bibinfo{person}{Albert
  Matveev}, \bibinfo{person}{Zhongshi Jiang}, \bibinfo{person}{Francis
  Williams}, \bibinfo{person}{Alexey Artemov}, \bibinfo{person}{Evgeny
  Burnaev}, \bibinfo{person}{Marc Alexa}, \bibinfo{person}{Denis Zorin}, {and}
  \bibinfo{person}{Daniele Panozzo}.} \bibinfo{year}{2019}\natexlab{}.
\newblock \showarticletitle{ABC: A big CAD model dataset for geometric deep
  learning}. In \bibinfo{booktitle}{\emph{Proceedings of the IEEE Conference on
  Computer Vision and Pattern Recognition}}. \bibinfo{pages}{9593--9603}.
\newblock


\bibitem[\protect\citeauthoryear{Li, Sung, Dubrovina, Yi, and Guibas}{Li
  et~al\mbox{.}}{2019}]%
        {Li2019SupervisedFO}
\bibfield{author}{\bibinfo{person}{Lingxiao Li}, \bibinfo{person}{Minhyuk
  Sung}, \bibinfo{person}{Anastasia Dubrovina}, \bibinfo{person}{L. Yi}, {and}
  \bibinfo{person}{Leonidas~J. Guibas}.} \bibinfo{year}{2019}\natexlab{}.
\newblock \showarticletitle{Supervised Fitting of Geometric Primitives to 3D
  Point Clouds}.
\newblock \bibinfo{journal}{\emph{2019 IEEE/CVF Conference on Computer Vision
  and Pattern Recognition (CVPR)}} (\bibinfo{year}{2019}),
  \bibinfo{pages}{2647--2655}.
\newblock


\bibitem[\protect\citeauthoryear{Li, Wu, Chrysanthou, Sharf, Cohen-Or, and
  Mitra}{Li et~al\mbox{.}}{2011}]%
        {li_globFit_sigg11}
\bibfield{author}{\bibinfo{person}{Yangyan Li}, \bibinfo{person}{Xiaokun Wu},
  \bibinfo{person}{Yiorgos Chrysanthou}, \bibinfo{person}{Andrei Sharf},
  \bibinfo{person}{Daniel Cohen-Or}, {and} \bibinfo{person}{Niloy~J. Mitra}.}
  \bibinfo{year}{2011}\natexlab{}.
\newblock \showarticletitle{GlobFit: Consistently Fitting Primitives by
  Discovering Global Relations}.
\newblock \bibinfo{journal}{\emph{ACM Transactions on Graphics}}
  \bibinfo{volume}{30}, \bibinfo{number}{4}, Article \bibinfo{articleno}{52}
  (\bibinfo{year}{2011}), \bibinfo{numpages}{12}~pages.
\newblock


\bibitem[\protect\citeauthoryear{Liu, Lehman, Molino, Such, Frank, Sergeev, and
  Yosinski}{Liu et~al\mbox{.}}{2018}]%
        {Liu2018}
\bibfield{author}{\bibinfo{person}{Rosanne Liu}, \bibinfo{person}{Joel Lehman},
  \bibinfo{person}{Piero Molino}, \bibinfo{person}{Felipe~Petroski Such},
  \bibinfo{person}{Eric Frank}, \bibinfo{person}{Alex Sergeev}, {and}
  \bibinfo{person}{Jason Yosinski}.} \bibinfo{year}{2018}\natexlab{}.
\newblock \showarticletitle{An Intriguing Failing of Convolutional Neural
  Networks and the CoordConv Solution}. In
  \bibinfo{booktitle}{\emph{Proceedings of the 32nd International Conference on
  Neural Information Processing Systems}} (Montr\'{e}al, Canada)
  \emph{(\bibinfo{series}{NIPS'18})}. \bibinfo{publisher}{Curran Associates
  Inc.}, \bibinfo{address}{Red Hook, NY, USA}, \bibinfo{pages}{9628–9639}.
\newblock


\bibitem[\protect\citeauthoryear{Mescheder, Oechsle, Niemeyer, Nowozin, and
  Geiger}{Mescheder et~al\mbox{.}}{2019}]%
        {OccupancyNetworks2019}
\bibfield{author}{\bibinfo{person}{Lars Mescheder}, \bibinfo{person}{Michael
  Oechsle}, \bibinfo{person}{Michael Niemeyer}, \bibinfo{person}{Sebastian
  Nowozin}, {and} \bibinfo{person}{Andreas Geiger}.}
  \bibinfo{year}{2019}\natexlab{}.
\newblock \showarticletitle{Occupancy Networks: Learning 3D Reconstruction in
  Function Space}. In \bibinfo{booktitle}{\emph{IEEE Conference on Computer
  Vision and Pattern Recognition (CVPR)}}.
\newblock


\bibitem[\protect\citeauthoryear{Nandi, Wilcox, Panchekha, Blau, Grossman, and
  Tatlock}{Nandi et~al\mbox{.}}{2018}]%
        {nandi2018functional}
\bibfield{author}{\bibinfo{person}{Chandrakana Nandi}, \bibinfo{person}{James~R
  Wilcox}, \bibinfo{person}{Pavel Panchekha}, \bibinfo{person}{Taylor Blau},
  \bibinfo{person}{Dan Grossman}, {and} \bibinfo{person}{Zachary Tatlock}.}
  \bibinfo{year}{2018}\natexlab{}.
\newblock \showarticletitle{Functional programming for compiling and
  decompiling computer-aided design}.
\newblock \bibinfo{journal}{\emph{Proceedings of the ACM on Programming
  Languages}} \bibinfo{volume}{2}, \bibinfo{number}{ICFP}
  (\bibinfo{year}{2018}), \bibinfo{pages}{1--31}.
\newblock


\bibitem[\protect\citeauthoryear{Nandi, Willsey, Anderson, Wilcox, Darulova,
  Grossman, and Tatlock}{Nandi et~al\mbox{.}}{2020}]%
        {nandi2020synthesizing}
\bibfield{author}{\bibinfo{person}{Chandrakana Nandi}, \bibinfo{person}{Max
  Willsey}, \bibinfo{person}{Adam Anderson}, \bibinfo{person}{James~R. Wilcox},
  \bibinfo{person}{Eva Darulova}, \bibinfo{person}{Dan Grossman}, {and}
  \bibinfo{person}{Zachary Tatlock}.} \bibinfo{year}{2020}\natexlab{}.
\newblock \showarticletitle{Synthesizing Structured CAD Models with Equality
  Saturation and Inverse Transformations}. In
  \bibinfo{booktitle}{\emph{Proceedings of the 41st ACM SIGPLAN Conference on
  Programming Language Design and Implementation}}. \bibinfo{pages}{31–44}.
\newblock


\bibitem[\protect\citeauthoryear{Nash, Ganin, Eslami, and Battaglia}{Nash
  et~al\mbox{.}}{2020}]%
        {nash2020polygen}
\bibfield{author}{\bibinfo{person}{Charlie Nash}, \bibinfo{person}{Yaroslav
  Ganin}, \bibinfo{person}{S.~M.~Ali Eslami}, {and} \bibinfo{person}{Peter~W.
  Battaglia}.} \bibinfo{year}{2020}\natexlab{}.
\newblock \showarticletitle{PolyGen: An Autoregressive Generative Model of 3D
  Meshes}.
\newblock \bibinfo{journal}{\emph{ICML}} (\bibinfo{year}{2020}).
\newblock


\bibitem[\protect\citeauthoryear{Nelder and Mead}{Nelder and Mead}{1965}]%
        {NeldMead65}
\bibfield{author}{\bibinfo{person}{John~A. Nelder} {and} \bibinfo{person}{Roger
  Mead}.} \bibinfo{year}{1965}\natexlab{}.
\newblock \showarticletitle{A simplex method for function minimization}.
\newblock \bibinfo{journal}{\emph{Computer Journal}}  \bibinfo{volume}{7}
  (\bibinfo{year}{1965}), \bibinfo{pages}{308--313}.
\newblock


\bibitem[\protect\citeauthoryear{Para, Bhat, Guerrero, Kelly, Mitra, Guibas,
  and Wonka}{Para et~al\mbox{.}}{2021}]%
        {para2021sketchgen}
\bibfield{author}{\bibinfo{person}{Wamiq~Reyaz Para},
  \bibinfo{person}{Shariq~Farooq Bhat}, \bibinfo{person}{Paul Guerrero},
  \bibinfo{person}{Tom Kelly}, \bibinfo{person}{Niloy Mitra},
  \bibinfo{person}{Leonidas Guibas}, {and} \bibinfo{person}{Peter Wonka}.}
  \bibinfo{year}{2021}\natexlab{}.
\newblock \showarticletitle{SketchGen: Generating Constrained CAD Sketches}. In
  \bibinfo{booktitle}{\emph{Advances in Neural Information Processing Systems
  (NeurIPS)}}.
\newblock


\bibitem[\protect\citeauthoryear{Sanchez, Fryazinov, and Pasko}{Sanchez
  et~al\mbox{.}}{2012}]%
        {sanchez2012efficient}
\bibfield{author}{\bibinfo{person}{Mathieu Sanchez}, \bibinfo{person}{Oleg
  Fryazinov}, {and} \bibinfo{person}{Alexander Pasko}.}
  \bibinfo{year}{2012}\natexlab{}.
\newblock \showarticletitle{Efficient evaluation of continuous signed distance
  to a polygonal mesh}. In \bibinfo{booktitle}{\emph{Proceedings of the 28th
  Spring Conference on Computer Graphics}}. \bibinfo{publisher}{Association for
  Computing Machinery}, \bibinfo{address}{New York, NY, USA},
  \bibinfo{pages}{101--108}.
\newblock


\bibitem[\protect\citeauthoryear{Schnabel, Wahl, and Klein}{Schnabel
  et~al\mbox{.}}{2007}]%
        {Schnabel2007EfficientRF}
\bibfield{author}{\bibinfo{person}{Ruwen Schnabel}, \bibinfo{person}{Roland
  Wahl}, {and} \bibinfo{person}{R. Klein}.} \bibinfo{year}{2007}\natexlab{}.
\newblock \showarticletitle{Efficient RANSAC for Point‐Cloud Shape
  Detection}.
\newblock \bibinfo{journal}{\emph{Computer Graphics Forum}}
  \bibinfo{volume}{26} (\bibinfo{year}{2007}).
\newblock


\bibitem[\protect\citeauthoryear{Schulz, Shamir, Baran, Levin, Sitthi-Amorn,
  and Matusik}{Schulz et~al\mbox{.}}{2017}]%
        {Schulz2017}
\bibfield{author}{\bibinfo{person}{Adriana Schulz}, \bibinfo{person}{Ariel
  Shamir}, \bibinfo{person}{Ilya Baran}, \bibinfo{person}{David I.~W. Levin},
  \bibinfo{person}{Pitchaya Sitthi-Amorn}, {and} \bibinfo{person}{Wojciech
  Matusik}.} \bibinfo{year}{2017}\natexlab{}.
\newblock \showarticletitle{Retrieval on Parametric Shape Collections}.
\newblock \bibinfo{journal}{\emph{ACM Trans. Graph.}} \bibinfo{volume}{36},
  \bibinfo{number}{1}, Article \bibinfo{articleno}{11} (\bibinfo{date}{Jan.}
  \bibinfo{year}{2017}), \bibinfo{numpages}{14}~pages.
\newblock
\showISSN{0730-0301}
\urldef\tempurl%
\url{https://doi.org/10.1145/2983618}
\showDOI{\tempurl}


\bibitem[\protect\citeauthoryear{Seff, Ovadia, Zhou, and Adams}{Seff
  et~al\mbox{.}}{2020}]%
        {Ari2020}
\bibfield{author}{\bibinfo{person}{Ari Seff}, \bibinfo{person}{Yaniv Ovadia},
  \bibinfo{person}{Wenda Zhou}, {and} \bibinfo{person}{Ryan~P. Adams}.}
  \bibinfo{year}{2020}\natexlab{}.
\newblock \showarticletitle{Sketch{G}raphs: A Large-Scale Dataset for Modeling
  Relational Geometry in Computer-Aided Design}. In
  \bibinfo{booktitle}{\emph{ICML 2020 Workshop on Object-Oriented Learning}}.
\newblock


\bibitem[\protect\citeauthoryear{Seff, Zhou, Richardson, and Adams}{Seff
  et~al\mbox{.}}{2022}]%
        {SeffVitruvion2021}
\bibfield{author}{\bibinfo{person}{Ari Seff}, \bibinfo{person}{Wenda Zhou},
  \bibinfo{person}{Nick Richardson}, {and} \bibinfo{person}{Ryan~P. Adams}.}
  \bibinfo{year}{2022}\natexlab{}.
\newblock \showarticletitle{Vitruvion: {A} Generative Model of Parametric {CAD}
  Sketches}. In \bibinfo{booktitle}{\emph{International Conference on Learning
  Representations (ICLR)}}.
\newblock


\bibitem[\protect\citeauthoryear{Sethian}{Sethian}{1999}]%
        {sethian1999level}
\bibfield{author}{\bibinfo{person}{J.A. Sethian}.}
  \bibinfo{year}{1999}\natexlab{}.
\newblock \bibinfo{booktitle}{\emph{Level Set Methods and Fast Marching
  Methods: Evolving Interfaces in Computational Geometry, Fluid Mechanics,
  Computer Vision, and Materials Science}}.
\newblock \bibinfo{publisher}{Cambridge University Press}.
\newblock
\showISBNx{9780521645577}
\showLCCN{98040859}
\urldef\tempurl%
\url{https://books.google.co.uk/books?id=ErpOoynE4dIC}
\showURL{%
\tempurl}


\bibitem[\protect\citeauthoryear{Sharma, Goyal, Liu, Kalogerakis, and
  Maji}{Sharma et~al\mbox{.}}{2018}]%
        {sharma2018csgnet}
\bibfield{author}{\bibinfo{person}{Gopal Sharma}, \bibinfo{person}{Rishabh
  Goyal}, \bibinfo{person}{Difan Liu}, \bibinfo{person}{Evangelos Kalogerakis},
  {and} \bibinfo{person}{Subhransu Maji}.} \bibinfo{year}{2018}\natexlab{}.
\newblock \showarticletitle{CSGNet: Neural Shape Parser for Constructive Solid
  Geometry}. In \bibinfo{booktitle}{\emph{IEEE Conference on Computer Vision
  and Pattern Recognition (CVPR)}}.
\newblock


\bibitem[\protect\citeauthoryear{Sharma, Liu, Maji, Kalogerakis, Chaudhuri, and
  Měch}{Sharma et~al\mbox{.}}{2020}]%
        {sharma2020parsenet}
\bibfield{author}{\bibinfo{person}{Gopal Sharma}, \bibinfo{person}{Difan Liu},
  \bibinfo{person}{Subhransu Maji}, \bibinfo{person}{Evangelos Kalogerakis},
  \bibinfo{person}{Siddhartha Chaudhuri}, {and} \bibinfo{person}{Radomír
  Měch}.} \bibinfo{year}{2020}\natexlab{}.
\newblock \bibinfo{title}{ParSeNet: A Parametric Surface Fitting Network for 3D
  Point Clouds}.
\newblock
\newblock
\showeprint[arxiv]{2003.12181}~[cs.CV]


\bibitem[\protect\citeauthoryear{Shervashidze, Schweitzer, van Leeuwen,
  Mehlhorn, and Borgwardt}{Shervashidze et~al\mbox{.}}{2011}]%
        {ShervashidzeSLMB11}
\bibfield{author}{\bibinfo{person}{Nino Shervashidze}, \bibinfo{person}{Pascal
  Schweitzer}, \bibinfo{person}{Erik~Jan van Leeuwen}, \bibinfo{person}{Kurt
  Mehlhorn}, {and} \bibinfo{person}{Karsten~M. Borgwardt}.}
  \bibinfo{year}{2011}\natexlab{}.
\newblock \showarticletitle{Weisfeiler-Lehman Graph Kernels}.
\newblock \bibinfo{journal}{\emph{Journal of Machine Learning Research}}
  \bibinfo{volume}{12} (\bibinfo{year}{2011}), \bibinfo{pages}{2539--2561}.
\newblock
\urldef\tempurl%
\url{http://dl.acm.org/citation.cfm?id=2078187}
\showURL{%
\tempurl}


\bibitem[\protect\citeauthoryear{Smirnov, Bessmeltsev, and Solomon}{Smirnov
  et~al\mbox{.}}{2021}]%
        {smirnov2021patches}
\bibfield{author}{\bibinfo{person}{Dmitriy Smirnov}, \bibinfo{person}{Mikhail
  Bessmeltsev}, {and} \bibinfo{person}{Justin Solomon}.}
  \bibinfo{year}{2021}\natexlab{}.
\newblock \showarticletitle{Learning Manifold Patch-Based Representations of
  Man-Made Shapes}. In \bibinfo{booktitle}{\emph{International Conference on
  Learning Representations (ICLR)}}.
\newblock


\bibitem[\protect\citeauthoryear{Tian, Luo, Sun, Ellis, Freeman, Tenenbaum, and
  Wu}{Tian et~al\mbox{.}}{2019}]%
        {tian2018learning}
\bibfield{author}{\bibinfo{person}{Yonglong Tian}, \bibinfo{person}{Andrew
  Luo}, \bibinfo{person}{Xingyuan Sun}, \bibinfo{person}{Kevin Ellis},
  \bibinfo{person}{William~T. Freeman}, \bibinfo{person}{Joshua~B. Tenenbaum},
  {and} \bibinfo{person}{Jiajun Wu}.} \bibinfo{year}{2019}\natexlab{}.
\newblock \showarticletitle{Learning to Infer and Execute 3D Shape Programs}.
  In \bibinfo{booktitle}{\emph{International Conference on Learning
  Representations (ICLR)}}.
\newblock


\bibitem[\protect\citeauthoryear{Uy, Chang, Sung, Goel, Lambourne, Birdal, and
  Guibas}{Uy et~al\mbox{.}}{2022}]%
        {uy2022point2cyl}
\bibfield{author}{\bibinfo{person}{Mikaela~Angelina Uy},
  \bibinfo{person}{Yen-yu Chang}, \bibinfo{person}{Minhyuk Sung},
  \bibinfo{person}{Purvi Goel}, \bibinfo{person}{Joseph Lambourne},
  \bibinfo{person}{Tolga Birdal}, {and} \bibinfo{person}{Leonidas Guibas}.}
  \bibinfo{year}{2022}\natexlab{}.
\newblock \showarticletitle{Point2Cyl: Reverse Engineering 3D Objects from
  Point Clouds to Extrusion Cylinders}. In \bibinfo{booktitle}{\emph{IEEE
  Conference on Computer Vision and Pattern Recognition (CVPR)}}.
\newblock


\bibitem[\protect\citeauthoryear{Uy, Huang, Sung, Birdal, and Guibas}{Uy
  et~al\mbox{.}}{2020}]%
        {uy2020deformation}
\bibfield{author}{\bibinfo{person}{Mikaela~Angelina Uy},
  \bibinfo{person}{Jingwei Huang}, \bibinfo{person}{Minhyuk Sung},
  \bibinfo{person}{Tolga Birdal}, {and} \bibinfo{person}{Leonidas Guibas}.}
  \bibinfo{year}{2020}\natexlab{}.
\newblock \showarticletitle{Deformation-aware 3d model embedding and
  retrieval}. In \bibinfo{booktitle}{\emph{European Conference on Computer
  Vision (ECCV)}}. Springer, \bibinfo{pages}{397--413}.
\newblock


\bibitem[\protect\citeauthoryear{Uy, Kim, Sung, Aigerman, Chaudhuri, and
  Guibas}{Uy et~al\mbox{.}}{2021}]%
        {uy2021joint}
\bibfield{author}{\bibinfo{person}{Mikaela~Angelina Uy},
  \bibinfo{person}{Vladimir~G. Kim}, \bibinfo{person}{Minhyuk Sung},
  \bibinfo{person}{Noam Aigerman}, \bibinfo{person}{Siddhartha Chaudhuri},
  {and} \bibinfo{person}{Leonidas Guibas}.} \bibinfo{year}{2021}\natexlab{}.
\newblock \showarticletitle{Joint Learning of 3D Shape Retrieval and
  Deformation}. In \bibinfo{booktitle}{\emph{IEEE Conference on Computer Vision
  and Pattern Recognition (CVPR)}}.
\newblock


\bibitem[\protect\citeauthoryear{Varady}{Varady}{2008}]%
        {VaradyAutomaticProcedures2008}
\bibfield{author}{\bibinfo{person}{Tamas Varady}.}
  \bibinfo{year}{2008}\natexlab{}.
\newblock \showarticletitle{Automatic Procedures to Create CAD Models from
  Measured Data}.
\newblock \bibinfo{journal}{\emph{Computer-Aided Design and Applications}}
  \bibinfo{volume}{5}, \bibinfo{number}{5} (\bibinfo{year}{2008}),
  \bibinfo{pages}{577--588}.
\newblock
\urldef\tempurl%
\url{https://doi.org/10.3722/cadaps.2008.577-588}
\showDOI{\tempurl}
\showeprint{https://www.tandfonline.com/doi/pdf/10.3722/cadaps.2008.577-588}


\bibitem[\protect\citeauthoryear{Vaswani, Shazeer, Parmar, Uszkoreit, Jones,
  Gomez, Kaiser, and Polosukhin}{Vaswani et~al\mbox{.}}{2017}]%
        {vaswani2017attention}
\bibfield{author}{\bibinfo{person}{Ashish Vaswani}, \bibinfo{person}{Noam
  Shazeer}, \bibinfo{person}{Niki Parmar}, \bibinfo{person}{Jakob Uszkoreit},
  \bibinfo{person}{Llion Jones}, \bibinfo{person}{Aidan~N. Gomez},
  \bibinfo{person}{Lukasz Kaiser}, {and} \bibinfo{person}{Illia Polosukhin}.}
  \bibinfo{year}{2017}\natexlab{}.
\newblock \showarticletitle{Attention is all you need}.
\newblock \bibinfo{journal}{\emph{Advances in Neural Information Processing
  Systems (NeurIPS)}} (\bibinfo{year}{2017}), \bibinfo{pages}{5998--6008}.
\newblock


\bibitem[\protect\citeauthoryear{Vinyals, Fortunato, and Jaitly}{Vinyals
  et~al\mbox{.}}{2015}]%
        {vinyals2015pointer}
\bibfield{author}{\bibinfo{person}{Oriol Vinyals}, \bibinfo{person}{Meire
  Fortunato}, {and} \bibinfo{person}{Navdeep Jaitly}.}
  \bibinfo{year}{2015}\natexlab{}.
\newblock \showarticletitle{Pointer networks}.
\newblock \bibinfo{journal}{\emph{Advances in Neural Information Processing
  Systems (NeurIPS)}} (\bibinfo{year}{2015}).
\newblock


\bibitem[\protect\citeauthoryear{Wang, Xu, Xu, Tagliasacchi, Zhou,
  Mahdavi-Amiri, and Zhang}{Wang et~al\mbox{.}}{2020}]%
        {wang2020pie}
\bibfield{author}{\bibinfo{person}{Xiaogang Wang}, \bibinfo{person}{Yuelang
  Xu}, \bibinfo{person}{Kai Xu}, \bibinfo{person}{Andrea Tagliasacchi},
  \bibinfo{person}{Bin Zhou}, \bibinfo{person}{Ali Mahdavi-Amiri}, {and}
  \bibinfo{person}{Hao Zhang}.} \bibinfo{year}{2020}\natexlab{}.
\newblock \showarticletitle{PIE-NET: Parametric Inference of Point Cloud
  Edges}. In \bibinfo{booktitle}{\emph{Advances in Neural Information
  Processing Systems (NeurIPS)}}, Vol.~\bibinfo{volume}{33}.
  \bibinfo{publisher}{Curran Associates, Inc.}, \bibinfo{pages}{20167--20178}.
\newblock


\bibitem[\protect\citeauthoryear{Willis, Jayaraman, Lambourne, Chu, and
  Pu}{Willis et~al\mbox{.}}{2021a}]%
        {willis2021engineering}
\bibfield{author}{\bibinfo{person}{Karl D.~D. Willis},
  \bibinfo{person}{Pradeep~Kumar Jayaraman}, \bibinfo{person}{Joseph~G.
  Lambourne}, \bibinfo{person}{Hang Chu}, {and} \bibinfo{person}{Yewen Pu}.}
  \bibinfo{year}{2021}\natexlab{a}.
\newblock \showarticletitle{Engineering Sketch Generation for Computer-Aided
  Design}. In \bibinfo{booktitle}{\emph{The 1st Workshop on Sketch-Oriented
  Deep Learning (SketchDL), CVPR 2021}}.
\newblock


\bibitem[\protect\citeauthoryear{Willis, Pu, Luo, Chu, Du, Lambourne,
  Solar-Lezama, and Matusik}{Willis et~al\mbox{.}}{2021b}]%
        {willis2020fusion}
\bibfield{author}{\bibinfo{person}{Karl D.~D. Willis}, \bibinfo{person}{Yewen
  Pu}, \bibinfo{person}{Jieliang Luo}, \bibinfo{person}{Hang Chu},
  \bibinfo{person}{Tao Du}, \bibinfo{person}{Joseph~G. Lambourne},
  \bibinfo{person}{Armando Solar-Lezama}, {and} \bibinfo{person}{Wojciech
  Matusik}.} \bibinfo{year}{2021}\natexlab{b}.
\newblock \showarticletitle{Fusion 360 Gallery: A Dataset and Environment for
  Programmatic CAD Construction from Human Design Sequences}.
\newblock \bibinfo{journal}{\emph{ACM Transactions on Graphics (TOG)}}
  \bibinfo{volume}{40}, \bibinfo{number}{4} (\bibinfo{year}{2021}).
\newblock


\bibitem[\protect\citeauthoryear{Wu, Xiao, and Zheng}{Wu et~al\mbox{.}}{2021}]%
        {wu2021deepcad}
\bibfield{author}{\bibinfo{person}{Rundi Wu}, \bibinfo{person}{Chang Xiao},
  {and} \bibinfo{person}{Changxi Zheng}.} \bibinfo{year}{2021}\natexlab{}.
\newblock \showarticletitle{DeepCAD: A Deep Generative Network for
  Computer-Aided Design Models}. In \bibinfo{booktitle}{\emph{Proceedings of
  the IEEE/CVF International Conference on Computer Vision (ICCV)}}.
\newblock


\bibitem[\protect\citeauthoryear{Xu, Peng, Cheng, Willis, and Ritchie}{Xu
  et~al\mbox{.}}{2021}]%
        {Xu2021InferringCM}
\bibfield{author}{\bibinfo{person}{Xianghao Xu}, \bibinfo{person}{Wenzhe Peng},
  \bibinfo{person}{Chin-Yi Cheng}, \bibinfo{person}{Karl D.~D. Willis}, {and}
  \bibinfo{person}{Daniel Ritchie}.} \bibinfo{year}{2021}\natexlab{}.
\newblock \showarticletitle{Inferring CAD Modeling Sequences Using Zone
  Graphs}.
\newblock \bibinfo{journal}{\emph{2021 IEEE/CVF Conference on Computer Vision
  and Pattern Recognition (CVPR)}} (\bibinfo{year}{2021}),
  \bibinfo{pages}{6058--6066}.
\newblock


\bibitem[\protect\citeauthoryear{Yan, Yang, Ma, Huang, Vouga, and Huang}{Yan
  et~al\mbox{.}}{2021}]%
        {Yan2021HPNetDP}
\bibfield{author}{\bibinfo{person}{Siming Yan}, \bibinfo{person}{Zhenpei Yang},
  \bibinfo{person}{Chongyang Ma}, \bibinfo{person}{Haibin Huang},
  \bibinfo{person}{Etienne Vouga}, {and} \bibinfo{person}{Qi-Xing Huang}.}
  \bibinfo{year}{2021}\natexlab{}.
\newblock \showarticletitle{HPNet: Deep Primitive Segmentation Using Hybrid
  Representations}.
\newblock \bibinfo{journal}{\emph{2021 IEEE/CVF International Conference on
  Computer Vision (ICCV)}} (\bibinfo{year}{2021}), \bibinfo{pages}{2733--2742}.
\newblock


\bibitem[\protect\citeauthoryear{Yu, Chen, Li, Sanghi, Shayani, Mahdavi-Amiri,
  and Zhang}{Yu et~al\mbox{.}}{2022}]%
        {yu2022capri}
\bibfield{author}{\bibinfo{person}{Fenggen Yu}, \bibinfo{person}{Zhiqin Chen},
  \bibinfo{person}{Manyi Li}, \bibinfo{person}{Aditya Sanghi},
  \bibinfo{person}{Hooman Shayani}, \bibinfo{person}{Ali Mahdavi-Amiri}, {and}
  \bibinfo{person}{Hao Zhang}.} \bibinfo{year}{2022}\natexlab{}.
\newblock \showarticletitle{CAPRI-Net: Learning Compact CAD Shapes with
  Adaptive Primitive Assembly}. In \bibinfo{booktitle}{\emph{Proceedings of the
  IEEE/CVF Conference on Computer Vision and Pattern Recognition}}.
  \bibinfo{pages}{11768--11778}.
\newblock


\end{thebibliography}

\ifincappendix
\clearpage
\appendix
\section{Appendix}
\newcommand{\imgwidth}{0.14}
\begin{figure*}[ht!]
    \begin{tabular}{c c c c c c}
        \includegraphics[width=\imgwidth\linewidth]{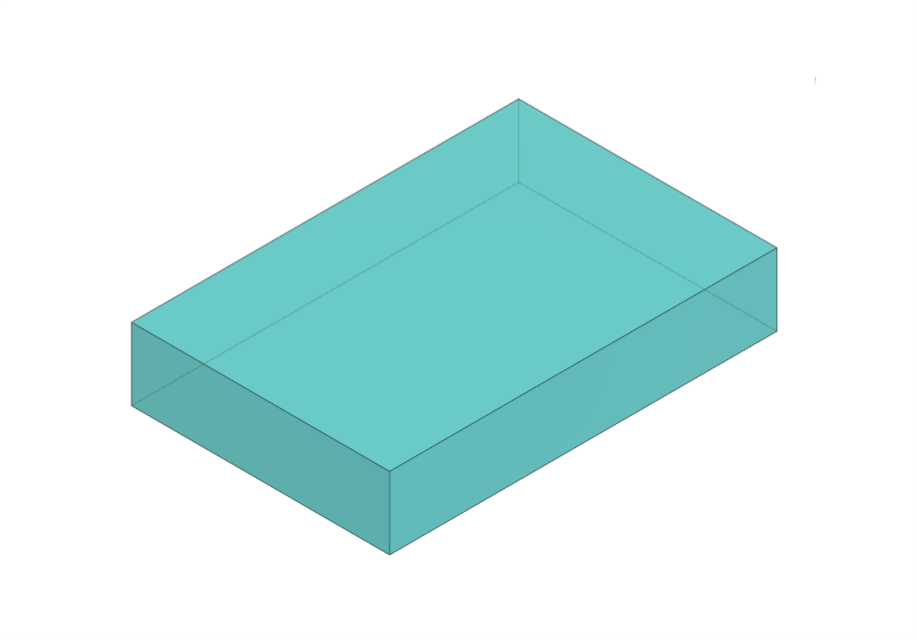}
        &  
        \includegraphics[width=\imgwidth\linewidth]{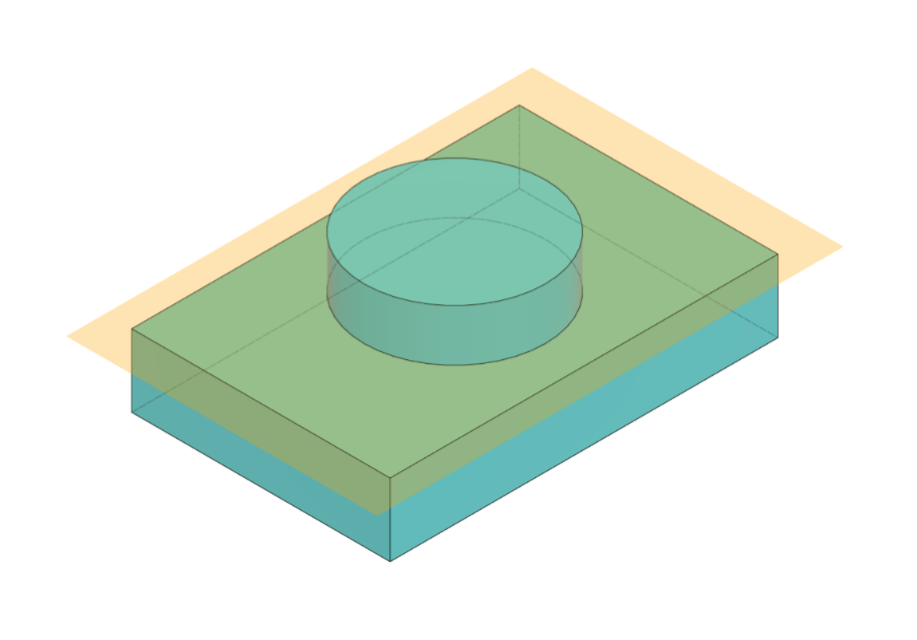}
        &      
        \includegraphics[width=\imgwidth\linewidth]{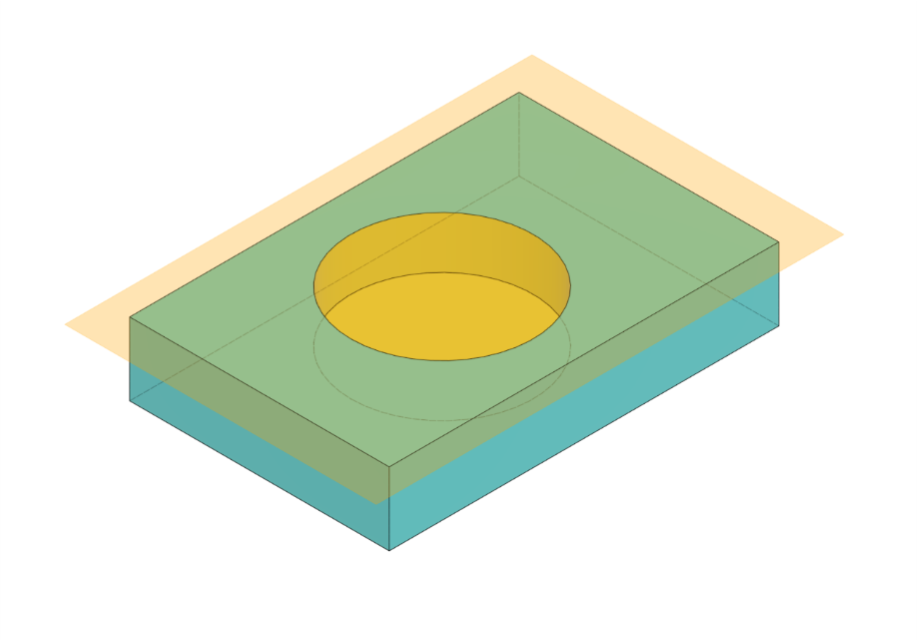}
        &
        \includegraphics[width=\imgwidth\linewidth]{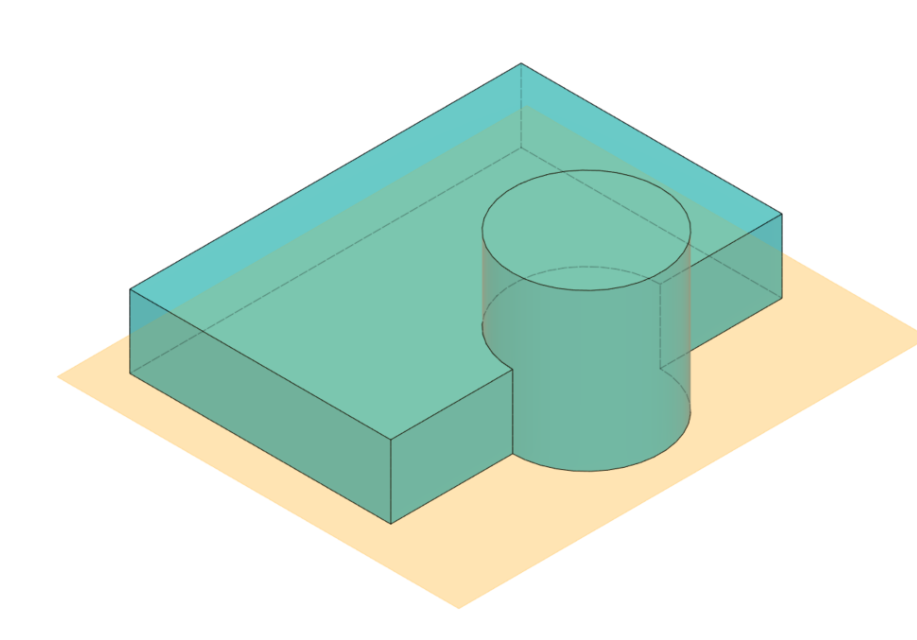}
        &
        \includegraphics[width=\imgwidth\linewidth]{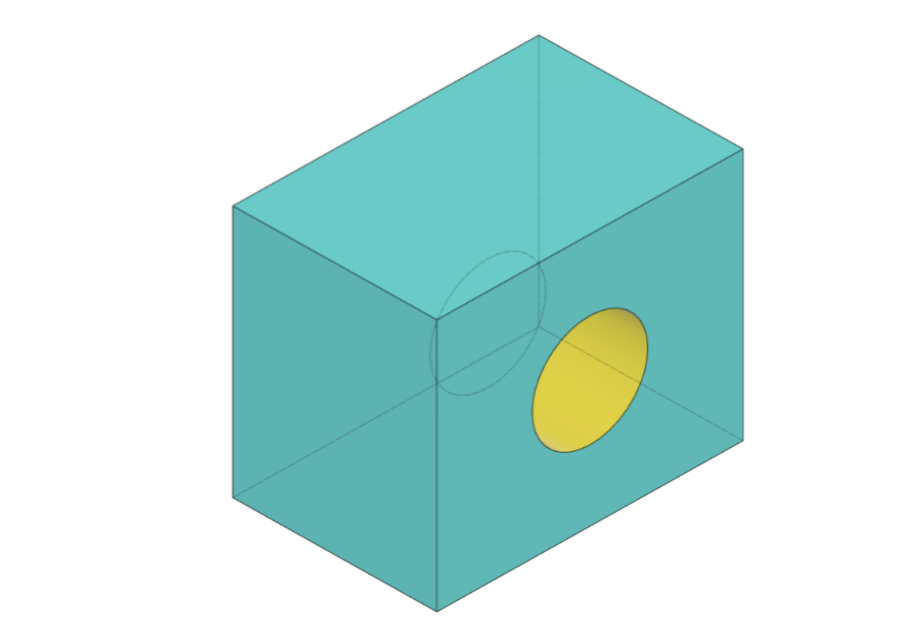}
        &
        \includegraphics[width=\imgwidth\linewidth]{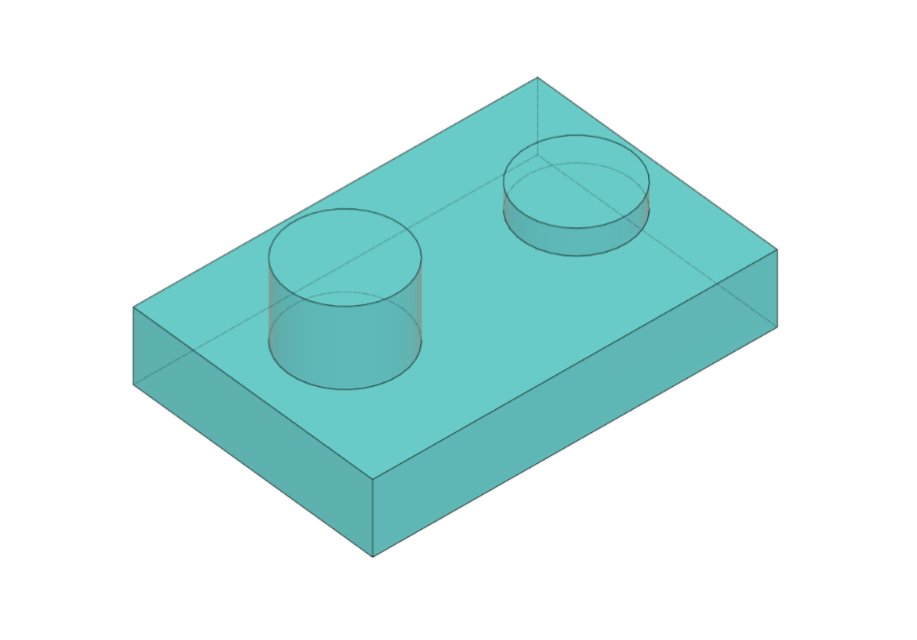}
        \\ 
        (a) 56.7\% & (b) 5.2\% & (c) 4.5\% & (d) 2.9\% & (e) 2.6\% & (f) 2.1\%
         \\
        \includegraphics[width=\imgwidth\linewidth]{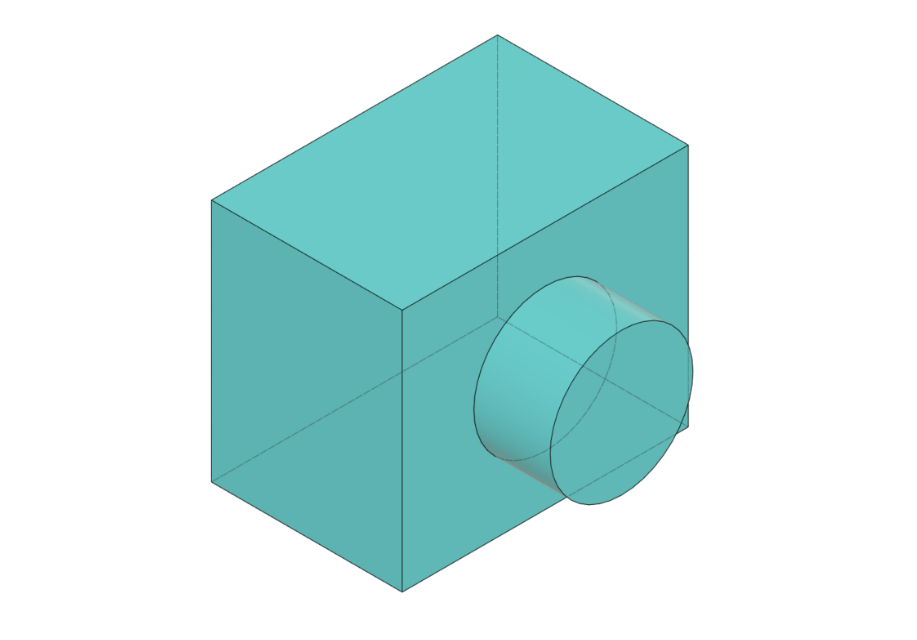}
        &      
        \includegraphics[width=\imgwidth\linewidth]{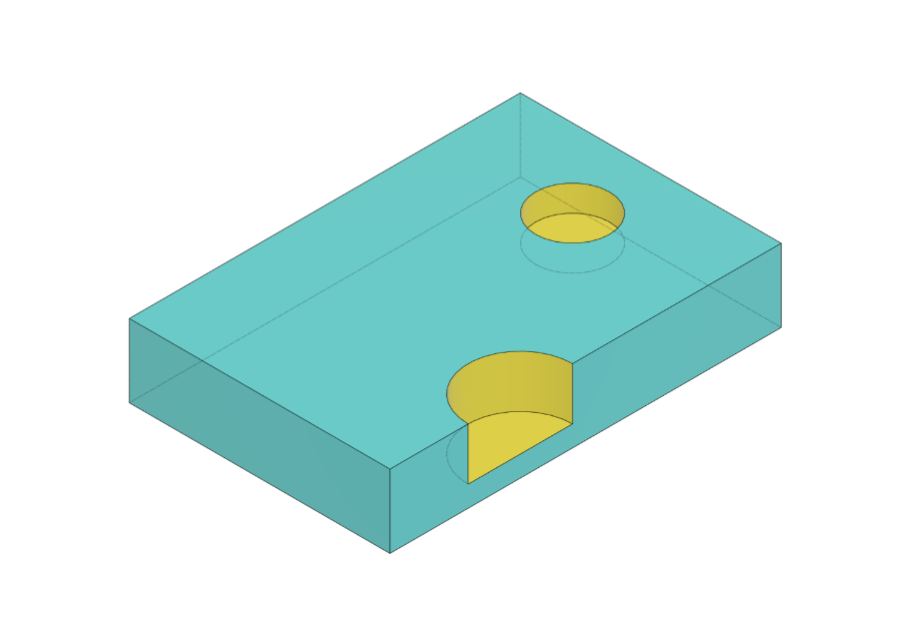}
        &
        \includegraphics[width=\imgwidth\linewidth]{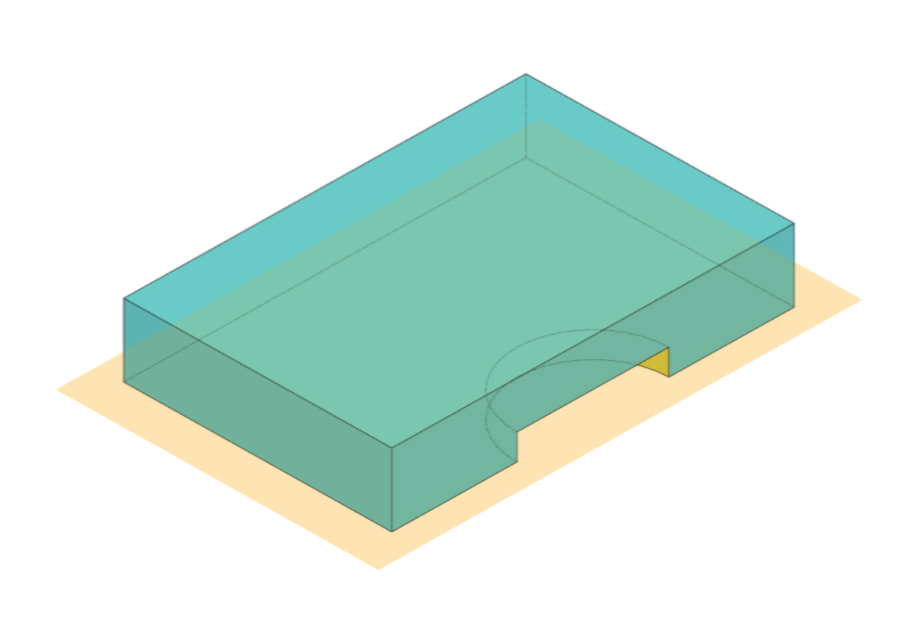}
        &
        \includegraphics[width=\imgwidth\linewidth]{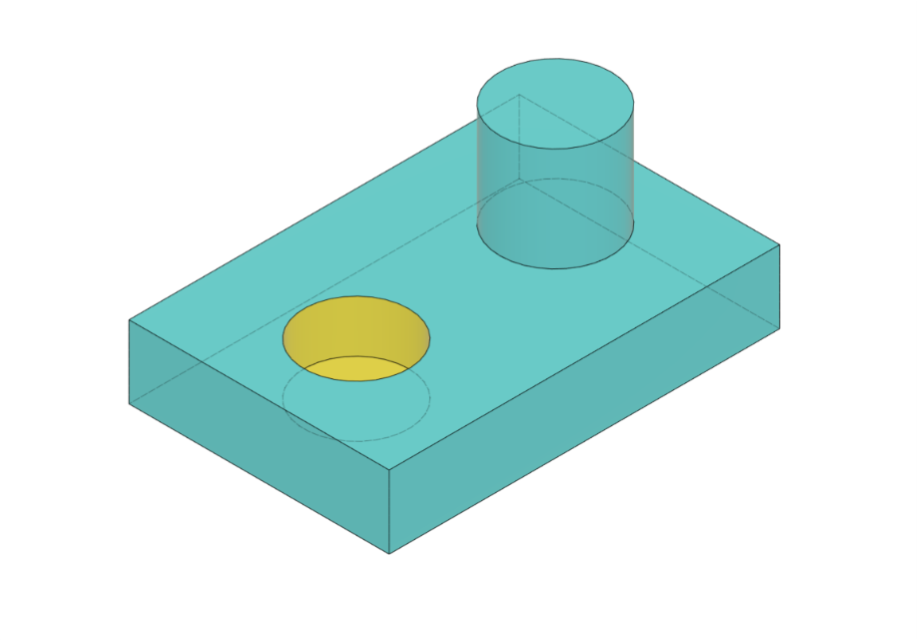}
        &
        \includegraphics[width=\imgwidth\linewidth]{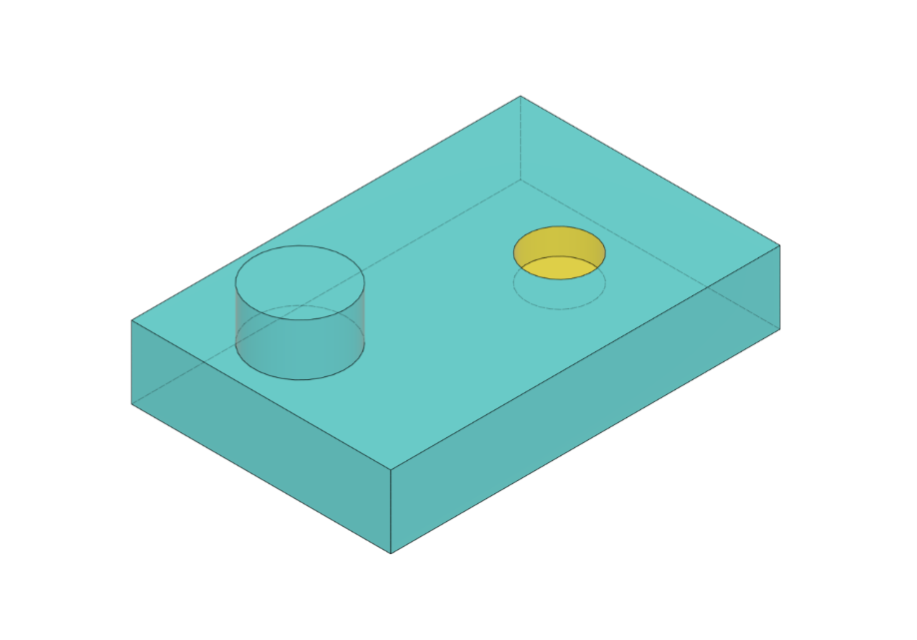}
        &       
        \includegraphics[width=\imgwidth\linewidth]{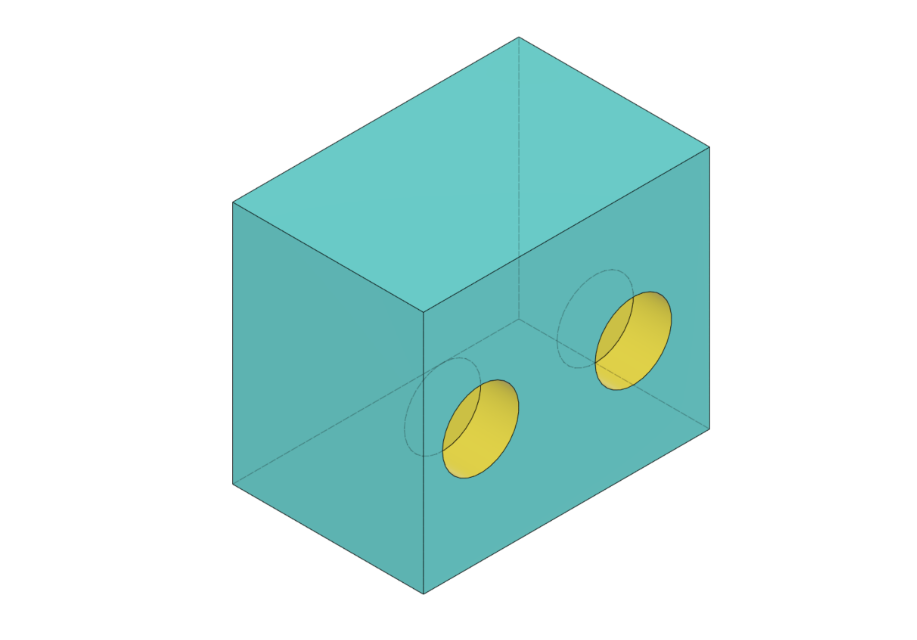}
        \\  
        (g) 1.5\% & (h) 1.2\% & (i) 0.9\% & (j) 0.8\% & (k) 0.4\% & (l) 0.4\%
        \\
        \includegraphics[width=\imgwidth\linewidth]{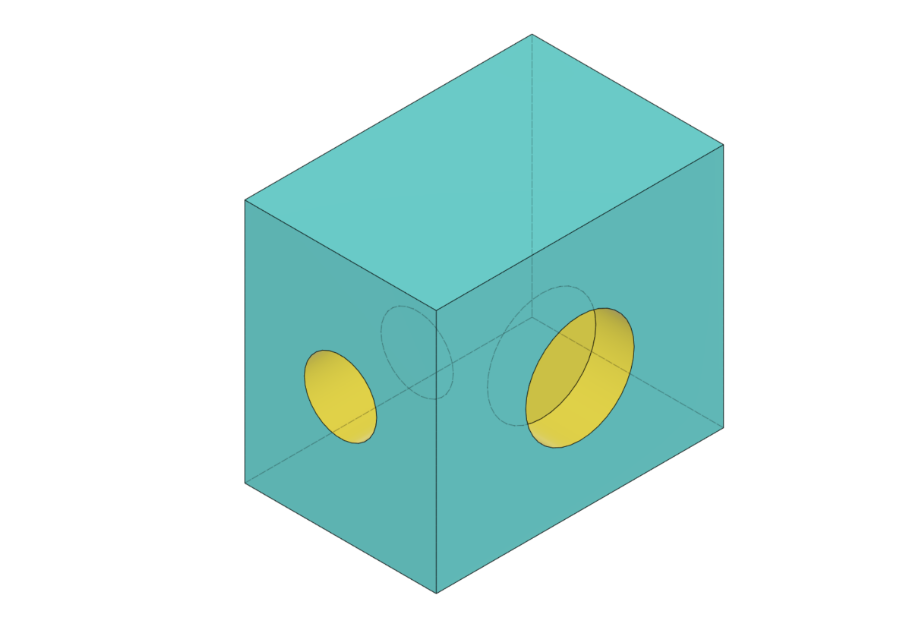}
        &
        \includegraphics[width=\imgwidth\linewidth]{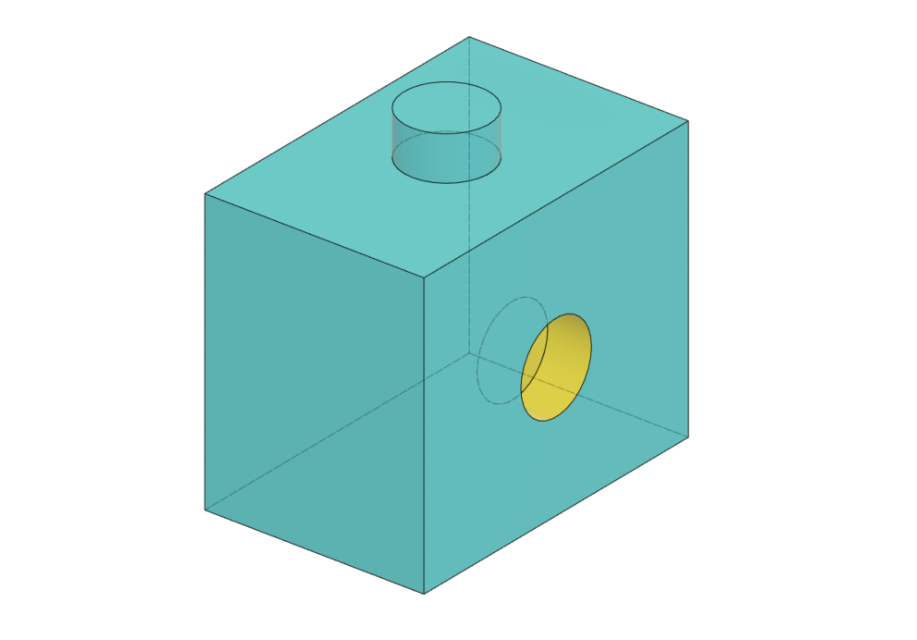}
        &
        \includegraphics[width=\imgwidth\linewidth]{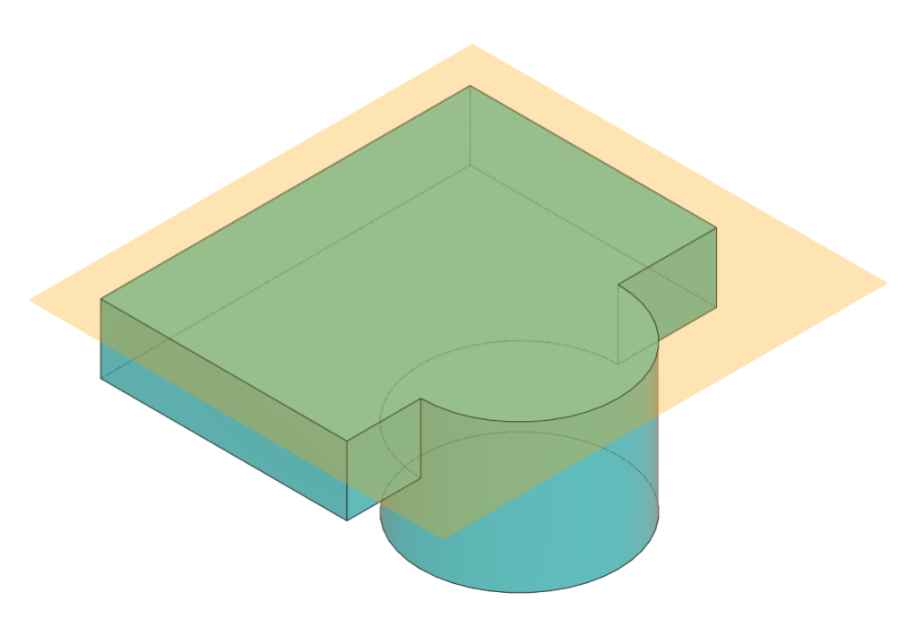}
        &
        \includegraphics[width=\imgwidth\linewidth]{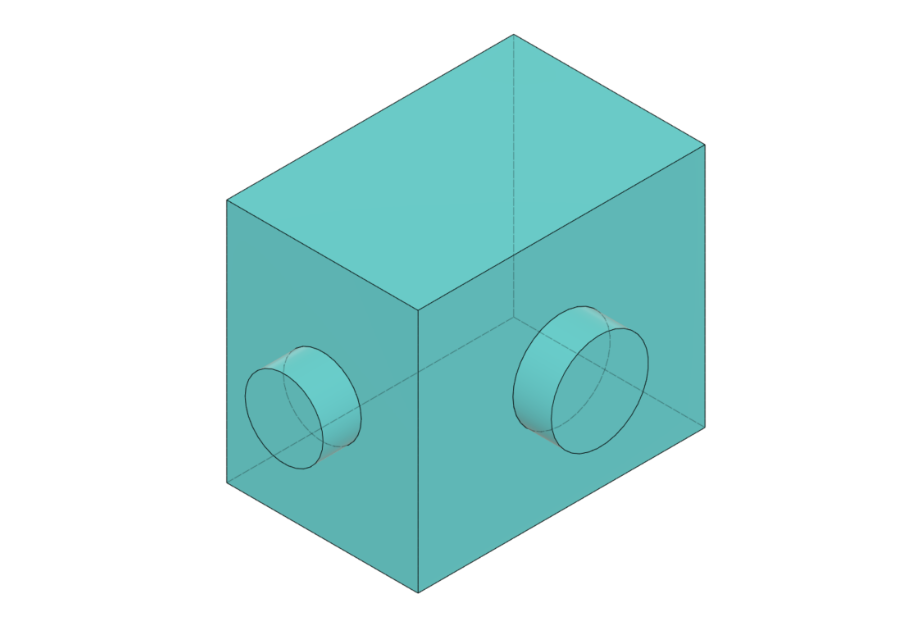}
        &      
        \includegraphics[width=\imgwidth\linewidth]{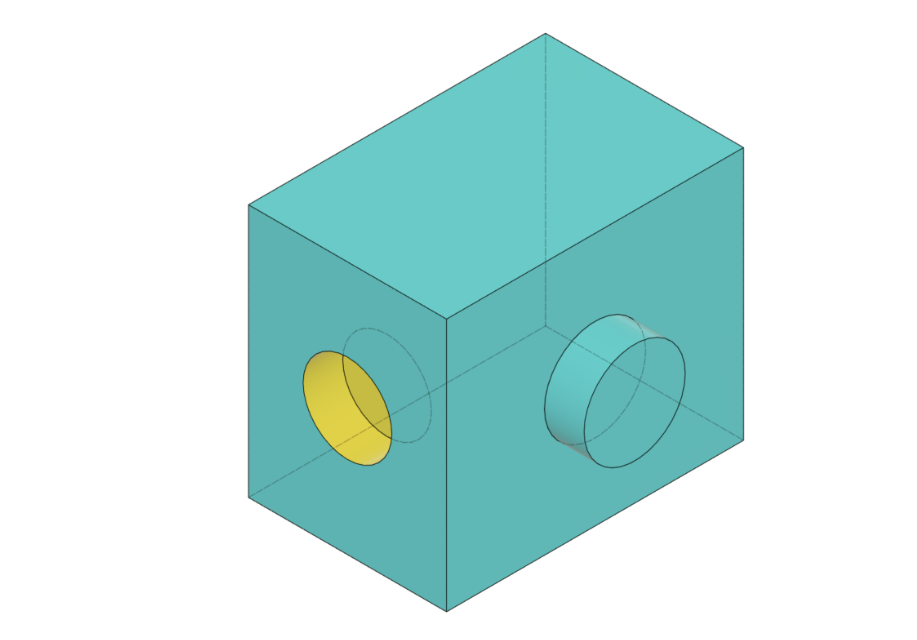}
        &        
        \includegraphics[width=\imgwidth\linewidth]{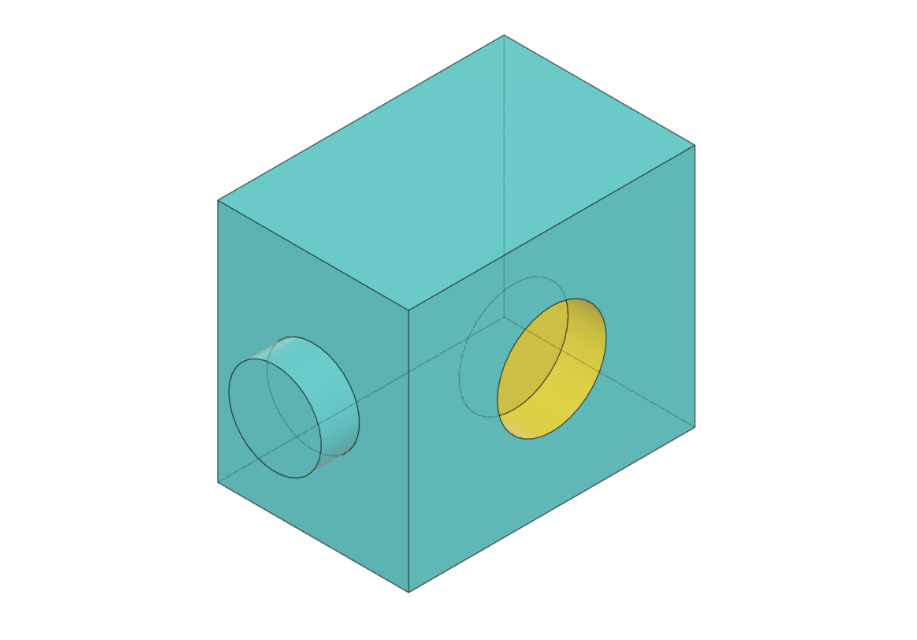}
        \\
        (m) 0.3\% & (n) 0.3\% & (o) 0.2\% & (p) 0.1\% & (q) 0.04\% & (r) 0.033\%
    \end{tabular}
            \caption{The frequency of common combinations of extrusions in the DeepCAD dataset.  Faces resulting from Boolean subtractions are shown in yellow.  Planes shown in orange are constrained to be coincident for pairs of extrusions as described in Section \sectionappendixenvelopearrays.   (a) Single extrusion. (b) Two joined extrusions sharing a common plane. (c) A base extrusion cut by a second extrusion. The end plane of both extrusions is shared.  (d) Two joined extrusions sharing a common start plane. (e) An extrusion with an orthogonal cut.  (f--r) Other combinations of extrusions.}
    \label{figure:decoderheads}
\end{figure*}

\subsection{Combinations of extrusions}
\label{section:decoderheads}
Figure \ref{figure:decoderheads} shows the 18 combinations of extrusion alignments, shared start/end planes and Boolean operations used by decoder modules in this work.  Note that this figure shows extrusions of rectangular and circular profiles as illustrations, while the real dataset utilizes more complex profile shapes.  For pairs of extrusions which have the same extrusion axis and share a common start or end plane (shown in orange in Figure \ref{figure:decoderheads}), the shared plane is encoded as an additional constraint as described in Appendix \sectionappendixenvelopearrays.  
\subsection{DeepCAD profile analysis and selection}
\label{section:appendix_profile_analysis}
We performed the following analysis on the 2D profiles from DeepCAD. Each profile loop was converted into a graph with curves as nodes and edges linking adjacent curves. Node attributes were defined based on curve type and whether lines were aligned with the coordinate system axes.   Horizontal and vertical lines are common in CAD profiles and can be considered as different to lines with arbitrary directions.  Edge attributes where used to indicate whether adjacent curves met with tangent continuity. We then compute the Weisfeiler Lehman graph hash \cite{ShervashidzeSLMB11} for these graphs.  The DeepCAD profile loops can then be organized into groups which have the same hash.  We select constrained parametric sketches which retain the same graph hash as the largest groups over a wide range of parametric variations.   In Figure \ref{figure:profile_shapes} we show 12 of the constrained parametric sketches used in this work, ranked by the size of the group of DeepCAD profiles with the same graph hash.  We see that circles and rectangles dominate the dataset.  The other shapes are frequently produced by other sketch generation works \cite{Ari2020, Ganin2021ComputerAidedDA, para2021sketchgen, SeffVitruvion2021}.  As these sketches are parametric, their parameters can be varied to form a family of geometries.  For example the L-shaped sketch in Figure \ref{figure:profile_shapes}g can be manipulated to build L-shaped profiles with any aspect ratio, while the constraints ensure that all lines remain vertical and horizontal as in the image.  To generate parametric variations of our template sketches we change the parameters in increments, flooding out from the initial parameter values using a priority queue as in Dijkstra's algorithm \cite{Dijkstra59anote}.  Parameter values which result in self-intersections or constraint solver failures do not propagate to further nodes.  The algorithm stops when the priority queue becomes empty or a set number of variations have been successfully generated.  Using this procedure we requested 1000 variations of the rectangular profile and 30 variations of all others except the circles (which have no degrees of freedom beyond center and scale).   This results in 1690 variations which are then used in the  search, retrieval and fitting procedure described in section \sectionsearchretrievalfit.

\begin{figure*}[t]
    \begin{tabular}{ c c c c c c }
        \includegraphics[width=\imgwidth\linewidth]{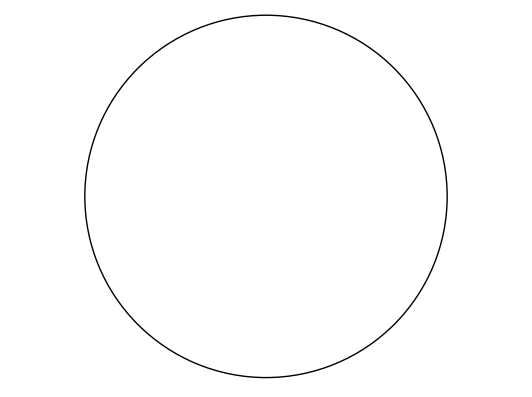}
        &  
        \includegraphics[width=\imgwidth\linewidth]{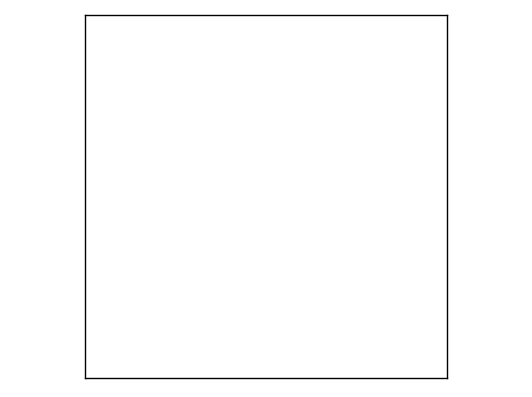}
        & 
        \includegraphics[width=\imgwidth\linewidth]{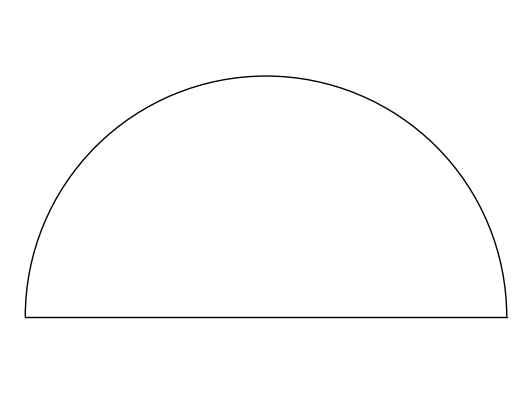}
        & 
        \includegraphics[width=\imgwidth\linewidth]{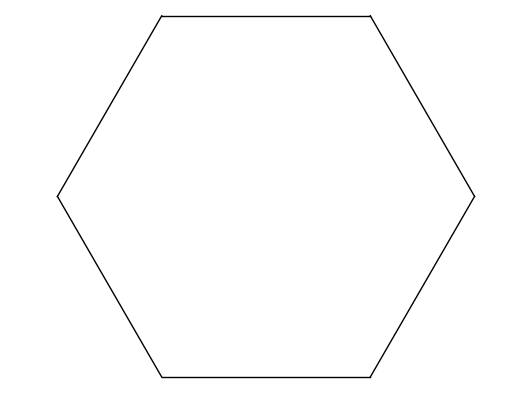}
        &  
        \includegraphics[width=\imgwidth\linewidth]{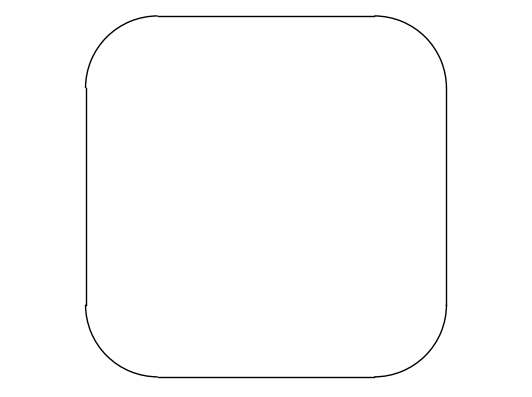}
        & 
        \includegraphics[width=\imgwidth\linewidth]{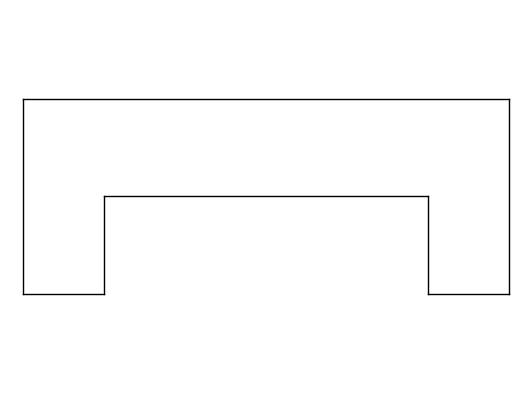}
        \\
        (a) Circle 41.53\% & (b) Rectangle 25.97\% & (c) Crescent 1.50\% & (d) Hexagon 0.95\% & (e) Rounded rect. 0.86\% & (f) U shape 0.78\% \\
                
        \includegraphics[width=\imgwidth\linewidth]{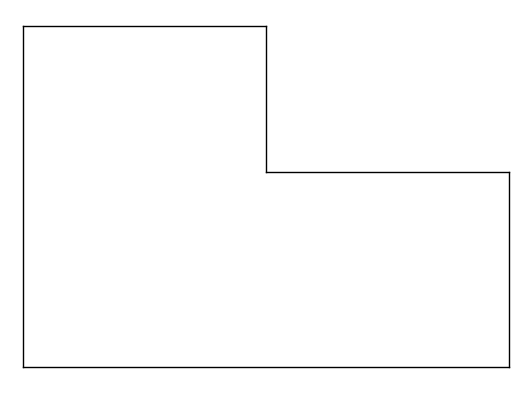}
        &  
        \includegraphics[width=\imgwidth\linewidth]{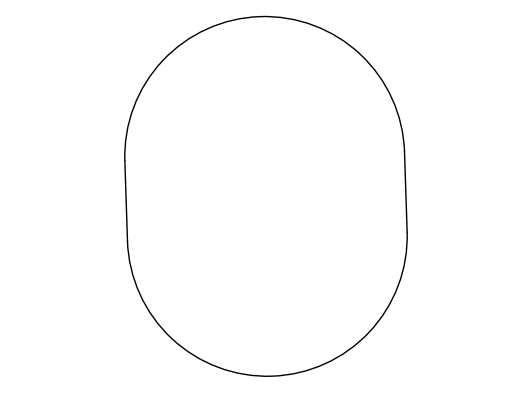}
        & 
        \includegraphics[width=\imgwidth\linewidth]{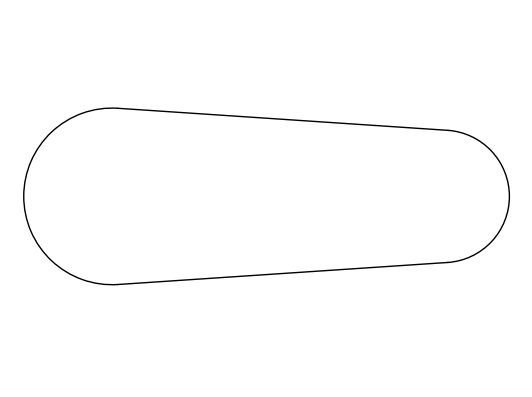}
        & 
        \includegraphics[width=\imgwidth\linewidth]{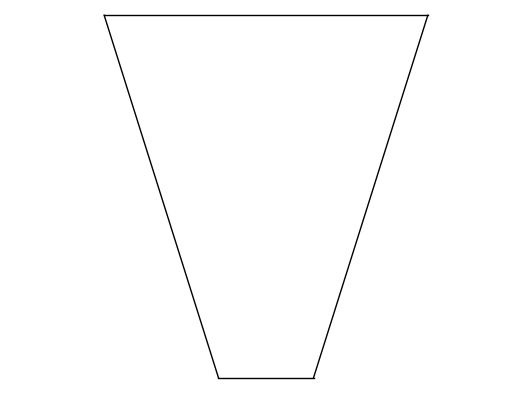}
        &  
        \includegraphics[width=\imgwidth\linewidth]{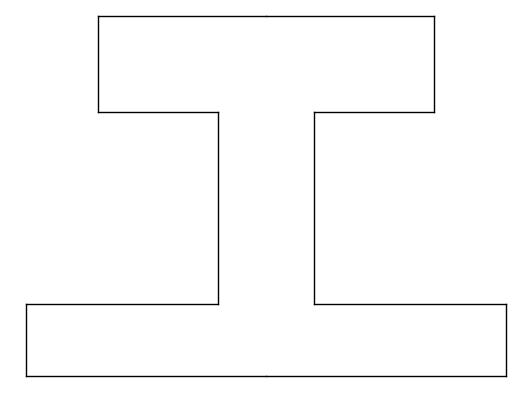}
        & 
        \includegraphics[width=\imgwidth\linewidth]{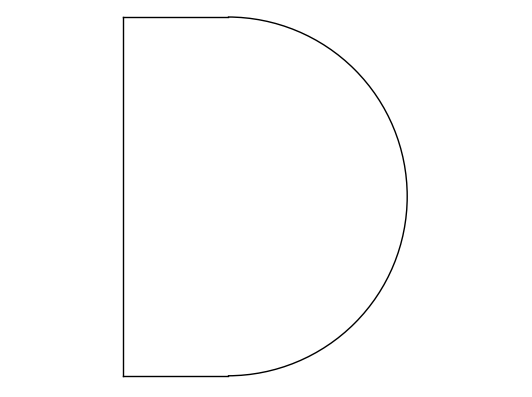}
        \\
        (g) L-shape 0.77\% & (h) Slot 0.75\% & (i) Tapered slot 0.67\% & (j) V-shape 0.67\% & (k) I-shape 0.34\% & (l) D-shape 0.25\% \\
    \end{tabular}
        \caption{Examples of the constrained parametric sketches used to generate profile loops in this work.  The graph hashing procedure described in section \ref{section:appendix_profile_analysis} was used to find profile loops in the DeepCAD dataset which have the same sequence of lines, arcs and important geometric properties (axis alignment, tangent continuity).   The percentage of all DeepCAD profile loops with the same graph hash is shown.}
    \label{figure:profile_shapes}
\end{figure*}
\subsection{Sharing envelope arrays}
\label{section:appendix_envelope_arrays}
\begin{figure}
  \begin{center}
  \end{center}
  \includegraphics[width=0.9\linewidth]{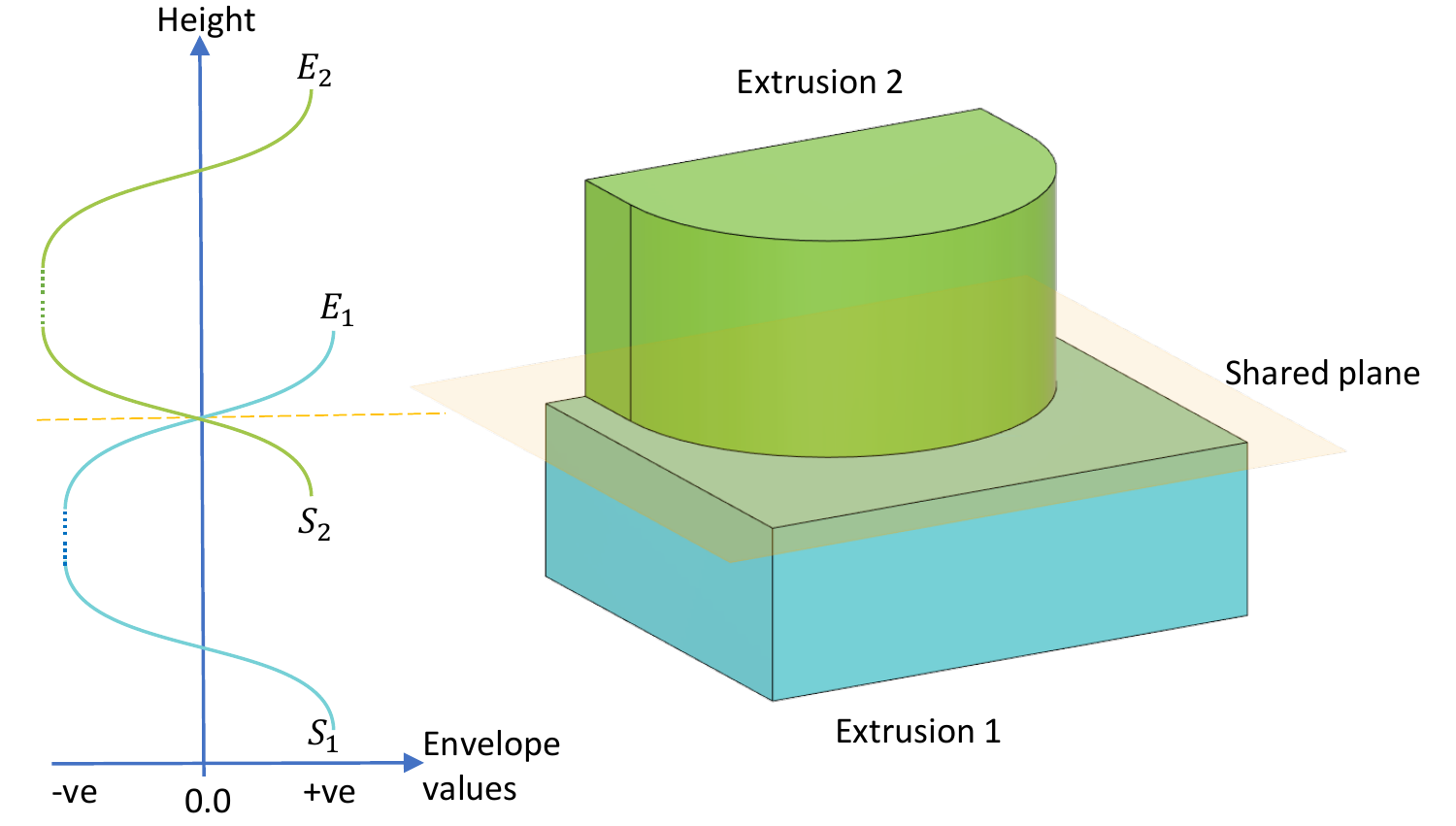}
  \caption{The envelope array values for the two extrusions sharing a common plane in the same configuration as Figure \ref{figure:decoderheads}b.}
  \label{fig:envelope_arrays}
\end{figure}
When creating CAD models from multiple extrusions which share a common plane, it is important that the extrusions start/end exactly at that plane.  Small differences between the end of one extrusion and the start of the next can cause a small gap in the model which is problematic for downstream workflows.  

In this work we carefully design the decoder modules to constrain shared planes to have identical positions for the planes shown in orange in Figure \ref{figure:decoderheads}.   This is done by decoding a single envelope array for the height of the shared plane, and using theses values to build the envelopes for each extrusion abutting the plane.    

Figure \ref{fig:envelope_arrays} shows an example of the envelope arrays for two abutting extrusions.  We use a single envelope decoder in the decoder module which decodes the values of $E_1$.  The values of $S_2$ are then computed as $S_2=-E_1$. 
\subsection{Operation decoder}
\label{section:appendix_operation_decoder}
The linear layers which decode the voxel embedding, $z$, into the individual embeddings for each extrusion have sizes:  128, 512, 512, 128$n$, where $n$ is the number of extrusions.  We use a ReLU non-linearity after each linear layer.   Following the last layer, the output tensor is split into the $n$ extrusion embeddings $e_0, e_1, ..., e_{n-1}$, each with size 128. 
\subsection{2D encoder}
\label{section:appendix_encoder2d}
The 2D CNN encoder used 6 Conv2d layers with leaky ReLU non-linearity.  The kernel size was 4 and the stride was 2.  The channel sizes were 1, 64, 64, 128, 256, 512, 64.
\subsection{2D decoder}
\label{section:appendix_decoder2d}
The decoder used 7 ConvTranspose2d layers, the first of which was a 1x1 conv.  After this the kernel size was 4, the stride was 2.  The channel sizes were 64, 512, 64, 64, 32, 32, 16, 1.   When decoding the 128 length extrusion embeddings, $e_i$, a linear layer was used to map the size down to the length 64 image embeddings used by the decoder. 
\subsection{2D image downsampler}
\label{section:appendix_downsampler}
An advantage of the suggested architecture is that the 2D profile images, generated by the image decoders, can be created with a higher resolution than the voxel grid in which the differentiable extrusions are created.  When working with data where the CAD construction sequence is known, the supervised component of the loss can also operate on the images at this higher resolution.   In this work the decoded profile images have resolution 128x128 and the differentiable extrusions are created with a voxel grid of resolution 64x64x64.   Before a profile is extruded it is downsampled using a Conv2d layer with fixed weights.  The kernel size was 3, stride was 2 and padding 1.  The weights were initialized to 1/3 and not passed to the optimizer.  
\subsection{Envelope decoder}
\label{section:appendix_envelope_decoder}
The 1D envelope decoders are analogous to the 2D image decoders used to create profile images.  They are implemented using 6 ConvTranspose1d layers with ReLU non-linearities between them. The channel sizes were 128, 512, 64, 64, 32, 32, 1.  The first layer had kernel size 1, while subsequent layers used kernel size 4.  The first two layers had a stride of 1 and subsequent layers had stride 2.
\subsection{Training}
\label{section:training}
Both the 2D and 3D autoencoders were trained with the Adam optimizer and a learning rate of $2\mathrm{e}{-4}$.  The 2D autoencoder was trained on 4 Quadro RTX 6000 GPUs and took 1 day 14h to converge. The 3D autoencoder was trained on 3 Quadro RTX 6000s.  The minimum in the validation loss was achieved after 12h.  Inference requires 2.2s per solid on CPU.  The majority of this time is spent in the profile fitting step which could be further optimized using a constraint solver mode which reuses some parts of the initial processing of the constraints to make subsequent solves faster. 
\subsection{Failure cases}
\begin{figure*}[t]
  \begin{center}
  \end{center}
  \includegraphics[width=\linewidth]{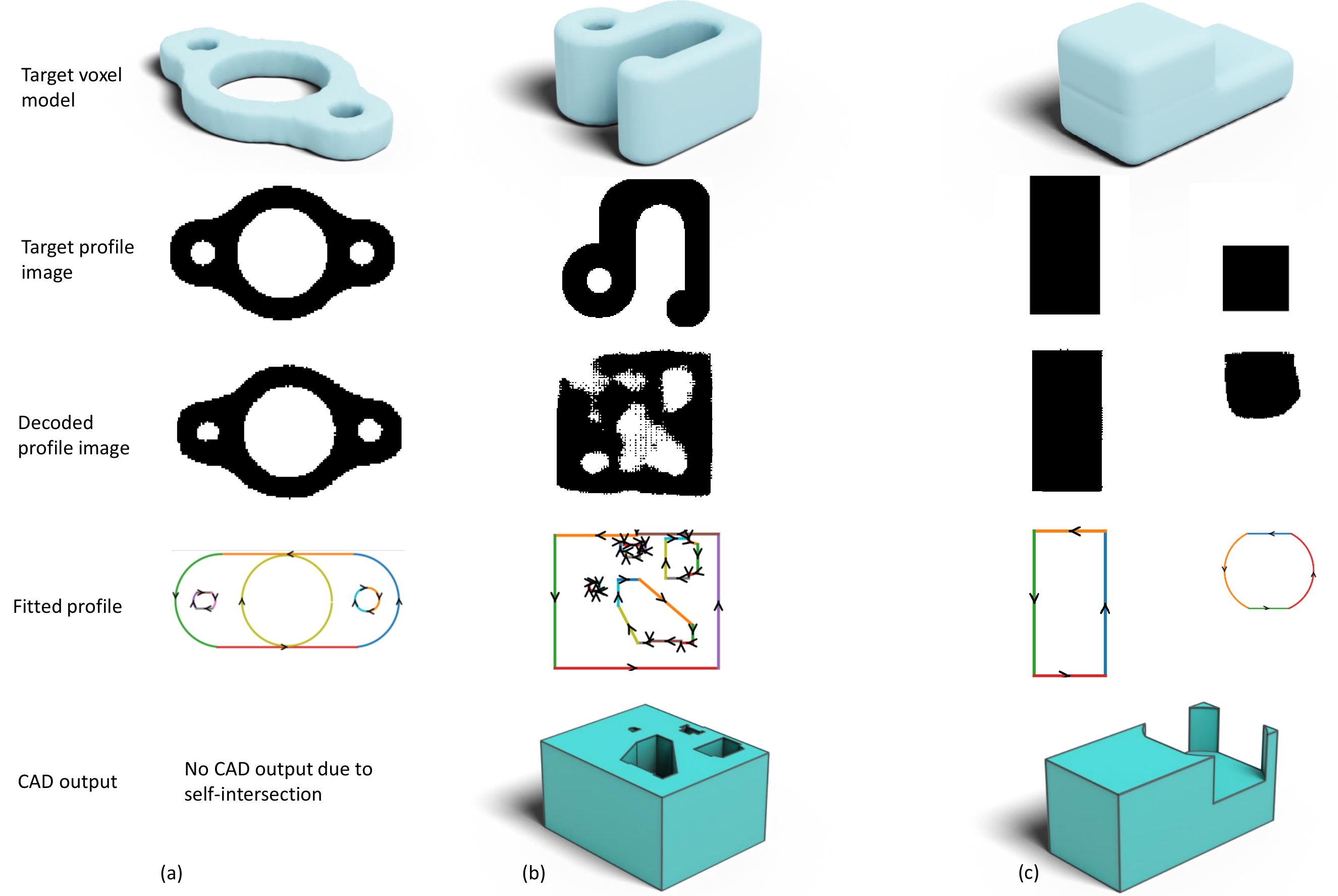}
  \caption{a) Failure mode in which the retrieved profiles self-intersect causing the constructed CAD to be invalid.  b) Failure mode in which the network is unable to decode the correct profile.  c) Failure mode in which misalignment between profiles causes the final solid model to be visually very different from the target.  This usually happens when one extrusion is subtracted from another.}
  \label{fig:failure_cases}
\end{figure*}
In this section we discuss the failure modes for our approach.   There are three common failure modes which we observe in the generated results.  Figure \ref{fig:failure_cases}a shows  a failure mode where the network correctly decodes the profile image, however the search, retrieval and fitting procedure results in profile loops which touch or intersect.  This causes the downstream CAD construction to fail the validity checking and valid CAD output cannot be generated.   Further processing of the 2D curves could detect cases like this and remove intersections, however touching loops as shown in this example would result in a non-manifold B-Rep body if extruded.  

Figure \ref{fig:failure_cases}b shows a case where the network fails to decode the profile with sufficient fidelity and the search, retrieval and fitting procedure produces loops which do not faithfully represent the target shape.  This typically happens for complex profiles.  We have observed that this failure mode is more common when training on a subset of the available data and believe training on larger collections of sketch and extrude data would alleviate this problem.

The third common failure mode is shown in \ref{fig:failure_cases}c.   Here multiple extrusions are combined to build the CAD model and having the correct alignment between the extruded profile is important for the quality of the result.   In this example the shape has almost equal probability of being made as the union of two extrusions ($5.2\%$ of DeepCAD files) or by subtracting one extrusion from the other ($4.5\%$ of DeepCAD files). The ground truth for this specific model was generated as the union of two extrusions, while the decoder module giving the result which most closely approximates the input voxels attempts to create the shape by subtracting a small profile from the larger one.   For the subtraction to remove material as expected, it is important the second profile is not smaller than the target shape.  In this case the fidelity of the second decoded profile image is insufficient for the fitted curves to align with the main profile, resulting in incorrect spikes at the far end of the regenerated CAD.  While the union and subtraction construction sequences are equally valid for designers, automated algorithms may derive some benefit from building models as the union of extrusions where possible.  It is hoped that this insight is useful for researchers attempting to derive canonical CAD sequences from real world data. 

\subsection{2D profile interpolation}
\newcommand{\interpimgwidth}{0.08}
\begin{figure*}[t]
    \begin{tabular}{ c c c c c c c c c c }
    
        \multirow{2}{14mm}[11.7mm]{\includegraphics[width=30.7mm]{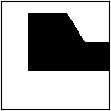}}
        &    
        &    
        \includegraphics[width=\interpimgwidth\linewidth]{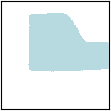}
        &    
        \includegraphics[width=\interpimgwidth\linewidth]{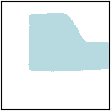}
        &    
        \includegraphics[width=\interpimgwidth\linewidth]{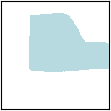}
        &    
        \includegraphics[width=\interpimgwidth\linewidth]{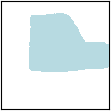}
        &    
        \includegraphics[width=\interpimgwidth\linewidth]{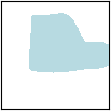}
        &  
        \includegraphics[width=\interpimgwidth\linewidth]{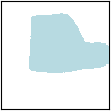}
        &  
        \includegraphics[width=\interpimgwidth\linewidth]{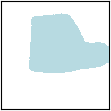}
        &  
        \includegraphics[width=\interpimgwidth\linewidth]{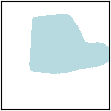}  \bigstrut
        \\
        &    
        &    
        \includegraphics[width=\interpimgwidth\linewidth]{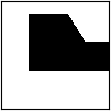}
        &    
        \includegraphics[width=\interpimgwidth\linewidth]{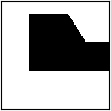}
        &    
        \includegraphics[width=\interpimgwidth\linewidth]{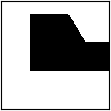}
        &    
        \includegraphics[width=\interpimgwidth\linewidth]{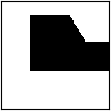}
        &    
        \includegraphics[width=\interpimgwidth\linewidth]{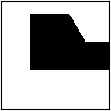}
        &  
        \includegraphics[width=\interpimgwidth\linewidth]{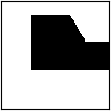}
        &  
        \includegraphics[width=\interpimgwidth\linewidth]{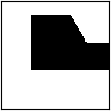}
        &  
        \includegraphics[width=\interpimgwidth\linewidth]{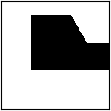}
        \\
        \\

        \includegraphics[width=\interpimgwidth\linewidth]{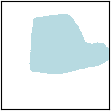}
        &    
        \includegraphics[width=\interpimgwidth\linewidth]{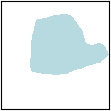}
        &    
        \includegraphics[width=\interpimgwidth\linewidth]{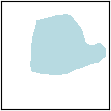}
        &    
        \includegraphics[width=\interpimgwidth\linewidth]{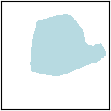}
        &    
        \includegraphics[width=\interpimgwidth\linewidth]{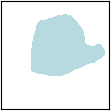}
        &    
        \includegraphics[width=\interpimgwidth\linewidth]{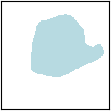}
        &    
        \includegraphics[width=\interpimgwidth\linewidth]{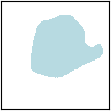}
        &  
        \includegraphics[width=\interpimgwidth\linewidth]{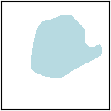}
        &  
        \includegraphics[width=\interpimgwidth\linewidth]{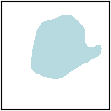}
        &  
        \includegraphics[width=\interpimgwidth\linewidth]{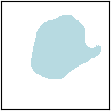}
        \\
        \includegraphics[width=\interpimgwidth\linewidth]{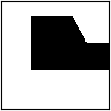}
        &    
        \includegraphics[width=\interpimgwidth\linewidth]{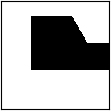}
        &    
        \includegraphics[width=\interpimgwidth\linewidth]{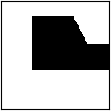}
        &    
        \includegraphics[width=\interpimgwidth\linewidth]{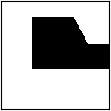}
        &    
        \includegraphics[width=\interpimgwidth\linewidth]{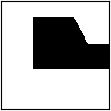}
        &    
        \includegraphics[width=\interpimgwidth\linewidth]{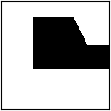}
        &    
        \includegraphics[width=\interpimgwidth\linewidth]{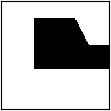}
        &  
        \includegraphics[width=\interpimgwidth\linewidth]{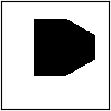}
        &  
        \includegraphics[width=\interpimgwidth\linewidth]{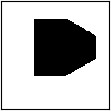}
        &  
        \includegraphics[width=\interpimgwidth\linewidth]{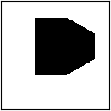}
        \\
        \\

        \includegraphics[width=\interpimgwidth\linewidth]{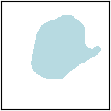}
        &    
        \includegraphics[width=\interpimgwidth\linewidth]{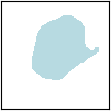}
        &    
        \includegraphics[width=\interpimgwidth\linewidth]{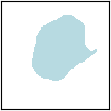}
        &    
        \includegraphics[width=\interpimgwidth\linewidth]{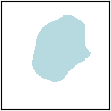}
        &    
        \includegraphics[width=\interpimgwidth\linewidth]{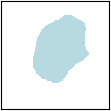}
        &    
        \includegraphics[width=\interpimgwidth\linewidth]{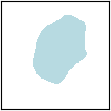}
        &    
        \includegraphics[width=\interpimgwidth\linewidth]{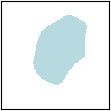}
        &  
        \includegraphics[width=\interpimgwidth\linewidth]{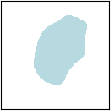}
        &  
        \includegraphics[width=\interpimgwidth\linewidth]{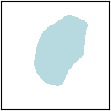}
        &  
        \includegraphics[width=\interpimgwidth\linewidth]{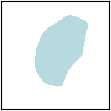}
        \\
        \includegraphics[width=\interpimgwidth\linewidth]{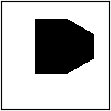}
        &    
        \includegraphics[width=\interpimgwidth\linewidth]{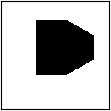}
        &    
        \includegraphics[width=\interpimgwidth\linewidth]{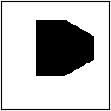}
        &    
        \includegraphics[width=\interpimgwidth\linewidth]{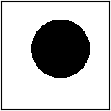}
        &    
        \includegraphics[width=\interpimgwidth\linewidth]{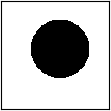}
        &    
        \includegraphics[width=\interpimgwidth\linewidth]{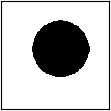}
        &    
        \includegraphics[width=\interpimgwidth\linewidth]{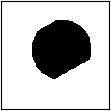}
        &  
        \includegraphics[width=\interpimgwidth\linewidth]{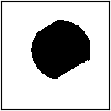}
        &  
        \includegraphics[width=\interpimgwidth\linewidth]{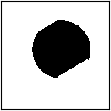}
        &  
        \includegraphics[width=\interpimgwidth\linewidth]{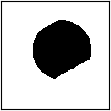}
        \\
        \\

        \includegraphics[width=\interpimgwidth\linewidth]{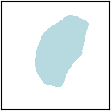}
        &    
        \includegraphics[width=\interpimgwidth\linewidth]{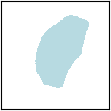}
        &    
        \includegraphics[width=\interpimgwidth\linewidth]{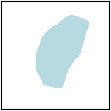}
        &    
        \includegraphics[width=\interpimgwidth\linewidth]{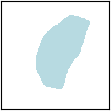}
        &    
        \includegraphics[width=\interpimgwidth\linewidth]{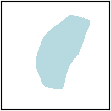}
        &    
        \includegraphics[width=\interpimgwidth\linewidth]{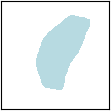}
        &    
        \includegraphics[width=\interpimgwidth\linewidth]{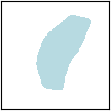}
        &  
        \includegraphics[width=\interpimgwidth\linewidth]{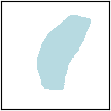}
        &  
        \includegraphics[width=\interpimgwidth\linewidth]{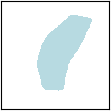}
        &  
        \includegraphics[width=\interpimgwidth\linewidth]{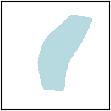}
        \\
        \includegraphics[width=\interpimgwidth\linewidth]{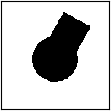}
        &    
        \includegraphics[width=\interpimgwidth\linewidth]{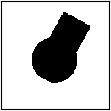}
        &    
        \includegraphics[width=\interpimgwidth\linewidth]{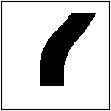}
        &    
        \includegraphics[width=\interpimgwidth\linewidth]{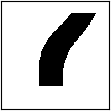}
        &    
        \includegraphics[width=\interpimgwidth\linewidth]{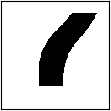}
        &    
        \includegraphics[width=\interpimgwidth\linewidth]{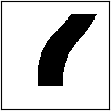}
        &    
        \includegraphics[width=\interpimgwidth\linewidth]{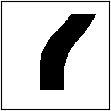}
        &  
        \includegraphics[width=\interpimgwidth\linewidth]{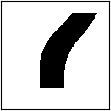}
        &  
        \includegraphics[width=\interpimgwidth\linewidth]{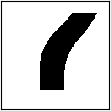}
        &  
        \includegraphics[width=\interpimgwidth\linewidth]{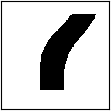}
        \\
        \\

        \includegraphics[width=\interpimgwidth\linewidth]{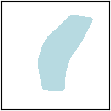}
        &    
        \includegraphics[width=\interpimgwidth\linewidth]{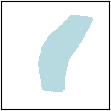}
        &    
        \includegraphics[width=\interpimgwidth\linewidth]{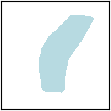}
        &    
        \includegraphics[width=\interpimgwidth\linewidth]{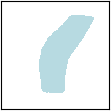}
        &    
        \includegraphics[width=\interpimgwidth\linewidth]{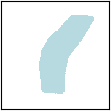}
        &    
        \includegraphics[width=\interpimgwidth\linewidth]{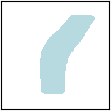}
        &    
        \includegraphics[width=\interpimgwidth\linewidth]{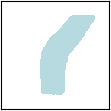}
        &  
        \includegraphics[width=\interpimgwidth\linewidth]{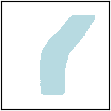}
        &  
        \multirow{2}{11mm}[12.0mm]{\includegraphics[width=30.7mm]{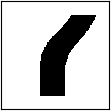}}
        &  
        \\
        \includegraphics[width=\interpimgwidth\linewidth]{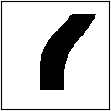}
        &    
        \includegraphics[width=\interpimgwidth\linewidth]{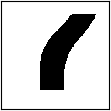}
        &    
        \includegraphics[width=\interpimgwidth\linewidth]{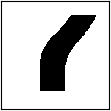}
        &    
        \includegraphics[width=\interpimgwidth\linewidth]{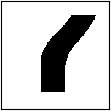}
        &    
        \includegraphics[width=\interpimgwidth\linewidth]{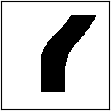}
        &    
        \includegraphics[width=\interpimgwidth\linewidth]{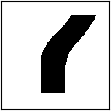}
        &    
        \includegraphics[width=\interpimgwidth\linewidth]{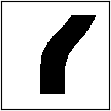}
        &  
        \includegraphics[width=\interpimgwidth\linewidth]{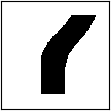}
        &  
        &  
        \\
    \end{tabular}
            \caption{2D profile interpolation.  The shapes of the start and end profiles are show in the top left and bottom right hand sides.  The light blue shapes show the output of the 2D deconvolution decoder and the black shapes are the retrieved and fitted profiles.  Changing the sketch parameters in this way allows the geometry to evolve smoothly, demonstrating the advantage of having a parametric model.  The gradual changes in the sketch geometry is better shown in the video in the supplementary material.}
    \label{fig:2d_interp}
\end{figure*}
\begingroup
\renewcommand{\arraystretch}{0.0}
\begin{figure*}[t]
\begin{tabular}{ c@{\hskip 0in} c@{\hskip 0in} c@{\hskip 0in} c@{\hskip 0.3in}  c@{\hskip 0in} c@{\hskip 0in} c@{\hskip 0in} c@{\hskip 0.3in}  c@{\hskip 0in} c }

    \multicolumn{4}{c}{Simple examples} & \multicolumn{4}{c}{Complex examples} & \multicolumn{2}{c}{DeepCAD failures}\\
    
    \includegraphics[width=\resultimgwidth\linewidth]{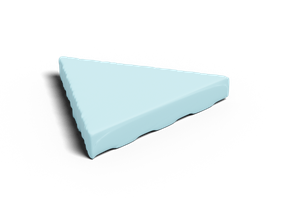} &
    \includegraphics[width=\resultimgwidth\linewidth]{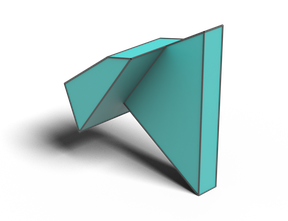} &
    \includegraphics[width=\resultimgwidth\linewidth]{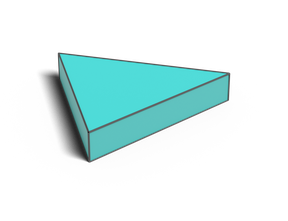} &
    \includegraphics[width=\resultimgwidth\linewidth]{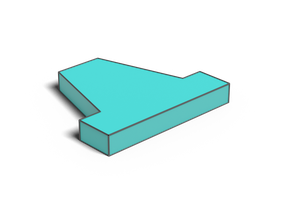} &

    \includegraphics[width=\resultimgwidth\linewidth]{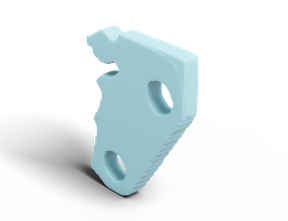} &
    \includegraphics[width=\resultimgwidth\linewidth]{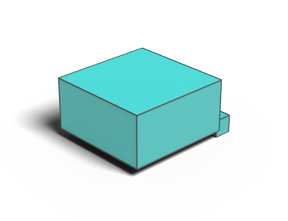} &
    \includegraphics[width=\resultimgwidth\linewidth]{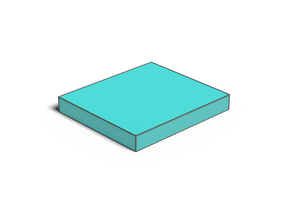} &
    \includegraphics[width=\resultimgwidth\linewidth]{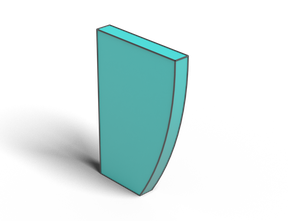} &

    \includegraphics[width=\resultimgwidth\linewidth]{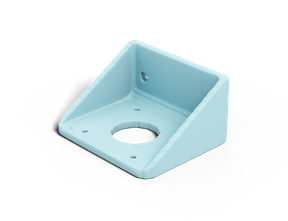} &
    \includegraphics[width=\resultimgwidth\linewidth]{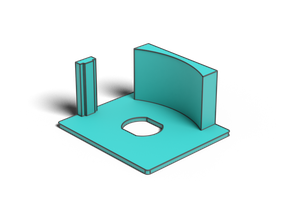} \\

    \includegraphics[width=\resultimgwidth\linewidth]{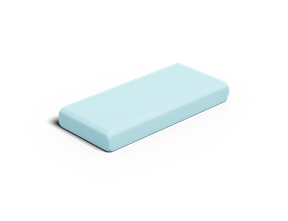} &
    \includegraphics[width=\resultimgwidth\linewidth]{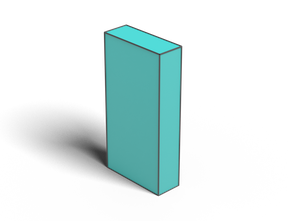} &
    \includegraphics[width=\resultimgwidth\linewidth]{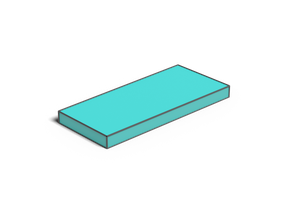} &
    \includegraphics[width=\resultimgwidth\linewidth]{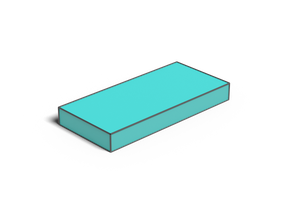} &
    
    \includegraphics[width=\resultimgwidth\linewidth]{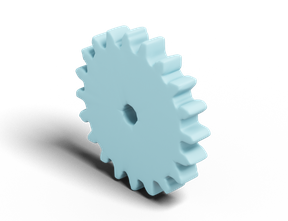} &
    \includegraphics[width=\resultimgwidth\linewidth]{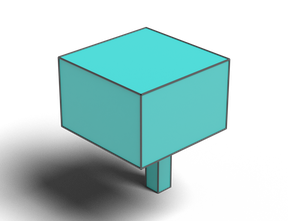} &
    \includegraphics[width=\resultimgwidth\linewidth]{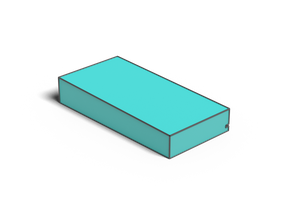} &
    \includegraphics[width=\resultimgwidth\linewidth]{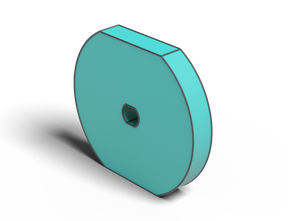} &
    
    \includegraphics[width=\resultimgwidth\linewidth]{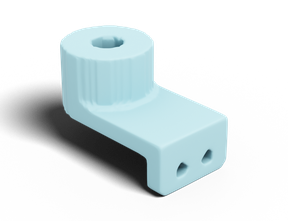} &
    \includegraphics[width=\resultimgwidth\linewidth]{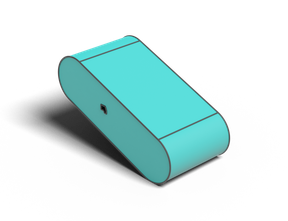} \\

    \includegraphics[width=\resultimgwidth\linewidth]{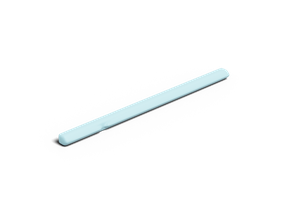} &
    \includegraphics[width=\resultimgwidth\linewidth]{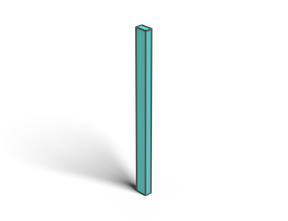} &
    \includegraphics[width=\resultimgwidth\linewidth]{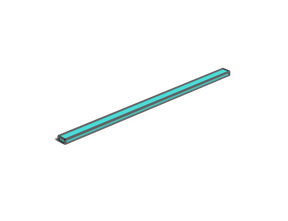} &
    \includegraphics[width=\resultimgwidth\linewidth]{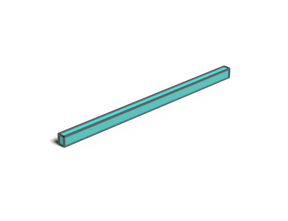} &

    \includegraphics[width=\resultimgwidth\linewidth]{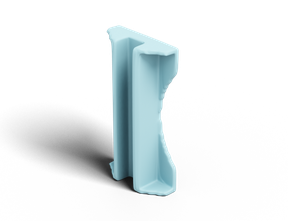} &
    \includegraphics[width=\resultimgwidth\linewidth]{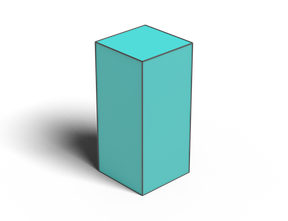} &
    \includegraphics[width=\resultimgwidth\linewidth]{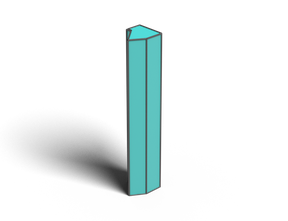} &
    \includegraphics[width=\resultimgwidth\linewidth]{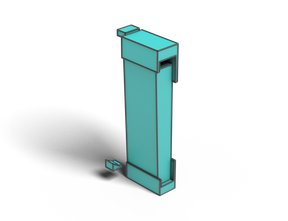} &
    
    \includegraphics[width=\resultimgwidth\linewidth]{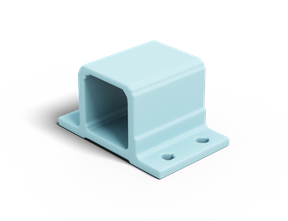} &
    \includegraphics[width=\resultimgwidth\linewidth]{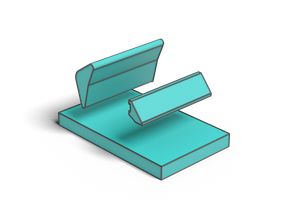} \\
    
    \includegraphics[width=\resultimgwidth\linewidth]{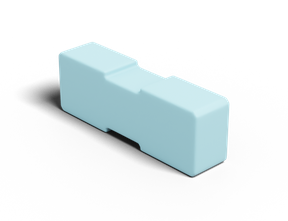} &
    \includegraphics[width=\resultimgwidth\linewidth]{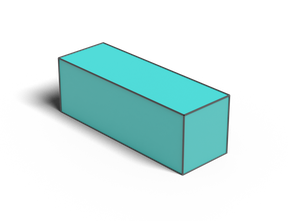} &
    \includegraphics[width=\resultimgwidth\linewidth]{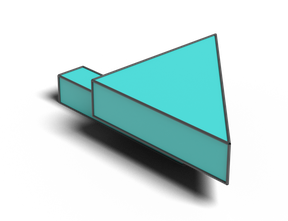} &
    \includegraphics[width=\resultimgwidth\linewidth]{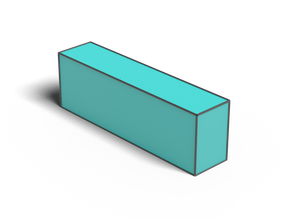} &

    \includegraphics[width=\resultimgwidth\linewidth]{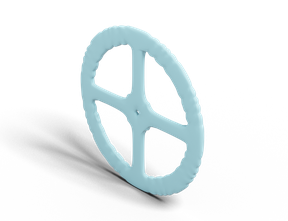} &
    \includegraphics[width=\resultimgwidth\linewidth]{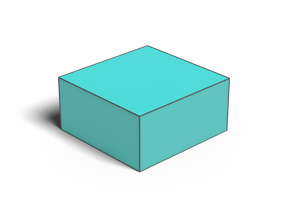} &
    \includegraphics[width=\resultimgwidth\linewidth]{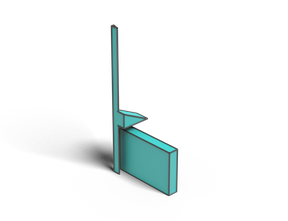} &
    \includegraphics[width=\resultimgwidth\linewidth]{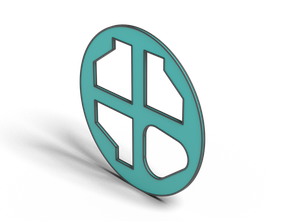} &
    
    \includegraphics[width=\resultimgwidth\linewidth]{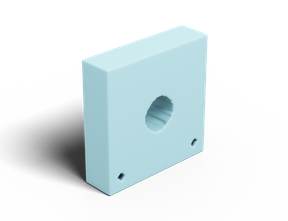} &
    \includegraphics[width=\resultimgwidth\linewidth]{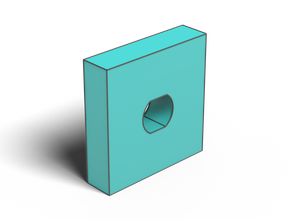} \\
    
    \includegraphics[width=\resultimgwidth\linewidth]{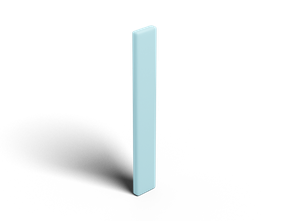} &
    \includegraphics[width=\resultimgwidth\linewidth]{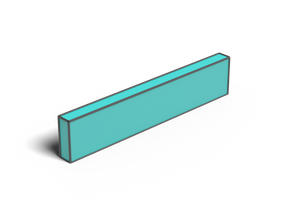} &
    \includegraphics[width=\resultimgwidth\linewidth]{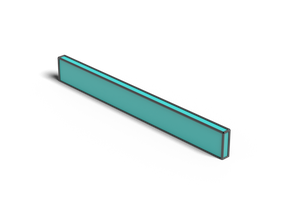} &
    \includegraphics[width=\resultimgwidth\linewidth]{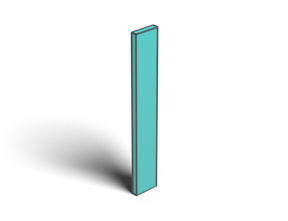} &

    \includegraphics[width=\resultimgwidth\linewidth]{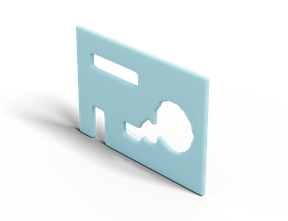} &
    \includegraphics[width=\resultimgwidth\linewidth]{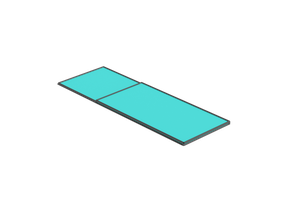} &
    \includegraphics[width=\resultimgwidth\linewidth]{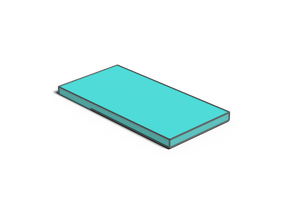} &
    \includegraphics[width=\resultimgwidth\linewidth]{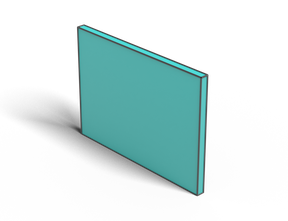} &
    
    \includegraphics[width=\resultimgwidth\linewidth]{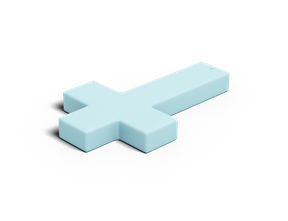} &
    \includegraphics[width=\resultimgwidth\linewidth]{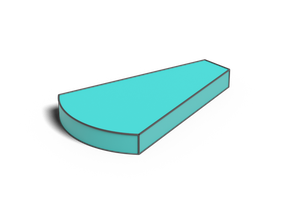} \\
    
    \includegraphics[width=\resultimgwidth\linewidth]{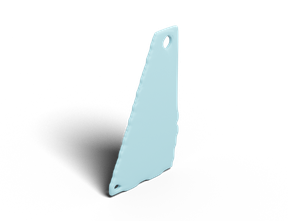} &
    \includegraphics[width=\resultimgwidth\linewidth]{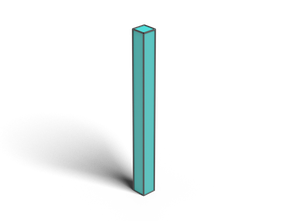} &
    \includegraphics[width=\resultimgwidth\linewidth]{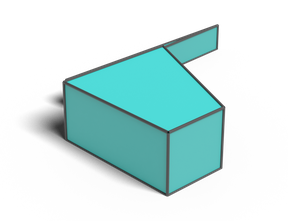} &
    \includegraphics[width=\resultimgwidth\linewidth]{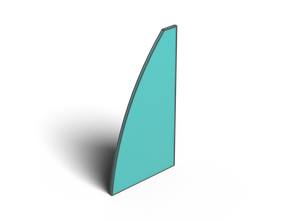} &

    \includegraphics[width=\resultimgwidth\linewidth]{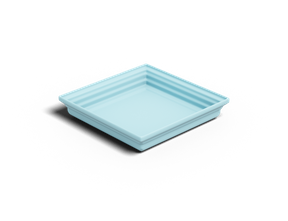} &
    \includegraphics[width=\resultimgwidth\linewidth]{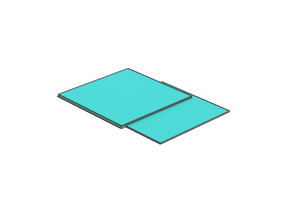} &
    \includegraphics[width=\resultimgwidth\linewidth]{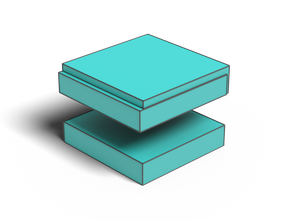} &
    \includegraphics[width=\resultimgwidth\linewidth]{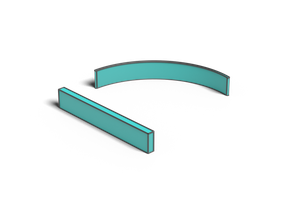} &
    
    \includegraphics[width=\resultimgwidth\linewidth]{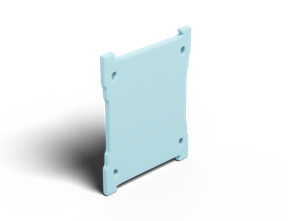} &
    \includegraphics[width=\resultimgwidth\linewidth]{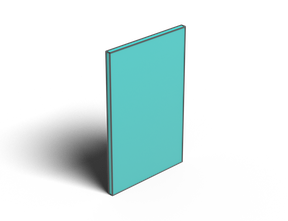} \\
    
    \includegraphics[width=\resultimgwidth\linewidth]{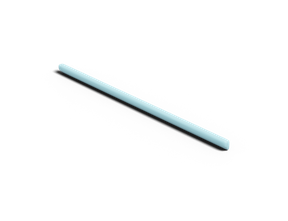} &
    \includegraphics[width=\resultimgwidth\linewidth]{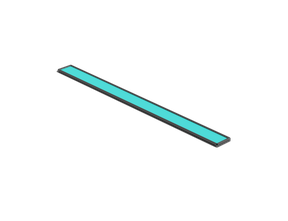} &
    \includegraphics[width=\resultimgwidth\linewidth]{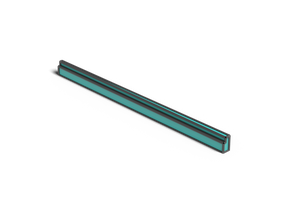} &
    \includegraphics[width=\resultimgwidth\linewidth]{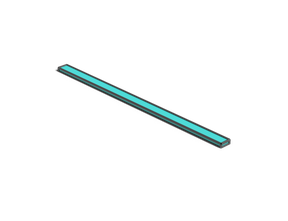} &

    \includegraphics[width=\resultimgwidth\linewidth]{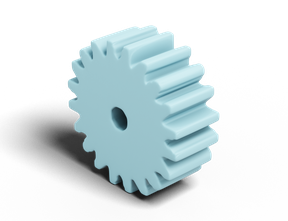} &
    \includegraphics[width=\resultimgwidth\linewidth]{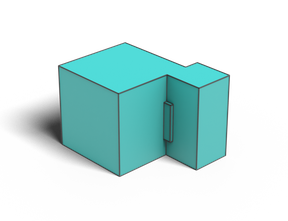} &
    \includegraphics[width=\resultimgwidth\linewidth]{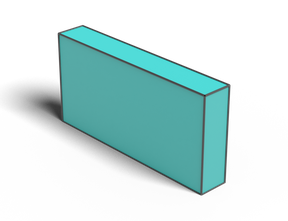} &
    \includegraphics[width=\resultimgwidth\linewidth]{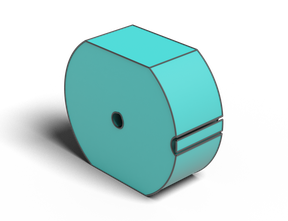} &
    
    \includegraphics[width=\resultimgwidth\linewidth]{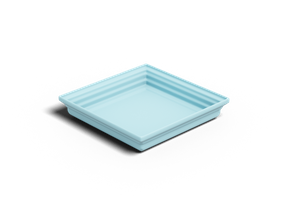} &
    \includegraphics[width=\resultimgwidth\linewidth]{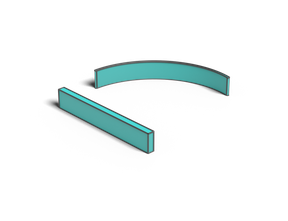} \\
    
    \includegraphics[width=\resultimgwidth\linewidth]{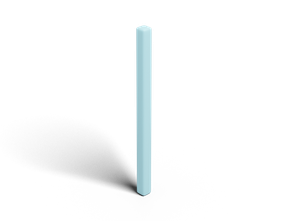} &
    \includegraphics[width=\resultimgwidth\linewidth]{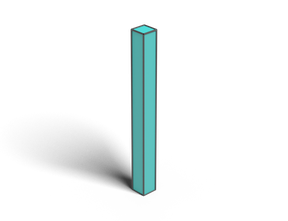} &
    \includegraphics[width=\resultimgwidth\linewidth]{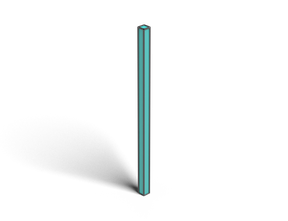} &
    \includegraphics[width=\resultimgwidth\linewidth]{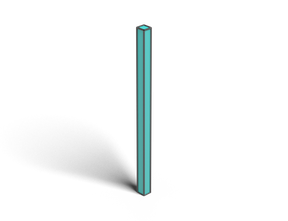} &

    \includegraphics[width=\resultimgwidth\linewidth]{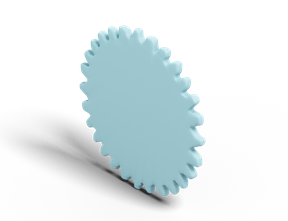} &
    \includegraphics[width=\resultimgwidth\linewidth]{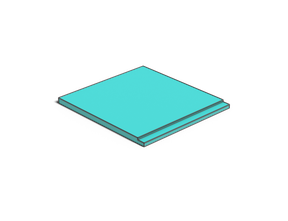} &
    \includegraphics[width=\resultimgwidth\linewidth]{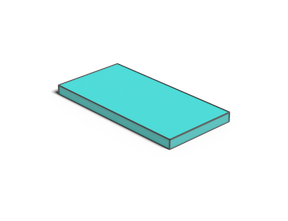} &
    \includegraphics[width=\resultimgwidth\linewidth]{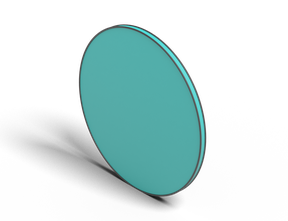} &
    
    \includegraphics[width=\resultimgwidth\linewidth]{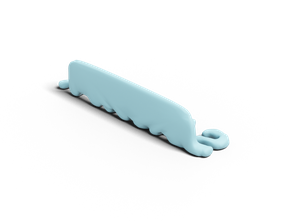} &
    \includegraphics[width=\resultimgwidth\linewidth]{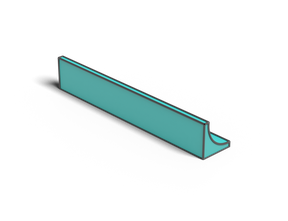} \\
    
    \includegraphics[width=\resultimgwidth\linewidth]{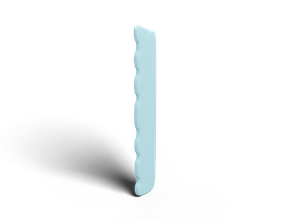} &
    \includegraphics[width=\resultimgwidth\linewidth]{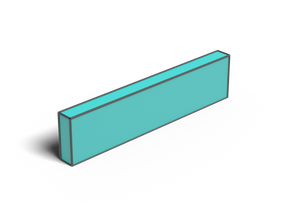} &
    \includegraphics[width=\resultimgwidth\linewidth]{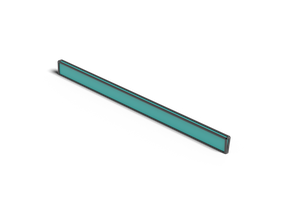} &
    \includegraphics[width=\resultimgwidth\linewidth]{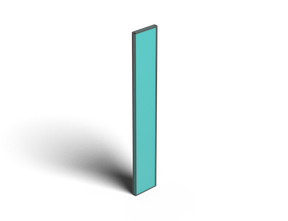} &

    \includegraphics[width=\resultimgwidth\linewidth]{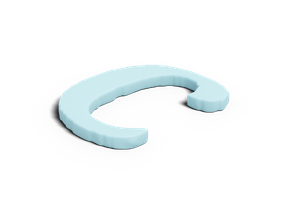} &
    \includegraphics[width=\resultimgwidth\linewidth]{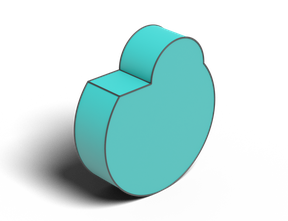} &
    \includegraphics[width=\resultimgwidth\linewidth]{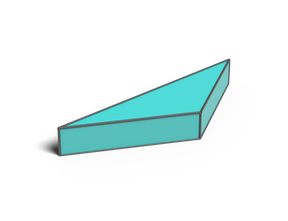} &
    \includegraphics[width=\resultimgwidth\linewidth]{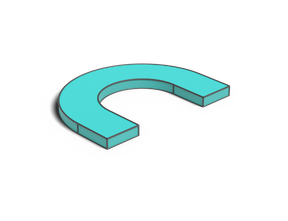} &
    
    \includegraphics[width=\resultimgwidth\linewidth]{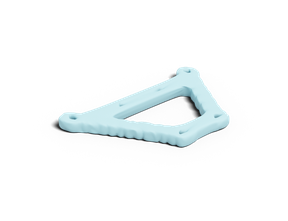} &
    \includegraphics[width=\resultimgwidth\linewidth]{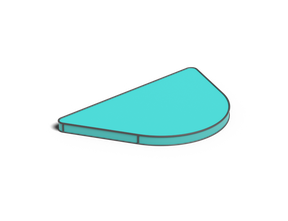} \\
    
    \includegraphics[width=\resultimgwidth\linewidth]{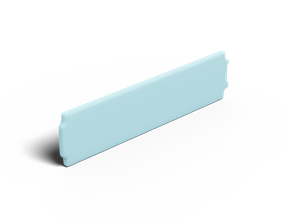} &
    \includegraphics[width=\resultimgwidth\linewidth]{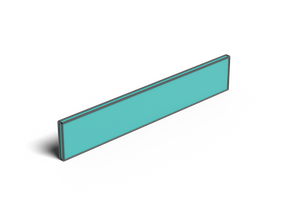} &
    \includegraphics[width=\resultimgwidth\linewidth]{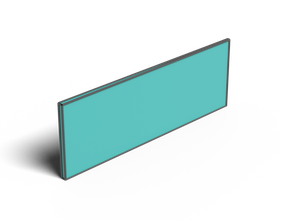} &
    \includegraphics[width=\resultimgwidth\linewidth]{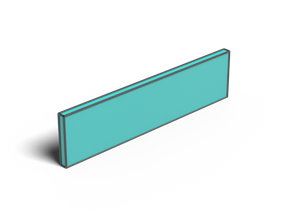} &

    \includegraphics[width=\resultimgwidth\linewidth]{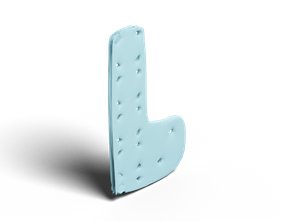} &
    \includegraphics[width=\resultimgwidth\linewidth]{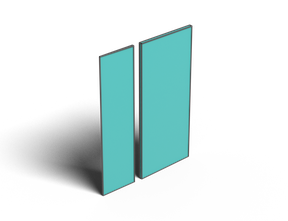} &
    \includegraphics[width=\resultimgwidth\linewidth]{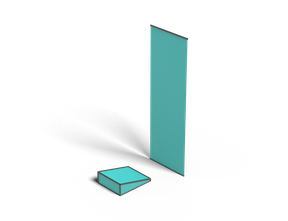} &
    \includegraphics[width=\resultimgwidth\linewidth]{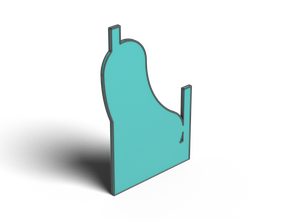} &
    
    \includegraphics[width=\resultimgwidth\linewidth]{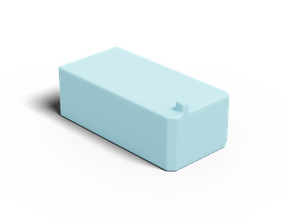} &
    \includegraphics[width=\resultimgwidth\linewidth]{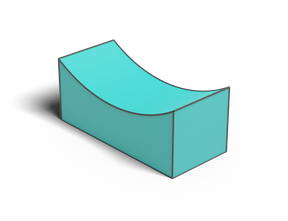} \\
    
    \includegraphics[width=\resultimgwidth\linewidth]{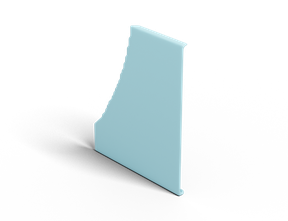} &
    \includegraphics[width=\resultimgwidth\linewidth]{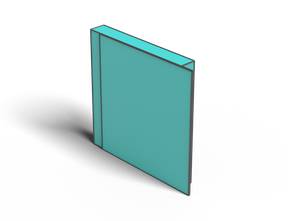} &
    \includegraphics[width=\resultimgwidth\linewidth]{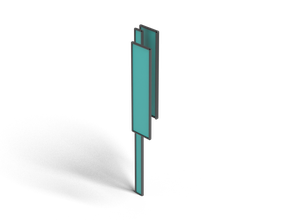} &
    \includegraphics[width=\resultimgwidth\linewidth]{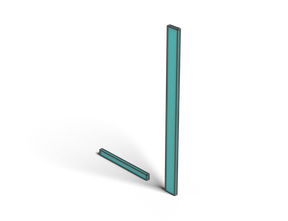} &

    \includegraphics[width=\resultimgwidth\linewidth]{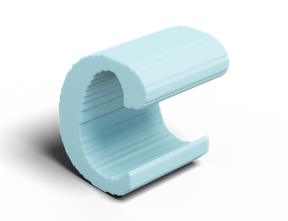} &
    \includegraphics[width=\resultimgwidth\linewidth]{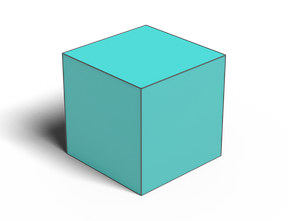} &
    \includegraphics[width=\resultimgwidth\linewidth]{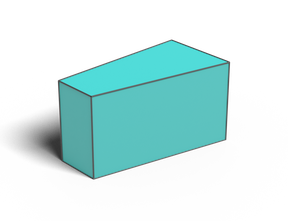} &
    \includegraphics[width=\resultimgwidth\linewidth]{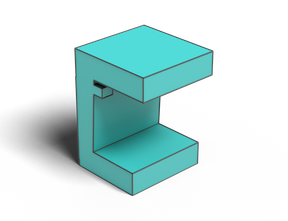} &
    
    \includegraphics[width=\resultimgwidth\linewidth]{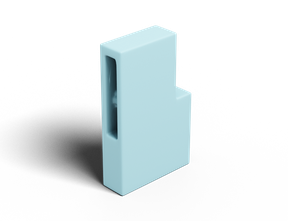} &
    \includegraphics[width=\resultimgwidth\linewidth]{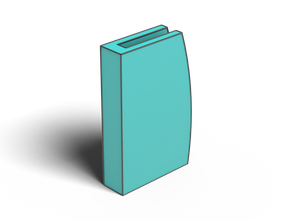} \\
    
    \includegraphics[width=\resultimgwidth\linewidth]{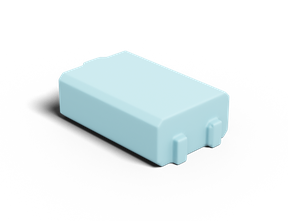} &
    \includegraphics[width=\resultimgwidth\linewidth]{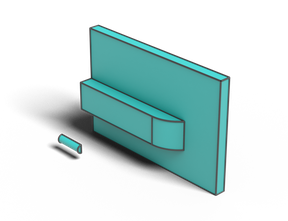} &
    \includegraphics[width=\resultimgwidth\linewidth]{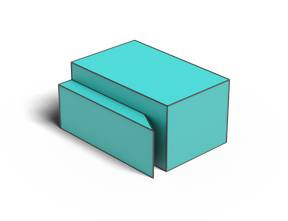} &
    \includegraphics[width=\resultimgwidth\linewidth]{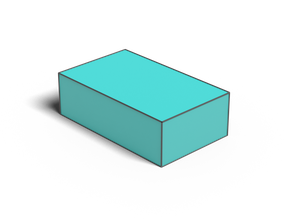} &

    \includegraphics[width=\resultimgwidth\linewidth]{figures/abc_experiment/all_methods2/00486647_7dc5b279f9f6e964add61af4_step_000_6_voxel.png} &
    \includegraphics[width=\resultimgwidth\linewidth]{figures/abc_experiment/all_methods2/00486647_7dc5b279f9f6e964add61af4_step_000_6_deepcad_pc.png} &
    \includegraphics[width=\resultimgwidth\linewidth]{figures/abc_experiment/all_methods2/00486647_7dc5b279f9f6e964add61af4_step_000_6_deepcad_voxel.png} &
    \includegraphics[width=\resultimgwidth\linewidth]{figures/abc_experiment/all_methods2/00486647_7dc5b279f9f6e964add61af4_step_000_6_ours.png} &
    
    \includegraphics[width=\resultimgwidth\linewidth]{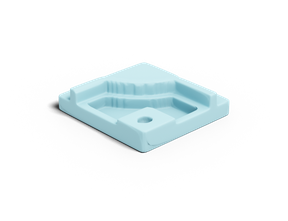} &
    \includegraphics[width=\resultimgwidth\linewidth]{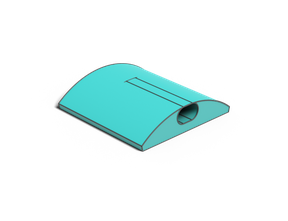} \\
    
    \includegraphics[width=\resultimgwidth\linewidth]{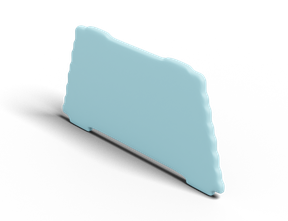} &
    \includegraphics[width=\resultimgwidth\linewidth]{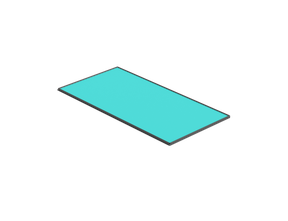} &
    \includegraphics[width=\resultimgwidth\linewidth]{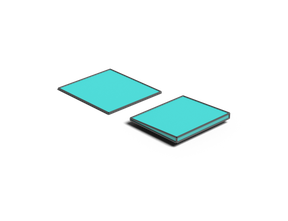} &
    \includegraphics[width=\resultimgwidth\linewidth]{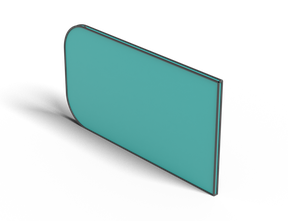} &

    \includegraphics[width=\resultimgwidth\linewidth]{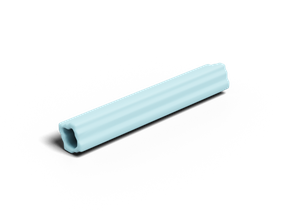} &
    \includegraphics[width=\resultimgwidth\linewidth]{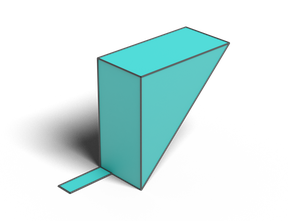} &
    \includegraphics[width=\resultimgwidth\linewidth]{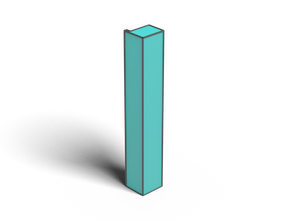} &
    \includegraphics[width=\resultimgwidth\linewidth]{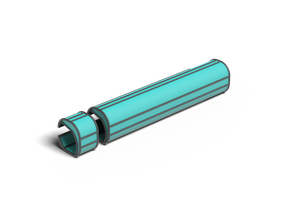} &
    
    \includegraphics[width=\resultimgwidth\linewidth]{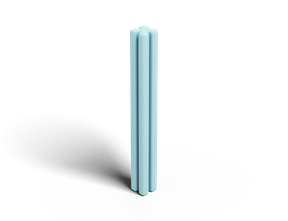} &
    \includegraphics[width=\resultimgwidth\linewidth]{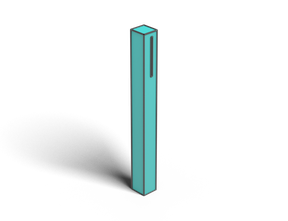} \\
    
    
    \includegraphics[width=\resultimgwidth\linewidth]{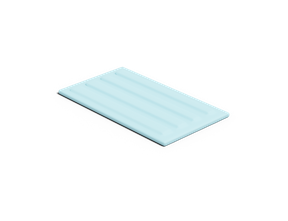} &
    \includegraphics[width=\resultimgwidth\linewidth]{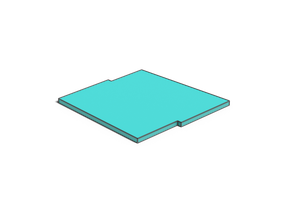} &
    \includegraphics[width=\resultimgwidth\linewidth]{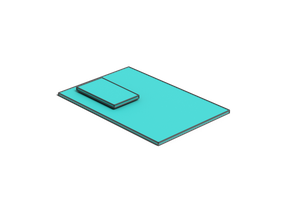} &
    \includegraphics[width=\resultimgwidth\linewidth]{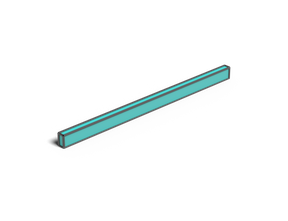} &

    \includegraphics[width=\resultimgwidth\linewidth]{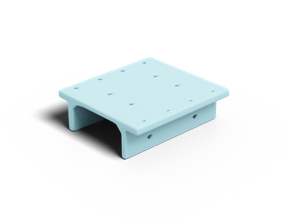} &
    \includegraphics[width=\resultimgwidth\linewidth]{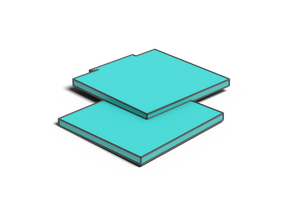} &
    \includegraphics[width=\resultimgwidth\linewidth]{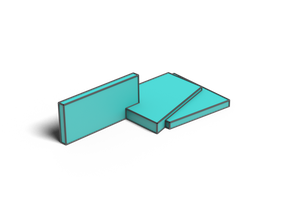} &
    \includegraphics[width=\resultimgwidth\linewidth]{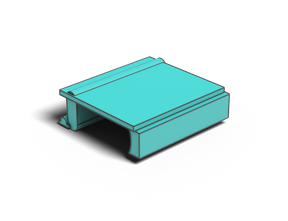} &
    
    \includegraphics[width=\resultimgwidth\linewidth]{figures/abc_experiment/dc_fails2/00633548_58cfbbfe334b731002b154a1_step_010_1_voxel.png} &
    \includegraphics[width=\resultimgwidth\linewidth]{figures/abc_experiment/dc_fails2/00633548_58cfbbfe334b731002b154a1_step_010_1_ours.png} \\
    
       (a) Target & (b) DC-P & (c) DC-V & (d) Ours & (a) Target & (b) DC-P & (c) DC-V & (d) Ours & (a) Target & (d) Ours
    \end{tabular}

\caption{Reconstructed models using voxelized solids from the ABC dataset as the target geometry.  The solids were sampled at random from those created during the experiment described in Section \sectionreconstructionthreed.   On the left we show models which all methods could create for simple target shapes.  In the middle we show more complex examples.  On the right we show examples where DeepCAD failed to generate valid solids, but our method was able to do so successfully.     (a) Target voxel model.  (b) DeepCAD using the point cloud encoder.  (c) DeepCAD using the voxel encoder.  (d) Ours}
\label{figure:abc_experiment}
\end{figure*}
\endgroup
The search, retrieval and fitting algorithm can also be used to interpolate between different CAD profiles.  Given a start and end profile, these can be embedded using the CNN encoder described in Appendix  \sectionappendixencodertwod.  The embeddings of the profiles, $z_{start}$ and $z_{end}$ can then be linearly interpolated
\begin{equation}
    z(t) = (1-t) ~z_{start} + t~z_{end}
\end{equation}
where the parameter $t$ is in the range $[0,1]$.   For any intermediate value of $z(t)$, an approximate shape can then be decoded using the deconvolution decoder described in Appendix \sectionappendixdecodertwod.  These decoded shapes are shown in light blue in Figure \ref{fig:2d_interp}.  While they produce a smooth interpolation between the start and end shapes, the geometry becomes rounded and distorted and the intermediate shapes do not look like geometry which a CAD designer would create. 

To recover CAD-like shapes we first create an embedding for each variation of the sketch templates described in Appendix \sectionappendixprofileanalysis.   This gives a set of embeddings $Z=\{z_0, z_1, ... z_{1689}\}$.  We then choose the template variation such that $|z_i-z(t)|$ is the minimum for any $z_i \in Z$.   This variation of the template  includes the sketch geometry, topology and also the sketch parameters which gave rise to the specific variation of the geometry with embedding $z_i$.   We then use the simplex algorithm \cite{NeldMead65} to fine tune the parameters of the sketch so that the extracted profile has the highest IoU with the approximate shape decoded from $z(t)$.   This results in the CAD profiles shown in black in Figure \ref{fig:2d_interp}.
In regions where the same sketch is retrieved for consecutive frames, this procedure allows the parameters of the sketch to be varied smoothly, providing a very gradual distortion of the retrieved and fitted shape.  This behavior is best observed in the video in the supplementary material, as the gradual changes in parameters are difficult to show in static images.  

An important property of this procedure is that the retrieved and fitted shapes always have the appearance of a profile which a CAD designer would create.  The constraints inside the sketches maintain certain important aspects of the design intent like horizontal and vertical lines and tangent continuity.   While the algorithm attempts modify the sketch parameters to give rise to the smoothest interpolation possible, there are still sudden changes to the profile when the closest sketch template changes.  Collections of constrained sketches which minimize the size of these jumps in the shape during interpolation are desirable as they can be thought of as providing a better \emph{covering} of the space of possible target shapes.

\fi

\end{document}